%% file: thesis.tex
\documentclass[11pt]{gsasthesis} % 10,11 and 12pt fonts allowed

\usepackage{etex} % extend the number of registers

\usepackage[margin={1.2in}]{geometry}
\usepackage[titletoc]{appendix}

\usepackage{rotating}

\usepackage{longtable}

\usepackage{natbib}
\setcitestyle{authoryear,round}

\usepackage[sc]{mathpazo}
\usepackage{courier}

\usepackage[utf8]{inputenc}
\usepackage[T1]{fontenc}

\usepackage[english]{babel}
\usepackage{blindtext}

\usepackage{amsmath}

\usepackage{microtype}

\usepackage[stable,multiple]{footmisc}

\RequirePackage[font=small,format=plain,labelfont=bf,textfont=it]{caption}
\title{Mitigating Catastrophic Forgetting and Mode Collapse in Text-to-Image Diffusion via Latent Replay} % needs to match title on DAC
\author{Aoi Otani} % full name as it appears on your GSAS record, needs
\degreename{Bachelor of Arts}
\department{Department of Organismic and Evolutionary Biology} % official name of department

\principaladvisor{Professor Gabriel Kreiman}

\begin{document}

%%%%%%%%%%%%%%%% FRONTMATTER %%%%%%%%%%%%%%%%

\pagenumbering{roman} % GSAS wants roman page numbers for frontmatter

% the following four pages are required in that order. The first two pages are
% not allowed to have page numbers, this is taken care of in the class file.
%\thesistitlepage

\begin{abstract}
Continual learning—the ability to acquire knowledge incrementally without forgetting previous skills—is fundamental to natural intelligence. While the human brain excels at this, artificial neural networks struggle with "catastrophic forgetting," where learning new tasks erases previously acquired knowledge. This challenge is particularly severe for text-to-image diffusion models, which generate images from textual prompts. Additionally, these models face "mode collapse," where their outputs become increasingly repetitive over time.
To address these challenges, we apply \emph{Latent Replay}, a neuroscience-inspired approach, to diffusion models. Traditional replay methods mitigate forgetting by storing and revisiting past examples, typically requiring large collections of images. Latent Replay instead retains only compact, high-level feature representations extracted from the model's internal architecture. This mirrors the hippocampal process of storing neural activity patterns rather than raw sensory inputs, reducing memory usage while preserving critical information.
Through experiments with five sequentially learned visual concepts, we demonstrate that Latent Replay significantly outperforms existing methods in maintaining model versatility. After learning all concepts, our approach retained 77.59\% Image Alignment (IA) on the earliest concept, 14\% higher than baseline methods, while maintaining diverse outputs. Surprisingly, random selection of stored latent examples outperforms similarity-based strategies.
Our findings suggest that Latent Replay enables efficient continual learning for generative AI models, paving the way for personalized text-to-image models that evolve with user needs without excessive computational costs.
\end{abstract}

% Center headings for table of contents, LOT, and LOF and make them smaller so
% that "Abstract", "Acknowledgments" and "Contents" all look alike. Comment out
% if you want the default. If you want more control, use the "tocloft" package.
\renewcommand{\contentsname}{\protect\centering\protect\Large Contents}
\renewcommand{\listtablename}{\protect\centering\protect\Large List of Tables}
\renewcommand{\listfigurename}{\protect\centering\protect\Large List of Figures}

\tableofcontents % Table of contents

% The rest of the front matter: Lists of tables, figures, dedication and
% acknowledment is optional. Comment out whatever you don't like
\listoftables
\listoffigures
\pagenumbering{arabic} % reset page numbering and switch to arabic

% Introductory chapter. Comment out if you don't have an intro chapter, but I
% think most committees expect you to have one.
% Don't number the intro chapter, but add to to the table of contents
\addcontentsline{toc}{chapter}{Introduction}

\chapter{Introduction}\label{ch:1}
\input{chapter1}

\chapter{Background and Related Work}\label{ch:2}
\input{chapter2}

\chapter{Methods}\label{ch:3}
\input{chapter3}

\chapter{Results and Discussion}\label{ch:4}
\input{chapter4}

\chapter{Conclusion}\label{ch:5}
\input{chapter5}

%%%%%%%%%%%%%%%% BACK MATTER %%%%%%%%%%%%%%%%

% Put appendices, bibliography, and supplemental materials here

% The bibliography may be single spaced within each entry, but must be
% double-spaced between each entry. Most bibliography styles leave space between
% entries, so that shouldn't be a problem.
\begin{singlespacing}
  % I like "References" better than "Bibliography"
  \renewcommand{\bibname}{References}

  % Any bibliohgraphy style that leaves space between entries is fine
  \bibliographystyle{plainnat}
  \bibliography{references}
\end{singlespacing}

\input{appendix}
\end{document}

%% file: chapter1.tex
\textit{"Intelligence is the ability to adapt to change."} 
\hfill --- Stephen Hawking

\section{Motivation and Background}
\label{sec:motivation}

\noindent
Through evolutionary processes, humans and other living organisms have developed remarkable abilities to continuously learn, modify, integrate, and utilize knowledge in response to environmental changes. Inspired by this biological paradigm, artificial intelligence (AI) research seeks to impart adaptability similar to computational models. This objective underlies the field of \emph{continual learning} (CL), in which a model assimilates new tasks or data distributions sequentially while seeking to retain proficiency on earlier tasks~\citep{parisi2019continual}.

Over the past several years, large-scale generative models have become powerful tools for AI-driven content creation. Among these, text-to-image models have garnered particular attention, enabling the automated synthesis of high-quality images from natural language prompts. Recent advances in generative modeling center on \emph{diffusion models}, which iteratively denoise latent representations to produce realistic and diverse images. Notable examples include Denoising Diffusion Probabilistic Models (DDPM) \citep{ho2020denoising} and the latent-space variant known as Stable Diffusion \citep{rombach2022high}. These methods have led to breakthroughs in personalization approaches such as DreamBooth~\citep{ruiz2022dreambooth} and Custom Diffusion \citep{kumari2022multiconcept}, allowing the integration of user-specific concepts with only a few examples.

Despite their success, these generative models are not intrinsically designed for sequential, lifelong updates. Typically, personalization is done for one concept or for multiple concepts in a one-time parallel manner; repeated fine-tuning of the same network on new concepts can inadvertently overwrite parameters essential for older concepts. This phenomenon, known as \emph{catastrophic forgetting}~\citep{mccloskey1989catastrophic, goodfellow2013empirical}, becomes especially problematic in text-to-image diffusion, where the goal is not just to classify an input but to generate a rich and varied distribution of images. Even when a model retains some capacity for an older concept, it may collapse to a narrow slice of that concept’s diversity—a subtle form of forgetting referred to as \emph{mode collapse} \citep{zhang2024clog}.

\section{Text-to-Image Diffusion Models}

Text-to-image synthesis has evolved rapidly, from early GAN-based approaches \citep{goodfellow2014gan} to more stable and diverse diffusion models \citep{ho2020denoising}. These diffusion models typically learn a reverse noising process that reconstructs a clean sample from corrupted latent codes, conditioned on textual embeddings. Stable Diffusion~\citep{rombach2022high} represents a significant advancement by operating in a lower-dimensional latent space rather than in pixel space, dramatically improving computational efficiency while maintaining generation quality.

The model architecture integrates several components: an autoencoder $\left(E, D\right)$ that compresses images into compact latent representations, a U-Net denoiser conditioned on text embeddings (from models like CLIP), and a decoder that transforms the processed latents back into images. This approach has enabled unprecedented capabilities in text-guided image creation, with applications ranging from creative design to scientific visualization.

Fine-tuning techniques such as DreamBooth~\citep{ruiz2022dreambooth}, Textual Inversion~\citep{gal2022textual}, or Custom Diffusion~\citep{kumari2022multiconcept} can rapidly adapt stable diffusion models to new user-centric concepts from limited examples. However, they generally assume all new concepts are introduced at once, rather than in a series of incremental updates. Consequently, performing multiple rounds of sequential training on distinct concepts invites catastrophic forgetting of previously learned knowledge, with the added risk of reduced diversity in older concepts.

\section{What is Continual Learning?}

In continual learning, a model is trained on a sequence of tasks 
\(\{T_1, T_2, \dots, T_n\}\), each arriving at a different point in time. 
After learning task \(T_i\), the model must still perform well on tasks 
\(T_1\) through \(T_{i-1}\). Without dedicated strategies to preserve earlier knowledge, standard deep learning methods often exhibit catastrophic forgetting. Research in CL has proposed several broad solutions to mitigate forgetting, broadly categorized as regularization-based techniques (e.g., Elastic Weight Consolidation, EWC), parameter isolation via dynamic architectures, and replay-based methods that interleave old data (or synthetic samples) alongside new training data.

Generative models pose unique challenges for continual learning because they must reconstruct entire data distributions instead of merely predicting class labels. In text-to-image settings, even partial forgetting can lead to blurry or repetitive outputs that fail to reflect the breadth of previously learned concepts. Recent work underscores these difficulties. For instance, \citet{zhang2024clog} shows that although replay methods can help mitigate forgetting for generative tasks, they often fail to fully prevent it in diffusion-based models, and mode collapse remains a serious concern. Similarly, \citep{sun2023create} (citation example) highlights that naive replay alone is insufficient for large-scale diffusion settings. Nevertheless, replay methods still offer important benefits over naive fine-tuning, motivating further research into more sophisticated replay strategies.

\emph{Latent replay} has emerged as a particularly memory-efficient variant of replay-based approaches, originally introduced by Pellegrini et al.\citep{pellegrini2020latent} in classification contexts. Rather than storing the full-resolution images, latent replay retains compact latent codes or activations, drastically reducing memory overhead. This strategy can be especially appealing for text-to-image diffusion models, given the high costs of storing raw image data.

\begin{figure}[ht]
\centering
\includegraphics[width=\textwidth]{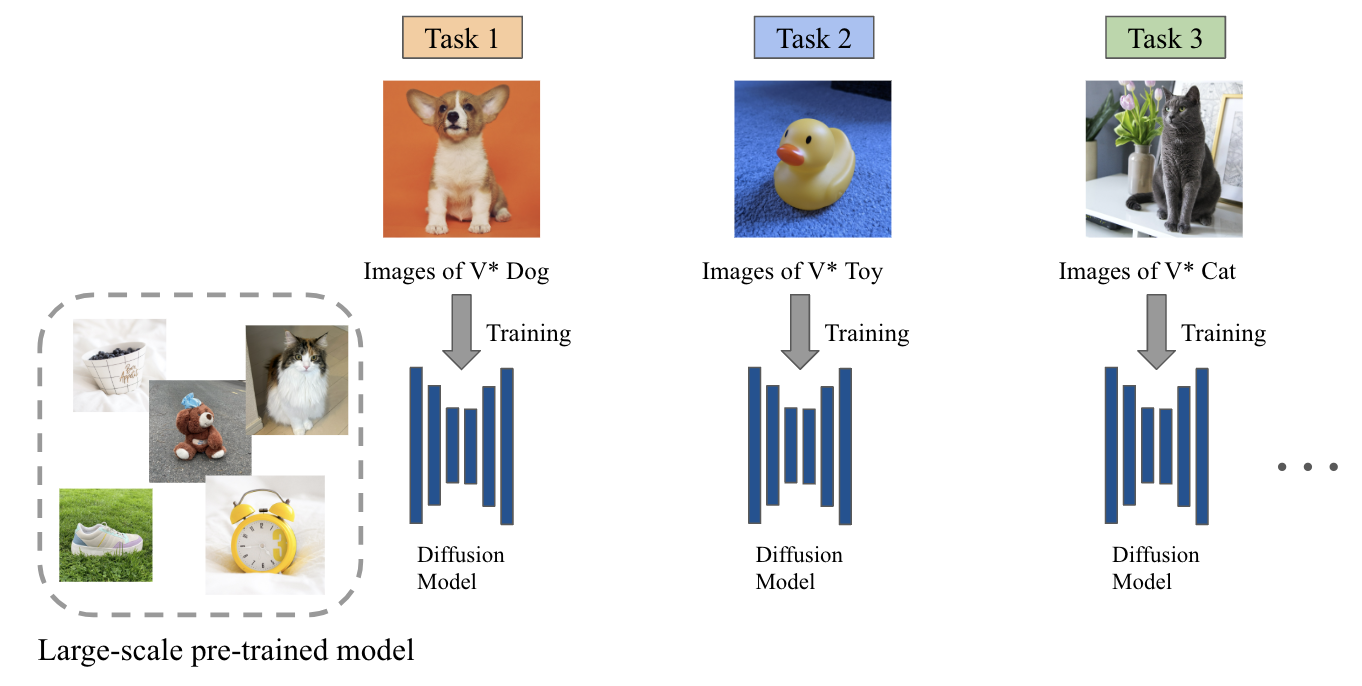}
\caption{Illustration of sequential fine-tuning in a text-to-image diffusion model. A pre-trained diffusion model is adapted to new concepts (dog, toy, cat) in a sequential manner.}
\label{fig:cl_diffusion}
\end{figure}

\section{Problem Statement}

Large text-to-image diffusion models, though capable of generating a vast range of high-quality images, are particularly prone to forgetting when fine-tuned sequentially on different tasks or concepts. As new data arrives, parameter updates can inadvertently overwrite those essential for generating older concepts. Even if some fidelity remains, the variety of outputs for older concepts may drop precipitously, indicating a form of mode collapse. Further complicating matters, storing full-resolution images for replay is often infeasible due to memory constraints, making it crucial to develop memory-efficient solutions to preserve generative diversity across tasks.

\section{Research Objectives}

The primary aim of this thesis is to determine whether \emph{latent replay} can be an effective way to mitigate catastrophic forgetting and mode collapse in text-to-image diffusion models. We focus on Stable Diffusion as our base architecture and investigate how to store and replay compressed latent codes of past images to remind the model of previously learned concepts. 
Our work is guided by four core objectives:
\begin{enumerate}
\item \textbf{Quantify catastrophic forgetting in diffusion models} when fine-tuned sequentially on new concept datasets. We will establish baseline measurements of how performance on generating earlier concepts deteriorates as new concepts are learned.
\item \textbf{Develop a latent replay mechanism} for text-to-image diffusion models. This involves determining what latent information to store (e.g., image latents, text embeddings, noise vectors) and how to utilize it during training of new tasks to rehearse past concepts.
\item \textbf{Evaluate the efficacy of latent replay in mitigating forgetting and mode collapse}, comparing it against other continual learning strategies or baselines. We will compare our approach to naive continual learning.
\item \textbf{Explore variants of latent replay}, including a similarity-based variant that selectively retrieves latents similar to the current training batch, examining whether this approach better preserves or inadvertently narrows prior knowledge.
\end{enumerate}

By operating in the latent space of a pre-trained model, we aim to efficiently refresh the model's memory of past concepts without requiring storage of full images or excessively large data. This approach is inspired by prior successes of latent replay, and we extend it to the generative diffusion domain.

\section{Thesis Contributions}

The core contribution of this thesis is the adaptation of latent replay—originally applied to classification tasks—to the domain of large-scale text-to-image diffusion. This fills a critical gap in generative continual learning by combining memory-efficient rehearsal with the need to maintain robust generation capabilities. Through extensive experiments, we show that latent replay often surpasses naive fine-tuning under tight memory constraints, preserving broader coverage of past concepts and reducing mode collapse. Our work also clarifies how similarity-based retrieval can interact with the generative manifold in ways that may be counterintuitive, offering insights that can guide future research on memory management in sequential generative tasks. By grounding the approach in both established replay methods and neuroscience-inspired ideas about rehearsal, we aim to bring text-to-image diffusion closer to genuine lifelong learning.

\section{Thesis Outline}

The remainder of this thesis is structured as follows. Chapter 2 delves further into the foundations of continual learning and generative models, summarizing key ideas and related work on catastrophic forgetting in text-to-image diffusion. Chapter 3 describes our methodology, detailing the experimental setup, fine-tuning procedures, and replay mechanisms. Chapter 4 presents quantitative and qualitative results, including statistical analyses of forgetting and mode collapse, and discusses the relative merits of latent replay compared to naive or image-based replay. Finally, Chapter 5 concludes with a summary of our findings and potential directions for future inquiry, including ways to scale latent replay to longer task sequences and to integrate more advanced sample selection strategies.

%% file: chapter2.tex
This chapter provides a comprehensive examination of the core concepts underlying this thesis. We begin by establishing the foundations of continual learning, exploring its basic formulation and the primary methods for mitigating catastrophic forgetting. We then examine text-to-image diffusion models, including their fundamental mechanisms and current personalization approaches. Next, we narrow our focus to the \emph{Continual Learning of Generative Models} (CLoG), highlighting the unique challenges of preserving both performance and diversity across sequential tasks. We identify a critical gap in existing methods for lifelong text-to-image generation, particularly regarding memory efficiency and mode collapse. Finally, we introduce \emph{Latent Replay} as our proposed solution and outline its theoretical advantages and the remaining gaps in the literature.

\section{Foundations of Continual Learning}
\label{sec:cl-theory-practice}

\subsection{Basic Formulation}
Continual learning involves learning a sequence of tasks $\mathcal{T}^{(1)}, \mathcal{T}^{(2)}, \dots, \mathcal{T}^{(T)}$ incrementally. Each task $\mathcal{T}^{(t)}$ has an input space $\mathcal{X}^{(t)}$, an output space $\mathcal{Y}^{(t)}$, and a training set
\begin{equation}
    \mathcal{D}^{(t)} = \{(x_j^{(t)}, y_j^{(t)})\}_{j=1}^{|\mathcal{D}^{(t)}|}
\end{equation}
drawn i.i.d.\ from the distribution $\mathcal{P}(\mathcal{X}^{(t)}, \mathcal{Y}^{(t)})$. The goal is to learn a function
\begin{equation}
    f: \bigcup_{t=1}^{T} \mathcal{X}^{(t)} \rightarrow \bigcup_{t=1}^{T} \mathcal{Y}^{(t)},
\end{equation}
which achieves good performance on every task $\mathcal{T}^{(t)}$, ensuring that the model retains its ability to perform previously learned tasks while acquiring new ones. While the specific criteria for performance may vary depending on the task, such as accuracy, alignment, or fidelity, the model should minimize degradation in its ability to meet the performance requirements specific to each task.

A key assumption in continual learning is that once a task is learned, its training data $\mathcal{D}^{(t)}$ is no longer fully accessible or is available only in limited form. This realistic constraint simulates the learning process of humans but also causes catastrophic forgetting in machine learning models, which refers to performance degradation on previous tasks due to parameter updates when learning new tasks~\citep{mccloskey1989catastrophic}. When neural networks are trained sequentially on multiple tasks, the weights optimized for earlier tasks are overwritten by updates for newer tasks, leading to rapid loss of previously acquired knowledge.

\subsection{Methods for Mitigating Catastrophic Forgetting}
To address catastrophic forgetting, researchers have proposed various strategies that can be broadly classified into three categories:

\subsubsection{Regularization-Based Approaches}
These methods add penalties to discourage significant deviations in parameters deemed crucial for previously learned tasks. For example, Elastic Weight Consolidation (EWC) \citep{kirkpatrick2017overcoming} leverages the Fisher Information Matrix, which measures how sensitive the model's outputs are to changes in each parameter, to quantify parameter importance and penalizes changes in those weights. Synaptic Intelligence (SI) \citep{zenke2017continual} tracks the contribution of each parameter over training and constrains updates accordingly. Learning without Forgetting (LwF) \citep{li2017learning} employs a distillation loss to preserve the output behavior on old tasks without storing their data.
These techniques primarily protect high-level parameters and are typically evaluated in classification rather than generative domains. For large diffusion models, direct parameter regularization can be either too weak, allowing forgetting, or too strong, impeding new concept learning~\citep{gao2023ddgr}. Moreover, knowledge-distillation-based approaches are more complex in a generative setting, as they must replicate the entire data distribution instead of just classifier outputs.

\subsubsection{Parameter-Isolation Methods}
These approaches partition the network so that each new task uses task-specific parameters. Progressive Networks \citep{rusu2016progressive} expand the model by adding a new column of neural network units (a new set of neurons and connections) for each task while keeping previously learned parameters fixed. Dynamically Expandable Networks \citep{lin2024dynamic} adaptively grow the architecture by introducing additional neurons or layers as needed to accommodate new tasks. Other techniques isolate parameters through masking or pruning \citep{mallya2018packnet, wortsman2020supermasks}, preventing updates for a new task from interfering with previous ones. Although parameter isolation can avoid catastrophic forgetting, it often limits cross-task generalization and leads to ever-growing model size. For text-to-image diffusion, isolation-based strategies would severely bloat an already large model, and still offer no mechanism for mixing previously learned concepts in new combinations.

\subsubsection{Replay-Based Approaches}
These techniques counter forgetting by interleaving data from previous tasks with new task data during training. Experience replay stores a limited buffer of real samples from each past task \citep{rebuffi2017icarl, riemer2018learning} and "replays" them alongside new data, with common sampling strategies including reservoir sampling and herding. Generative replay, instead of storing raw data, synthesizes examples from earlier tasks on demand using an often separate generative model \citep{shin2017continual, wu2018memorygan, cong2020ganreplay}. These generated samples are then used to reinforce old knowledge. Latent Replay stores intermediate representations, such as latent activations, rather than raw inputs \citep{pellegrini2020latent}, thereby reducing storage overhead while preserving essential information about previous tasks. In text-to-image diffusion, naively replaying high-resolution images is memory-intensive, while purely generative replay can degrade when the model inadvertently generates lower-quality samples over time---leading to compounding errors~\citep{zhang2024clog}. Thus, replay must be carefully designed to preserve both memory efficiency and distributional fidelity.

\subsection{Empirical Findings and Limitations}
Most continual learning research focuses on classification tasks, which involve low-dimensional, discrete outputs (class labels). Although well-suited for accuracy-based metrics, these methods do not fully address the more complex demands of high-dimensional, continuous outputs like those produced by generative models. When applied to generative tasks, standard continual learning approaches often fall short because they were designed primarily for discriminative boundaries rather than preserving entire distributions.

This limitation has motivated more specialized research into continual learning for generative models, which we explore after first establishing the foundations of text-to-image diffusion, the specific generative model domain that we focus on in this thesis.

\section{Text-to-Image Diffusion Models}

This section provides the foundation on text-to-image diffusion models necessary to understand how these models function and why they present both unique opportunities and challenges for continual learning.

\subsection{Diffusion Model Fundamentals}

Diffusion models have emerged as state-of-the-art methods for image generation. These models work by learning to reverse a gradual noising process, effectively recovering a clean image from pure noise. The breakthrough work of \citet{ho2020denoising} introduced Denoising Diffusion Probabilistic Models (DDPMs), which achieved image generation quality comparable to GANs while offering better mode coverage and training stability.

The diffusion process consists of two main phases:

\textbf{Forward Process:} A Markov chain that gradually adds Gaussian noise to data:
\begin{equation}
    q(x_t|x_{t-1}) = \mathcal{N}(x_t; \sqrt{1-\beta_t}x_{t-1}, \beta_t\mathbf{I})
\end{equation}
where \(\beta_t\) is a noise schedule.

\textbf{Reverse Process:} A learned denoising process that recovers the original data:
\begin{equation}
    p_\theta(x_{t-1}|x_t) = \mathcal{N}(x_{t-1}; \mu_\theta(x_t, t), \Sigma_\theta(x_t, t))
\end{equation}

In text-to-image settings, the model conditions on textual embeddings, allowing the denoiser to incorporate semantic constraints. This conditioning enables users to generate diverse images for different prompts without retraining from scratch.

\subsection{Latent Diffusion Models}
\citet{rombach2022high} introduced Latent Diffusion Models (LDMs) that operate in a lower-dimensional latent space rather than pixel space. An autoencoder $(E, D)$ is used: $E$ compresses images $x$ into latent codes $z$, and $D$ reconstructs images from these codes. The diffusion process itself happens in the latent space, significantly reducing computational overhead. A pretrained text encoder (e.g., CLIP from \citet{radford2021clip}) provides text embeddings that condition the U-Net denoiser via cross-attention. This approach underlies Stable Diffusion, which is known for generating high-quality, high-resolution images efficiently.

\subsection{Concept Customization in Diffusion Models}

Pretrained text-to-image models can be personalized to learn user-specific concepts (a new object, style, or person) from a handful of images. Methods such as DreamBooth~\citep{ruiz2022dreambooth}, Textual Inversion~\citep{gal2022textual}, and Custom Diffusion~\citep{kumari2022multiconcept} fine-tune or augment components of the diffusion pipeline with minimal data. However, these techniques often assume a single or simultaneous multi-concept update, not a sequential, lifelong scenario. Consequently, if multiple concepts are introduced one by one, catastrophic forgetting remains likely unless an explicit continual learning mechanism is used~\citep{smith2023continual}.

\section{Continual Learning of Generative Models}

In this section, we explore how continual learning applies specifically to generative models, including challenges and the specialized approaches that have been developed to address them.

\subsection{Formulation for continual learning of Generative Models}

Continual Learning of Generative Models (CLoG) follows a similar sequential training protocol but targets the learning of generative tasks
\(\mathcal{T}^{(1)}, \mathcal{T}^{(2)}, \dots, \mathcal{T}^{(T)}\) incrementally. Each task \(\mathcal{T}^{(t)}\) has an input space \(\mathcal{X}^{(t)}\) (generation conditions) and output space \(\mathcal{Y}\) (generation targets), and a training set

\[
\mathcal{D}^{(t)} = \{(x^{(t)}_j, y^{(t)}_j)\}_{j=1}^{|\mathcal{D}^{(t)}|}
\]

drawn \textit{i.i.d.} from the distribution \(\mathcal{P}(\mathcal{X}^{(t)}, \mathcal{Y}^{(t)})\).

The goal of CLoG is to learn a mapping:

\[
f : \bigcup_{t=1}^{T} \mathcal{X}^{(t)} \to \bigcup_{t=1}^{T} \mathcal{Y}^{(t)}
\]

that can achieve good performance on each task \(\mathcal{T}^{(t)}\).

The generation conditions can take various forms such as text \citep{li2019controllable, zhang2023controlnet}, images \citep{zhai2021lit, zhang2023controlnet}, or label indices \citep{ho2021classifier}, while the generation targets can span different modalities such as images \citep{ramesh2021zero, saharia2022photorealistic}, audio \citep{huang2018music, vandenOord2016wavenet}, or 3D objects \citep{shi2023deep, zeng2022lion}. In this thesis we focus on text-to-image generation, where textual descriptions serve as input conditions, and the model generates corresponding images as output.

\subsection{Unique Challenges in Continual Learning of Generative Models}

The key difference between classification-based continual learning and CLoG lies in the input space \( \mathcal{X} \) and output space \( \mathcal{Y} \). In image generation, the input \( x \) may be some label conditions (e.g., one-hot class) or instructions (text or images), and the output \( y \in \mathbb{R}^{C \times H \times W} \) should be images, where \( C \), \( H \), and \( W \) denote the number of channels, height, and width, respectively. CLoG is more challenging due to several factors. 

First, the output space inherently possesses a significantly larger cardinality, whereas the output of classification-based continual learning is typically limited to discrete class indices. Second, classification-based continual learning typically requires only a simple mechanism to model a categorical distribution, such as a linear mapping or a multilayer perceptron (MLP) head \citep{popescu2009mlp}, whereas CLoG necessitates more sophisticated generative models such as variational autoencoders (VAEs) \citep{kingma2013vae}, generative adversarial networks (GANs) \citep{goodfellow2014gan}, or score-based models \citep{song2020scorebased}.

A particularly unique challenge in CLoG is \emph{mode collapse}, where the model produces outputs with reduced diversity, effectively forgetting entire modes of the data distribution \citep{wu2018memorygan, zhang2024clog}. In a continual learning setting, even minor parameter shifts when learning new tasks can lead to the disappearance of previously learned modes, thereby compromising the overall generative diversity. This represents a subtle form of forgetting that may not be captured by traditional accuracy metrics but drastically affects the model's practical utility.

The problem of mode collapse also occurs due to task imbalance in continual learning scenarios. The inherent disparity between current task data (fully available) and previous task data (limited examples in the replay buffer) creates a situation where the model is disproportionately exposed to new data distributions. This imbalance can cause the model to gradually lose the ability to generate diverse outputs for earlier tasks.

Memory constraints further compound the issue, as replay buffer limitations impose severe constraints on retaining previous data. Only a small subset of earlier examples can be stored, and \citet{zhang2024clog} report that replay-based methods often struggle to preserve the full range of modes when memory is tight, potentially compounding the loss of generative diversity over time.

Additionally, traditional replay-based strategies can sometimes lead to replay-driven overfitting on the limited examples available in the replay buffer. This causes the model to repeatedly train on low-quality or redundant samples, ultimately leading to both catastrophic forgetting and diminished generative diversity. 

Empirical studies reveal that different generative architectures exhibit varying susceptibility to these challenges. GANs are particularly prone to mode collapse in CLoG settings, often generating nearly identical outputs for previous concepts after sequential training. Diffusion models demonstrate better resilience but still suffer from diversity loss, especially when replay buffer sizes are severely constrained. Moreover, traditional continual learning metrics primarily focus on accuracy or fidelity but fail to capture diversity loss, highlighting the need for diversity-sensitive metrics that can detect mode collapse even when task-specific accuracy remains high.

These findings demonstrate that preserving both model performance and output diversity requires specialized approaches beyond traditional continual learning methods, particularly for generative models that must maintain diverse outputs across sequential tasks.

\subsection{Current Approaches in CLoG}
Several methods have been proposed to address the challenges of continual learning in generative models. \citet{cong2020ganreplay} introduced GAN Memory, a technique that uses specialized architecture modulations to prevent forgetting in GANs. The approach maintains task-specific statistics and employs them to modulate feature maps during generation, helping to preserve the distinct characteristics of each learned task.
\citet{seo2023lfsgan} proposed Lifelong Few-Shot GAN, which employs Low-rank Factorized Transformations to efficiently adapt a pre-trained GAN to new tasks with minimal forgetting. This method decomposes network adaptations into low-rank components that can be selectively activated for different tasks, allowing parameter-efficient continual learning.
For diffusion models, \citet{gao2023ddgr} designed DDGR (Diffusion-based Dynamic Generative Replay), a classifier-guided approach that generates synthetic examples from previously learned tasks to mitigate forgetting. By incorporating task-specific guidance during the sampling process, DDGR can generate diverse examples representative of earlier tasks.
While these approaches help maintain knowledge of earlier tasks, none fully solve the dual challenges of catastrophic forgetting and mode collapse, especially under tight replay-memory constraints. This persistent gap motivates our exploration of Latent Replay as a potential solution for text-to-image diffusion models.

\section{The Gap in Lifelong Text-to-Image Generation}
\label{sec:gap}

The intersection of text-to-image diffusion models and continual learning represents an important but underexplored research area. While text-to-image diffusion models have demonstrated remarkable capabilities for concept customization, adapting them for sequential learning presents several unresolved challenges.

As mentioned above, standard personalization methods such as DreamBooth~\citep{ruiz2022dreambooth}, Textual Inversion~\citep{gal2022textual}, and Custom Diffusion~\citep{kumari2022multiconcept} suffer from catastrophic forgetting when multiple concepts are introduced sequentially.

More recent parameter-efficient adaptation techniques for diffusion models, such as C-LoRA~\citep{smith2023continual} and STAMINA~\citep{smith2024stamina}, aim to improve scalability across multiple tasks. However, these approaches focus primarily on reducing parameter overhead rather than explicitly addressing the mode collapse problem that arises in replay-limited environments, as identified by \citet{zhang2024clog}.

The high-resolution outputs of modern diffusion models create significant memory challenges for traditional replay-based continual learning. Storing raw high-resolution images (typically 512×512 or larger) for replay quickly becomes impractical as the number of concepts increases. Using a limited set of examples risks accelerating overfitting and mode collapse if the buffer is too small or unrepresentative of the full concept distribution.

Additionally, the multi-stage denoising process of diffusion models adds complexity to the replay mechanism. Representations at different noise levels may capture different aspects of a concept, raising questions about the optimal point in the diffusion process to store and replay information. This technical complexity has not been thoroughly addressed in existing continual learning approaches for diffusion models.

These unresolved challenges create a clear need for memory-efficient replay strategies specifically designed for diffusion-based generative models. Such methods should maintain both fidelity and diversity in continual text-to-image generation settings while working within practical memory constraints.

\section{Latent Replay for Continual Learning}
\label{sec:latent-replay}

Latent Replay represents a promising approach for continual learning in text-to-image diffusion models. Originally introduced by \citet{pellegrini2020latent} for classification tasks, it has shown impressive results in maintaining performance while significantly reducing memory requirements.

In Pellegrini's original approach, the term Latent Replay refers to storing activations at intermediate layers of the network rather than raw input data. Instead of keeping past input examples (like images) in memory, their method captures and stores activation volumes from a middle layer of the network. When learning new tasks, these stored intermediate representations are injected directly at the corresponding layer, bypassing the need to recompute the forward pass through earlier layers. To maintain representation stability and ensure the stored activations remain valid, they proposed slowing down learning in all layers below the Latent Replay point, while allowing layers above to learn at full speed. This technique was specifically designed to enable continual learning on resource-constrained devices like smartphones, where storing and processing raw data would be prohibitively expensive.

\subsection{Core Concept and Memory Efficiency}
Replay-based strategies mitigate forgetting by periodically reminding the model of older tasks through exposure to previously learned examples. However, in large-scale text-to-image diffusion, storing raw images is often prohibitive in terms of memory requirements. Latent Replay addresses this limitation by storing compressed representations extracted from the model's encoder or an intermediate network layer, drastically reducing memory overhead.
This approach is particularly efficient in the context of Stable Diffusion, where a $512\times512$ RGB image (approximately 3MB) is compressed to a $64\times64\times4$ latent representation (approximately 64KB), cutting storage requirements by roughly $98\%$. Such compression allows for maintaining a much larger and more diverse set of examples within the same memory budget.

\subsection{Challenges in Application to Generative Models}
While promising, adapting Latent Replay from classification to generative diffusion models presents several technical challenges. Originally proposed for classification tasks, Latent Replay has primarily served discriminative goals where maintaining decision boundaries is sufficient. Generative tasks, however, require preserving detailed distributional information, as the model must capture the full range of possible outputs rather than just class separations.
This distinction places higher demands on the information content of stored representations. Inadequate preservation of feature details can lead to distorted or low-diversity reconstructions. Additionally, text-to-image models operate at multiple noise levels, and the alignment between latent codes and text embeddings must be maintained throughout the denoising process. This creates complexity in determining how and when to inject replayed latents during training.
Despite these challenges, the efficiency and potential benefits of Latent Replay make it a compelling approach for addressing continual learning in text-to-image diffusion models. In the next chapter, we describe our methodology for implementing and evaluating Latent Replay for this specific application.

\section{Summary and Research Direction}
\label{sec:summary-ch2}

This chapter provided an overview of continual learning, with an emphasis on the unique challenges in generative settings. The high-dimensional outputs of image-generation models complicate attempts to mitigate catastrophic forgetting, and the phenomenon of mode collapse further exacerbates the issue. Text-to-image diffusion models, such as Stable Diffusion, are especially relevant due to their success and widespread adoption in various domains, yet they are not intrinsically designed for sequential concept learning.
Traditional continual learning methods often do not fully resolve the memory and diversity challenges seen in lifelong text-to-image tasks. Latent Replay presents a promising pathway: it stores compact latent codes rather than full-resolution images, making it more scalable while retaining enough distributional information to remind the model of older modes. This approach naturally aligns with the compressed nature of latent diffusion models.
In the following chapter, we describe our proposed methodology for experiments including implementation of Latent Replay. We also discuss how we measure performance in terms of alignment and diversity, both crucial for evaluating the quality and breadth of generative outputs in a continual learning scenario.

%% file: chapter3.tex
This chapter presents an application of Latent Replay (LR), a strategy designed to mitigate catastrophic forgetting in continual learning. Our objective is to introduce LR for the first time in text-to-image diffusion models, evaluating its effectiveness in reducing forgetting while preserving output diversity. We compare LR against standard approaches including Naive Fine-tuning (Naive) and Experience Replay (ER), while also proposing a novel variant called Similarity-Based Latent Replay (SLR). We examine how these methods perform across different memory constraints and learning scenarios, ensuring the model retains past knowledge while continuing to acquire new information.
%%%%%%%%%%%%%%%%%%%%%%%%%%%%%%%%%%%%%%%%%%%%%%

\section{Datasets and Preprocessing}
\label{sec:dataset-and-preprocessing}

\subsection{Data Collection and Structure}
Following DreamBooth~\citep{ruiz2022dreambooth} and Custom Diffusion~\citep{kumari2022multiconcept}, we selected five object types representing distinct image generation concepts \emph{dog}, \emph{toy}, \emph{cat}, \emph{backpack}, \emph{plushie}, covering diverse visual domains and semantic relationships. For each concept, we collected 8--15 high-quality RGB images from public repositories, showcasing the concept from multiple angles, environments, and exemplars. The relatively small image count per concept aligns with the few-shot personalization paradigm of DreamBooth.

For each concept, we created 20 diverse prompt templates (e.g., "a photo of V* concept", "V* concept in a garden", "an oil painting of V* concept") with a unique identifier token "V*" in the text prompts (e.g., "a photo of V* cat"). The special token V* serves as a learnable identifier within the text encoder's embedding space, allowing the model to form a unique association between this rare token and the visual characteristics of each concept, following the approach established in DreamBooth~\citep{ruiz2022dreambooth}. This resulted in approximately 160-300 training samples per concept (8-15 images × 20 prompts), while still maintaining the few-shot nature of the approach in terms of unique visual examples.

\subsection{Preprocessing Pipeline}
We apply a standardized pipeline to each image:
\begin{enumerate}
    \item Resize to \(512\times512\) pixels (maintaining aspect ratio when possible).
    \item Center crop to ensure a square aspect ratio.
    \item Convert to RGB if necessary.
    \item Normalize to the \([-1,1]\) range, suitable for the Variational Autoencoder (VAE) latent representation.
\end{enumerate}

During training, each image is periodically paired with multiple prompts in a cyclical manner to ensure broad coverage and improve the model's generalization across different contexts and styles.

\subsection{Memory Usage and Representation}

To compare ER and LR fairly, we track their respective memory footprints.

\begin{table}[h]
    \centering
    \caption{Memory Requirements for ER and LR}
    \label{tab:memory-usage}
    \begin{tabular}{l c c}
        \hline
        \textbf{Method} & \textbf{Storage Type} & \textbf{Memory per Sample} \\
        \hline
        ER & \(512\times512\) float32 image & \(\sim3\) MB \\
        LR & \(64\times64\times4\) latent representation & \(\sim64\) KB \\
        \hline
    \end{tabular}
\end{table}

The ratio of raw image size to latent size is approximately \(48:1\). To ensure fair comparisons, we define different memory settings where both ER and LR use comparable total storage:

\begin{table}[h]
    \centering
    \caption{Memory Allocation Across Different Buffer Sizes}
    \label{tab:memory-settings}
    \begin{tabular}{l c c}
        \hline
        \textbf{Memory Setting} & \textbf{ER (Images)} & \textbf{LR (Latents)} \\
        \hline
        Small  & 10 images (\(\sim30\) MB)  & 480 latents (\(\sim30\) MB) \\
        Medium & 20 images (\(\sim60\) MB)  & 960 latents (\(\sim60\) MB) \\
        Large  & 100 images (\(\sim300\) MB) & 4800 latents (\(\sim300\) MB) \\
        \hline
    \end{tabular}
\end{table}

Because the specific model we use, \texttt{CompVis/stable-diffusion-v1-4} (a widely-used implementation of Stable Diffusion), is pretrained on a broad distribution of Internet images, certain common categories (e.g., \emph{dog} or \emph{cat}) may be relatively easier for the model to learn and retain. To leverage these pretrained priors, we freeze the text encoder and VAE while fine-tuning only the U-Net component.

\section{Baselines and Ablation Studies}

We evaluate the effectiveness of LR by comparing it against several continual learning approaches for sequential text-to-image generation tasks. Our study includes five methods: Naive, ER, LR, SLR, and an Offline Upper Bound. Each method is implemented using the same underlying model architecture and training protocol to ensure fair comparison.

All replay-based methods (ER, LR, and SLR) share a common approach to balancing current and past tasks through a weighted loss function:
\begin{equation}
    \mathcal{L}_{\text{total}} = (1-\lambda_{\text{memory}})\mathcal{L}_{\text{current}} + \lambda_{\text{memory}}\mathcal{L}_{\text{memory}}
\end{equation}
where $\mathcal{L}_{\text{current}}$ is computed on current task samples and $\mathcal{L}_{\text{memory}}$ is computed on samples retrieved from the memory buffer. Both loss terms represent diffusion denoising losses, measuring how well the model predicts the noise added during the forward diffusion process. During training, we optimize the model's parameters to minimize this combined loss, allowing it to learn new concepts while retaining knowledge of previous ones. We set $\lambda_{\text{memory}} = 0.5$ for all experiments unless otherwise specified, providing equal weight to both current and past tasks—a balance we found effective in our ablation studies for managing the trade-off between retaining previous knowledge and acquiring new information.

\subsection{Naive Fine-tuning (Naive)}
Naive Fine-tuning (Naive) sequentially fine-tunes the model on each task without any memory retention, serving as a baseline for catastrophic forgetting. The implementation follows standard fine-tuning procedures for diffusion models, where the U-Net backbone is updated while the VAE and text encoder remain frozen. For each task, the model optimizes the denoising loss on current examples without any mechanism to preserve knowledge of previous tasks.

\subsection{Experience Replay (ER)}
Experience Replay (ER) retains a subset of previously seen images and reintroduces them during training to reinforce past knowledge. To maintain a representative distribution of past tasks, memory updates use reservoir sampling (a technique that maintains a fixed-size random sample over a data stream without knowing the stream's size in advance). During training, stored samples are replayed alongside new task data, contributing to the weighted loss function described above to mitigate forgetting.

\subsection{Latent Replay (LR)}
Latent Replay (LR) is a memory-efficient alternative to ER for classification tasks. We adapt this approach to the text-to-image diffusion domain. Instead of storing raw images, LR retains compact latent representations extracted from the VAE, significantly reducing storage requirements while maintaining high visual fidelity.

In our implementation, the VAE-encoded latent vectors (dimension $4 \times 64 \times 64$) are stored along with their corresponding prompt tokens. During training, for each batch of current task data (batch size 1), we retrieve an equal-sized batch of latent vectors from the memory buffer. These replayed latents are processed alongside current task samples, using the same weighted loss approach as ER. 

Our memory management employs reservoir sampling to ensure an unbiased selection of stored latents, allowing the model to retain prior knowledge while still adapting efficiently to new tasks.

\begin{figure}[htbp]
    \centering
    \includegraphics[width=\textwidth]{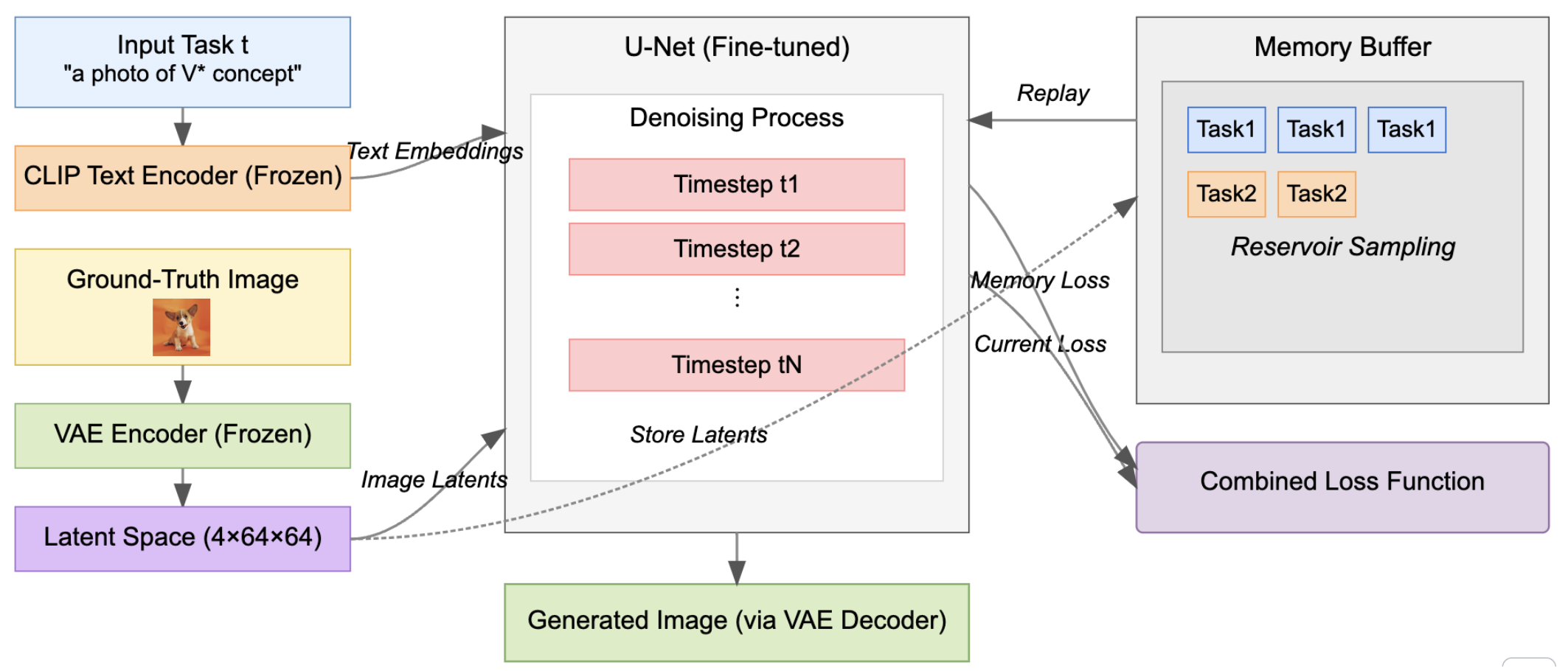}
    \caption{Architectural overview of Latent Replay for text-to-image diffusion models. The input concept is processed through the frozen CLIP text encoder and VAE encoder, producing latent representations (4×64×64). These latents flow to the fine-tuned U-Net for the denoising process and are also stored in a memory buffer using reservoir sampling. During training, latents from previous tasks are retrieved from memory and replayed along with current task latents. The model optimizes a combined loss function that balances learning new concepts while preserving knowledge of past ones.}
    \label{fig:latent-replay}
\end{figure}

\subsection{Similarity-based Latent Replay (SLR)}
Building on standard LR, we propose a novel approach called Similarity-based Latent Replay (SLR), which employs a more selective approach to memory retrieval based on the semantic relevance of stored latent vectors to the current training batch. The key intuition behind our proposed SLR is that not all previous examples are equally relevant to the current learning task, and retrieving semantically similar latents may provide more effective knowledge transfer with less interference.

Our implementation of SLR differs from standard LR in the retrieval mechanism:

\begin{enumerate}
    \item \textbf{Query Formation:} For each batch of current task samples, we extract and average their latent representations to form a query vector:
    \begin{equation}
        \mathbf{q} = \frac{1}{|\mathcal{B}|}\sum_{\mathbf{x} \in \mathcal{B}} E(\mathbf{x})
    \end{equation}
    where $\mathcal{B}$ is the current batch and $E$ is the VAE encoder.
    
    \item \textbf{Similarity Computation:} We calculate the cosine similarity between this query and all latent vectors $\mathbf{z}_i$ in the memory buffer.
    
    \item \textbf{Top-$k$ Selection:} We select the $k$ latents with highest similarity scores for replay in the current training step, where $k$ is a hyperparameter (typically set to 4).
\end{enumerate}

This approach differs from random selection in standard LR, which draws samples uniformly from the buffer regardless of relevance to the current training examples. SLR tests the hypothesis that semantically related examples from previous tasks provide stronger signals for preserving task-specific knowledge without sacrificing adaptability to new tasks. For example, when learning the \emph{cat} concept after \emph{dog}, SLR might prioritize retrieving dog latents that share visual characteristics with cats (such as similar poses or backgrounds), potentially creating stronger anchors for preserving both concepts. The same weighted loss function is used as with other replay methods.

\subsection{Offline Upper Bound}
The Offline Upper Bound serves as an empirical upper bound for performance by training on all tasks simultaneously rather than sequentially. Unlike the continual learning approaches, the Offline Upper Bound has access to all task data at once, eliminating the challenge of catastrophic forgetting. During training, we create combined datasets of all tasks up to the current task, ensuring equal representation in each training batch. The Offline Upper Bound provides a ceiling for the performance achievable by continual learning methods, as it represents the ideal scenario where all data is available simultaneously.

\subsection{Ablation Studies}

Ablation studies systematically isolate and evaluate the contribution of individual components or hyperparameters of our proposed methods. By selectively modifying or removing specific elements while keeping others constant, we determine the relative importance of each factor and identify optimal configurations. In our work, we conducted several ablation studies to thoroughly analyze the impact of various design choices:

\paragraph{Memory Size Variation.}
We tested the three memory configurations described in Table~\ref{tab:memory-settings}: small (10 images/480 latents), medium (20/960), and large (100/4800) buffers to determine how buffer size affects forgetting mitigation and mode collapse prevention across all methods.

\paragraph{Replay Weight Analysis.}
We experimented with \(\lambda_{\text{memory}}\) in \(\{0.1,0.3, 0.5,0.7,0.9\}\). Lower weights speed new-task learning but risk older-task forgetting; higher weights preserve older tasks but slow new-task adaptation. 

\paragraph{Training Stability Factors.}
We explored how thresholding at loss=1.0 vs. 1.5, fixed vs. random replay batches, and up to five restarts per task influenced final performance. 

\paragraph{Task Order Effects.}
To examine how the sequence of tasks affects forgetting patterns, we evaluated alternative orderings of our five concepts. Specifically, we compared the original sequence (dog \(\to\) toy \(\to\) cat \(\to\) backpack \(\to\) plushie) with the reversed order (plushie \(\to\) backpack \(\to\) cat \(\to\) toy \(\to\) dog), measuring IA, TA, and Diversity metrics across all methods. This investigation helps determine whether forgetting is more influenced by temporal position or by specific semantic relationships between sequentially learned concepts.

\subsection{Training Process}
The training loop for all methods follows a similar structure, with method-specific modifications for replay mechanisms. Given a sequence of tasks, training is conducted as follows:
\begin{enumerate} 
    \item For each task, a dataset of concept-specific images and prompts is constructed. 
    \item Images are encoded into the latent space using the frozen VAE. 
    \item Random noise is added to the latent representations according to the diffusion schedule. 
    \item The U-Net is trained in mini-batches of size 1 using a DataLoader that shuffles the task-specific dataset. The model predicts the noise conditioned on text embeddings, with gradient accumulation every 4 steps to simulate larger batch sizes while maintaining memory efficiency. This process continues for multiple epochs until reaching the maximum training steps.
    \item After training on each task, we evaluate performance by generating ten images per learned concept using the standard prompt: "a photo of V* concept"
    where V* is the learned identifier token. 
\end{enumerate}

For replay-based methods, additional steps are incorporated:
\begin{enumerate}
    \item[6.] Samples are retrieved from memory (either raw images for ER or latent representations for LR).
    \item[7.] A separate replay loss is computed on the retrieved samples.
    \item[8.] The total loss is computed as a weighted sum of the current task loss and the replay loss.
\end{enumerate}

Training stability is monitored by tracking the loss throughout the training process. If instability is detected (i.e., the loss exceeds a predefined threshold of 1.0–1.5, depending on the method), training is restarted from scratch. 

\section{Training Configuration}

We fine-tune a Stable Diffusion model, initialized with \texttt{CompVis/stable-diffusion-v1-4} hosted on Hugging Face \citep{compvis_sd}, on a sequence of five tasks. Each task corresponds to a distinct concept (e.g., \emph{dog}, \emph{toy}, \emph{cat}, \emph{backpack}, \emph{plushie}) and includes a small set of representative images plus 20 text prompts. The model is trained using configurations from prior literature~\citep{sun2023create}, as summarized in Table~\ref{tab:training-config}.

\begin{table}[h]
    \centering
    \caption{Training Configuration for Fine-Tuning Stable Diffusion}
    \label{tab:training-config}
    \begin{tabular}{l l}
        \hline
        \textbf{Parameter} & \textbf{Value} \\
        \hline
        Model & Stable Diffusion (\texttt{CompVis/stable-diffusion-v1-4}) \\
        Batch size & 1 \\
        Optimizer & AdamW (8-bit) \\
        Learning rate & \(1\times10^{-4}\) \\
        Weight decay & \(1\times10^{-2}\) \\
        Gradient Accumulation & 4 steps \\
        Mixed Precision & fp16 \\
        Gradient Clipping & Max norm = 1.0 \\
        Max Steps per Task & 800 \\
        Warm-up Steps & 50 \\
        Fine-tuned Component & U-Net only (860M parameters) \\
        VAE & Frozen \\
        Text Encoder & Frozen \\
        \hline
    \end{tabular}
\end{table}

Throughout training, we monitor loss to detect instability and ensure training convergence. For Naive and Offline, if the loss exceeds 1.0, training is restarted. For Replay-based Training methods, if the combined loss exceeds 1.5, training is restarted.

Each task may be retried up to five times with a new random seed if the loss remains unstable. Additionally, we enforce a minimum of 100 steps before allowing early termination, ensuring each task is sufficiently learned.

\section{Evaluation Metrics}

To assess the effectiveness of our approach, we evaluate three key metrics: Image Alignment (IA), Text Alignment (TA), and Diversity. Additionally, we compute the Task Forgetting Rate (TFR) to quantify performance degradation across sequential learning tasks. We further conduct a qualitative analysis of both catastrophic forgetting and mode collapse, providing visual inspections to complement quantitative results.

\subsection{Image Alignment (IA)}
Image Alignment (IA) measures the similarity between generated images and reference images from the training dataset. Following the methodology of DreamBooth~\citep{ruiz2022dreambooth} and Custom Diffusion~\citep{kumari2022multiconcept}, we compute IA using CLIP embeddings. Specifically, we extract feature representations using a pretrained CLIP-L model~\citep{radford2021clip} and compute cosine similarity between generated images and their corresponding training images. The final IA score is obtained by averaging cosine similarity across all samples.

\subsection{Text Alignment (TA)}
Text Alignment (TA) evaluates how well the generated images align with their corresponding textual descriptions. We measure TA using CLIP-based similarity between image embeddings and text embeddings, ensuring that the generated outputs accurately represent their associated prompts.

\subsection{Task Forgetting Rate (TFR)}
To fairly evaluate generation performance in our lifelong learning setting, we generate ten images per concept per prompt for each generation task. After the model has observed all tasks, we adopt the following metrics to assess lifelong generation:

\begin{itemize}
    \item \textbf{TFR-IA (Forgetting Index for Image Alignment)}: This metric quantifies the decrease in IA scores over sequential tasks. Following~\cite{smith2023continual}, it is computed as:
    \[
    \text{TFR-IA} = \frac{1}{k-1} \sum_{\ell=1}^{k-1} I_{\ell, \ell} - I_{k, \ell}
    \]
    where \( I_{\ell, \ell} \) represents the IA score for the \(\ell\)-th generation task immediately after training, and \( I_{k, \ell} \) denotes the IA score for the \(\ell\)-th task after learning the \(k\)-th task.

    \item \textbf{TFR-TA (Forgetting Index for Text Alignment)}: This metric follows the same formulation but evaluates text alignment degradation:
    \[
    \text{TFR-TA} = \frac{1}{k-1} \sum_{\ell=1}^{k-1} T_{\ell, \ell} - T_{k, \ell}
    \]
    where \( T_{\ell, \ell} \) and \( T_{k, \ell} \) denote TA scores before and after learning the \(k\)-th task, respectively.
\end{itemize}

In practice, these TFR values are computed offline by referencing the stored IA and TA scores after each task. They summarize how much performance has deteriorated for earlier tasks.

\subsection{Diversity} 
Diversity measures whether the model collapses to generating repetitive samples. We employ the Vendi Score~\citep{naeem2022vendi}, a metric specifically designed to quantify diversity in machine learning-generated outputs.
The Vendi Score is based on entropy estimates in high-dimensional spaces and evaluates intra-task variation within generated samples. Unlike traditional diversity measures that rely on pairwise distances (e.g., LPIPS), the Vendi Score is scale-invariant and applicable to arbitrary feature spaces. Formally, given a set of generated samples \(\mathcal{S} = \{s_1, s_2, \dots, s_n\}\), the Vendi Score is computed as:

\[
\mathcal{V}(\mathcal{S}) = \exp\left(\frac{1}{n} \sum_{i=1}^{n} \log \frac{1}{d_i}\right)
\]

where \( d_i \) represents the local density of sample \( s_i \) in the feature space, calculated based on the distances to neighboring points. A higher Vendi Score indicates greater intra-task diversity, reducing concerns about mode collapse in generated images.

\subsection{Qualitative Analysis}

In addition to quantitative evaluations, we conduct a qualitative analysis of catastrophic forgetting by visually inspecting the generated outputs for Task 1 (\emph{dog}) after training on each task (e.g., generation output for the \emph{dog} task after completing the \emph{cat} task). This provides insights into how early learned concepts degrade over time.

To further analyze intra-task diversity, we visually inspect four generated images after learning each task, showcasing how variations occur from each task. By displaying four outputs instead of one, we assess whether the model successfully generates diverse samples within a task or collapses to producing nearly identical outputs. 

\section{Experimental Setup}

Experiments run on a GPU-accelerated cluster (single GPU, 16 CPU cores) with SLURM-based scheduling. Each experimental configuration is repeated with ten different random seeds to gauge stability. We report mean and standard deviation of IA, TA, and Diversity across these ten runs, as well as forgetting curves showing how earlier tasks degrade after subsequent training.

This setup comprehensively assesses both numerical performance and memory efficiency of Naive, ER, LR, SLR, and the Offline upper bound.

\section{Statistical Analysis}
We compute the mean and standard deviation of IA, TA, and Diversity scores across the ten runs with different random seeds. To rigorously assess whether the observed performance differences between methods are statistically significant, we employ the non-parametric Wilcoxon signed-rank test with significance level $\alpha = 0.05$. This test is appropriate given the potential non-normality of our metrics and the relatively small sample size of ten runs per method, as it makes no assumptions about the underlying distributions.

For each metric (IA, TA, Diversity) and each task, we performed pairwise comparisons between methods using a two-sided Wilcoxon signed-rank test with significance level $\alpha = 0.05$. Additionally, we applied the Benjamini-Hochberg procedure to control the false discovery rate across multiple comparisons.

Our analysis particularly focuses on statistically validating the comparative advantages of LR over both Naive and ER, as well as examining the relative performance of different LR variants. Results of these statistical comparisons are presented in Chapter 4 alongside our quantitative performance measurements.

%% file: chapter4.tex
In this chapter, we present and analyze our quantitative and qualitative findings from training five methods --- \textit{Naive Fine-tuning (Naive)}, \textit{Experience Replay (ER)}, \textit{Latent Replay (LR)}, \textit{Similarity-Based Latent Replay (SLR)}, and an \textit{Offline} upper bound --- on a sequence of five tasks (\emph{dog}, \emph{toy}, \emph{cat}, \emph{backpack}, and \emph{plushie}). We evaluate performance according to three main metrics introduced in Chapter~3: Image Alignment (IA), Text Alignment (TA), Diversity, and a qualitative analysis. Each method was trained over ten runs with different random seeds. For our main analysis, we focus on the most memory-constrained condition with a small buffer setting: 10 images (\(\sim30\) MB) for ER and 480 latents (\(\sim30\) MB) for LR and SLR. We also present results from ablation studies with larger buffer sizes.
%(make this more comprehensive)

\section{Quantitative Results}
\label{sec:quant_results}

\subsection{Metric Plots}

\begin{figure}[ht]
    \centering
    \includegraphics[width=\linewidth]{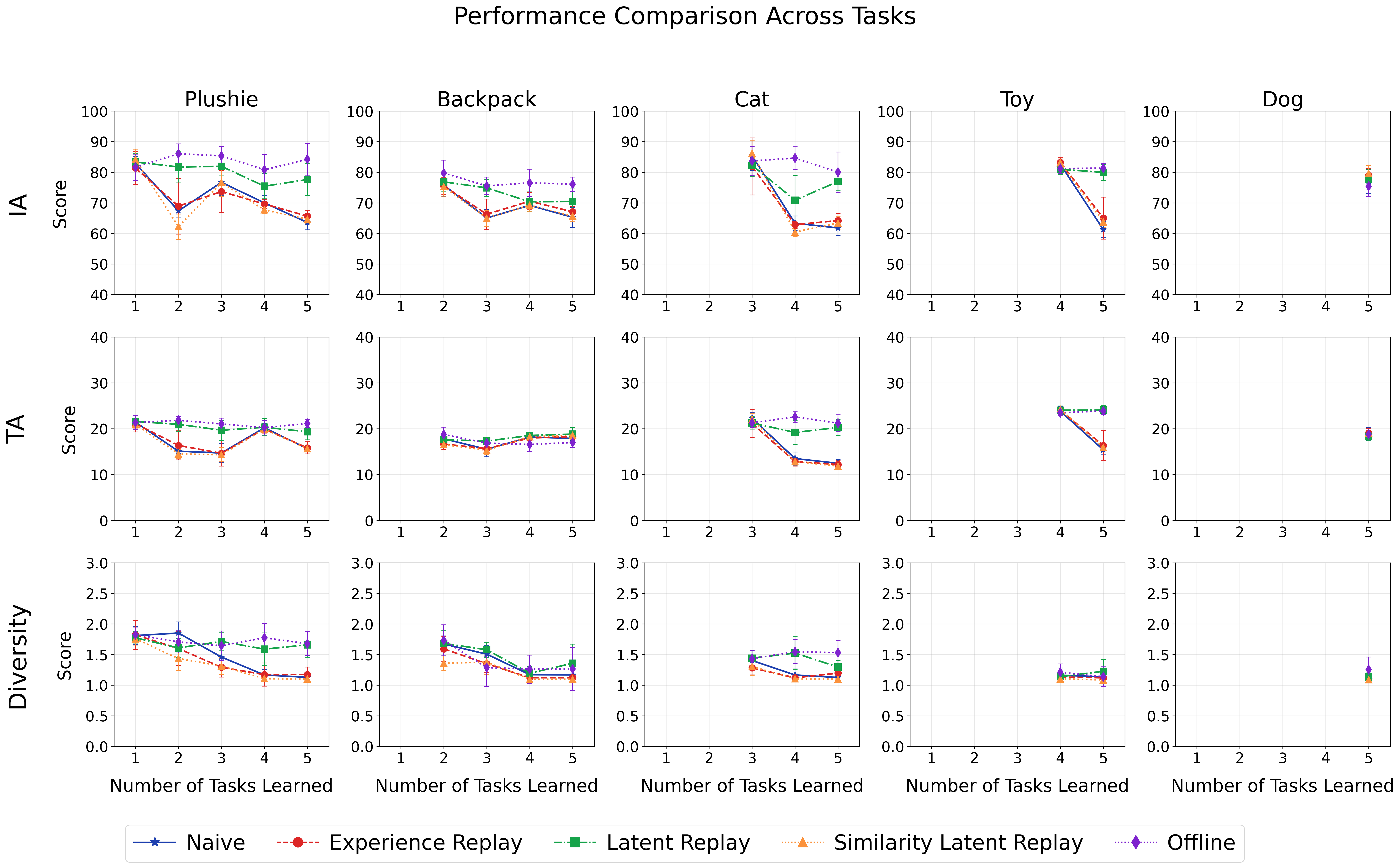}
    \caption{Performance metrics across sequential tasks showing LR's consistent advantages. Top: Image Alignment, where LR maintains higher visual fidelity to reference images. Middle: Text Alignment, where LR preserves stronger prompt-image correspondence. Bottom: Diversity scores, where LR prevents mode collapse with variation approaching Offline training. In all metrics, LR significantly outperforms Naive, ER, and SLR methods, especially for earlier concepts, despite using limited memory. Error bars represent one standard deviation across ten runs.}
    \label{fig:combinedplots}
\end{figure}

Figures~\ref{fig:combinedplots} show how IA, TA, and Diversity change as new tasks are introduced, using the small buffer size across ten runs. Each figure has subplots for each concept (\emph{dog}, \emph{toy}, \emph{cat}, \emph{backpack}, \emph{plushie}), with the y‐axis denoting the metric value and the x‐axis representing the total number of tasks learned. Error bars denote one standard deviation across ten runs.
A steep decline in IA or TA on early tasks indicates catastrophic forgetting, whereas stable or increasing scores imply strong knowledge retention.

\subsection{Metric Tables}

Tables~\ref{tab:final_IA}--\ref{tab:final_Diversity} summarize the final (post‐Task 5) results for IA, TA, and Diversity. Values represent mean and standard deviation over ten runs, allowing for comparison of end‐stage retention.

\begin{table}[h!]
\centering
\caption{Final IA scores (\%) after learning all 5 tasks. Mean $\pm$ std across 10 runs.}
\label{tab:final_IA}
\begin{tabular}{lccccc}
\textbf{Method} & \textbf{Dog} & \textbf{Toy} & \textbf{Cat} & \textbf{Backpack} & \textbf{Plushie} \\
\hline
Naive            & 63.56 $\pm$ 2.43 & 65.25 $\pm$ 3.28 & 61.77 $\pm$ 2.41 & 61.23 $\pm$ 2.61 & 76.95 $\pm$ 3.96 \\
Experience Replay & 65.64 $\pm$ 1.91 & 67.05 $\pm$ 1.68 & 64.19 $\pm$ 2.36 & 64.98 $\pm$ 6.86 & 78.89 $\pm$ 2.25 \\
Latent Replay     & 77.59 $\pm$ 5.30 & 70.43 $\pm$ 4.36 & 76.94 $\pm$ 2.81 & 79.96 $\pm$ 2.61 & 77.72 $\pm$ 3.24 \\
Similarity LR     & 64.62 $\pm$ 1.48 & 65.61 $\pm$ 1.64 & 63.39 $\pm$ 2.24 & 63.73 $\pm$ 1.76 & 79.61 $\pm$ 2.65 \\
Offline           & 84.25 $\pm$ 5.18 & 76.06 $\pm$ 2.35 & 79.99 $\pm$ 6.60 & 81.32 $\pm$ 1.49 & 75.40 $\pm$ 3.37 \\
\end{tabular}
\end{table}

\begin{table}[h!]
\centering
\caption{Final TA scores (\%) after learning all 5 tasks. Mean $\pm$ std across 10 runs.}
\label{tab:final_TA}
\begin{tabular}{lccccc}
\textbf{Method} & \textbf{Dog} & \textbf{Toy} & \textbf{Cat} & \textbf{Backpack} & \textbf{Plushie} \\
\hline
Naive            & 15.64 $\pm$ 0.67 & 17.87 $\pm$ 0.70 & 12.46 $\pm$ 0.87 & 15.38 $\pm$ 0.92 & 18.86 $\pm$ 1.37 \\
Experience Replay & 15.81 $\pm$ 1.32 & 18.26 $\pm$ 0.42 & 12.18 $\pm$ 0.90 & 16.33 $\pm$ 3.29 & 19.01 $\pm$ 1.05 \\
Latent Replay     & 19.33 $\pm$ 1.78 & 18.81 $\pm$ 1.41 & 20.26 $\pm$ 1.79 & 24.05 $\pm$ 1.01 & 18.34 $\pm$ 0.98 \\
Similarity LR     & 15.64 $\pm$ 0.46 & 18.58 $\pm$ 0.35 & 11.85 $\pm$ 0.56 & 15.87 $\pm$ 0.70 & 19.03 $\pm$ 0.45 \\
Offline           & 21.12 $\pm$ 0.88 & 17.00 $\pm$ 1.15 & 21.20 $\pm$ 1.83 & 23.87 $\pm$ 0.82 & 18.77 $\pm$ 0.94 \\
\end{tabular}
\end{table}

\begin{table}[h!]
\centering
\caption{Final Diversity scores after learning all 5 tasks. Mean $\pm$ std across 10 runs.}
\label{tab:final_Diversity}
\begin{tabular}{lccccc}
\textbf{Method} & \textbf{Dog} & \textbf{Toy} & \textbf{Cat} & \textbf{Backpack} & \textbf{Plushie} \\
\hline
Naive            & 1.13 $\pm$ 0.08 & 1.17 $\pm$ 0.09 & 1.12 $\pm$ 0.05 & 1.14 $\pm$ 0.05 & 1.15 $\pm$ 0.10 \\
Experience Replay & 1.17 $\pm$ 0.12 & 1.12 $\pm$ 0.06 & 1.20 $\pm$ 0.14 & 1.12 $\pm$ 0.05 & 1.11 $\pm$ 0.04 \\
Latent Replay     & 1.66 $\pm$ 0.21 & 1.36 $\pm$ 0.31 & 1.29 $\pm$ 0.10 & 1.23 $\pm$ 0.19 & 1.13 $\pm$ 0.08 \\
Similarity LR     & 1.10 $\pm$ 0.03 & 1.10 $\pm$ 0.03 & 1.10 $\pm$ 0.04 & 1.09 $\pm$ 0.02 & 1.09 $\pm$ 0.03 \\
Offline           & 1.68 $\pm$ 0.20 & 1.27 $\pm$ 0.35 & 1.53 $\pm$ 0.20 & 1.14 $\pm$ 0.16 & 1.25 $\pm$ 0.21 \\
\end{tabular}
\end{table}

\begin{table}[h!]
\centering
\caption{Task Forgetting Rate (TFR) after learning all tasks. Lower values indicate better retention of previous knowledge.}
\label{tab:TFR}
\begin{tabular}{lcc}
\textbf{Method} & \textbf{TFR-IA} & \textbf{TFR-TA} \\
\hline
Naive            & 16.16 $\pm$ 2.71 & 4.39 $\pm$ 0.95 \\
Experience Replay & 13.59 $\pm$ 2.28 & 4.05 $\pm$ 1.15 \\
Latent Replay     &  4.84 $\pm$ 1.72 & 0.84 $\pm$ 0.46 \\
Similarity LR     & 18.48 $\pm$ 1.56 & 5.37 $\pm$ 0.67 \\
Offline           &  0.34 $\pm$ 0.89 & 0.39 $\pm$ 0.32 \\
\end{tabular}
\end{table}

\subsection{IA and TA Observations}

IA measures how closely model-generated images resemble their corresponding training images, while TA evaluates the accuracy of textual descriptions in conditioning the generation process.

From Table~\ref{tab:final_IA} and Figure~\ref{fig:combinedplots}, \textbf{Offline} yields the highest IA on most tasks, as expected, since it trains on all data from all tasks simultaneously. \textbf{LR} notably outperforms both \textbf{Naive} and \textbf{ER} on final IA for the earliest task (\emph{dog}), indicating that LR's 480-latent buffer more effectively preserves that concept. LR's IA also remains comparatively high on \emph{cat} and \emph{backpack} after \emph{plushie} is introduced, suggesting stronger resilience to forgetting. 

\textbf{Naive} exhibits the largest drop in IA from Task~1 to Task~5 --- dog’s final IA is around 63.56\%, well below LR’s 77.59\%. ER provides some improvement over Naive but still remains behind LR on tasks like \emph{dog}, \emph{cat}, and \emph{toy}, implying that storing only ten raw images for previous tasks is insufficient to fully preserve earlier knowledge.

Regarding TA, shown in Table~\ref{tab:final_TA} and Figure~\ref{fig:combinedplots}, \textbf{Offline} generally achieves the highest scores, though LR achieves respectable TA gains relative to Naive or ER for early tasks. Notably, LR displays a significant jump on \emph{backpack} (24.05\% TA) vs.\ ER (16.33\%) and Naive (15.38\%), suggesting that replaying older latents need not interfere with new textual details.

Table~\ref{tab:TFR} further quantifies forgetting via TFR-IA and TFR-TA. \textbf{Offline} approaches near‐zero forgetting (TFR-IA=0.34, TFR-TA=0.39), thanks to simultaneous access to all task data. LR shows notably low forgetting relative to Naive or ER, underscoring the benefit of a 480-latent buffer. Naive and SLR, by contrast, exhibit more pronounced forgetting. Interestingly, SLR's unexpected underperformance (TFR-IA=18.48\%, even worse than Naive at 16.16\%) indicates that similarity-based selection does not necessarily outperform random latent sampling in this setting. We explore the mechanisms behind this counterintuitive finding in 
Section \ref{sec:ablation_detailed}.

\subsection{Diversity Observations}

Diversity measures the extent to which the generated images vary within a given concept, ensuring that the model does not suffer from mode collapse, where it repeatedly generates nearly identical images instead of maintaining a broad distribution.

Turning to Diversity in Figure~\ref{fig:combinedplots} and Table~\ref{tab:final_Diversity}, \textbf{Offline} and \textbf{LR} maintain stronger diversity than \textbf{Naive} or \textbf{ER}, especially for earlier learned concepts. After learning all five tasks, LR achieves a diversity score of 1.66 for the earliest task (\emph{dog}), comparable to Offline's 1.68, and substantially higher than Naive (1.13) and ER (1.17). LR's diversity advantage gradually diminishes for more recently learned concepts, with scores decreasing from 1.66 (\emph{dog}) to 1.13 (\emph{plushie}). This pattern suggests that LR's latent representation more effectively preserves the distribution of earlier concepts, while all methods achieve similar diversity for the most recently learned task. ER shows moderate diversity but remains consistently below LR across most tasks, indicating that LR's 480-latent buffer better captures the concepts' manifold than ER's 10-image buffer, more effectively mitigating mode collapse for earlier tasks.

\section{Statistical Significance Analysis}
\label{sec:statistical_analysis}

Table~\ref{tab:statistical_tests} summarizes the significant findings from our analysis, focusing on the critical comparison between LR and ER, as well as between LR and Naive fine-tuning.

\begin{table}[h!]
\centering
\caption{Summary of statistically significant differences between methods ($p < 0.05$ after correction).}
\label{tab:statistical_tests}
\begin{tabular}{lll}
\hline
\textbf{Comparison} & \textbf{Metric} & \textbf{Significant Differences} \\
\hline
LR vs. ER & Image Alignment & Dog ($p = 0.002$), Cat ($p = 0.003$), \\
 &  & Backpack ($p = 0.004$) \\
\cline{2-3}
 & Text Alignment & Dog ($p = 0.012$), Cat ($p = 0.005$), \\
 &  & Backpack ($p = 0.007$) \\
\cline{2-3}
 & Diversity & Dog ($p = 0.003$), Toy ($p = 0.031$) \\
\hline
LR vs. Naive & Image Alignment & Dog ($p = 0.001$), Cat ($p = 0.002$), \\
 &  & Backpack ($p = 0.004$), Toy ($p = 0.022$) \\
\cline{2-3}
 & Text Alignment & Dog ($p = 0.007$), Cat ($p = 0.003$), \\
 &  & Backpack ($p = 0.005$) \\
\cline{2-3}
 & Diversity & Dog ($p = 0.002$), Toy ($p = 0.026$) \\
\hline
LR vs. SLR & Image Alignment & Dog ($p = 0.003$), Cat ($p = 0.005$), \\
 &  & Backpack ($p = 0.004$) \\
\cline{2-3}
 & Diversity & Dog ($p = 0.004$), Cat ($p = 0.029$) \\
\hline
Offline vs. LR & Image Alignment & Dog ($p = 0.028$) \\
\cline{2-3}
 & TFR-IA & Overall ($p = 0.012$) \\
\hline
\end{tabular}
\end{table}

\subsection{Interpretation of Statistical Findings}

These statistical analyses confirm the following findings:

\begin{enumerate}
\item \textbf{LR vs. Naive:} The significant advantages of LR over Naive fine-tuning across nearly all tasks validates that LR effectively mitigates catastrophic forgetting.

\item \textbf{LR vs. SLR:} Standard LR significantly outperforms SLR on IA for early tasks and diversity metrics, suggesting that the similarity selection mechanism may be too restrictive in our experimental setting.

\item \textbf{Offline vs. LR:} While Offline training maintains a statistical edge over LR on the earliest task (Dog) and overall forgetting (TFR-IA), the differences on later tasks are not statistically significant, indicating that LR approaches Offline performance on many dimensions despite using far less memory.

\item \textbf{Diversity:} Statistical significance in diversity metrics, particularly for Dog and Toy tasks, confirms that LR better preserves the generative diversity of earlier concepts compared to both ER and Naive approaches.
\end{enumerate}

For the Plushie task (Task 5), differences between methods were generally not statistically significant, which aligns with our expectation that all methods perform well on the most recently learned task.

These statistical findings strengthen our conclusion that LR offers significant advantages for mitigating catastrophic forgetting and preserving generative diversity in sequential text-to-image diffusion fine-tuning, particularly for earlier tasks in the sequence.

\section{Qualitative Analysis}
\label{sec:qual_analysis}

\begin{figure*}[tp]
    \centering
    \resizebox{\textwidth}{!}{%
    \begin{tabular}{c@{\hspace{0.5em}}c@{\hspace{0.5em}}c@{\hspace{0.5em}}c@{\hspace{0.5em}}c@{\hspace{0.5em}}c}
        & \textbf{Dog} & \textbf{Toy} & \textbf{Cat} & \textbf{Backpack} & \textbf{Plushie} \\
        \rotatebox{90}{\parbox{3cm}{\centering\textbf{Naive}}} &
        \includegraphics[height=0.18\textwidth, keepaspectratio]{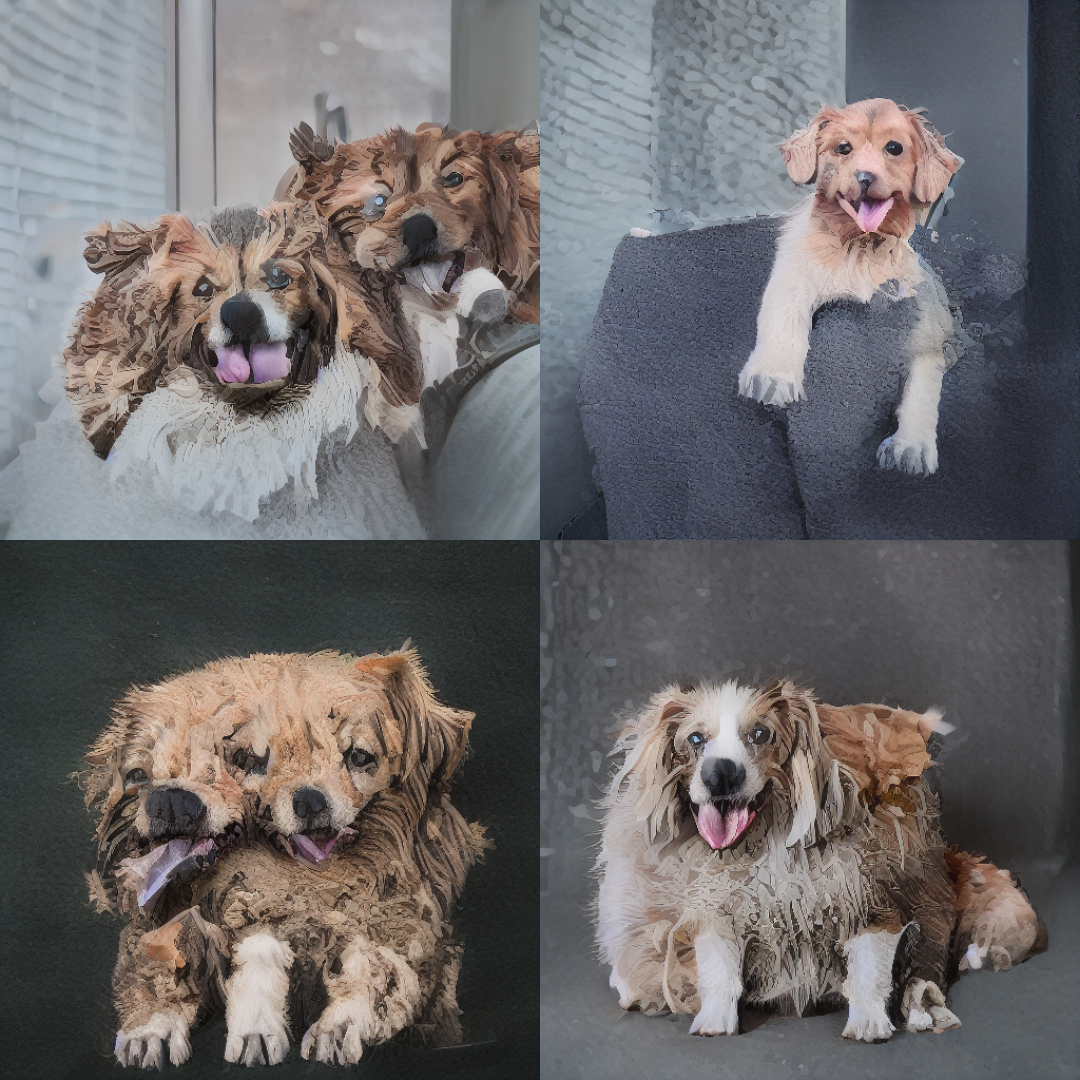} &
        \includegraphics[height=0.18\textwidth, keepaspectratio]{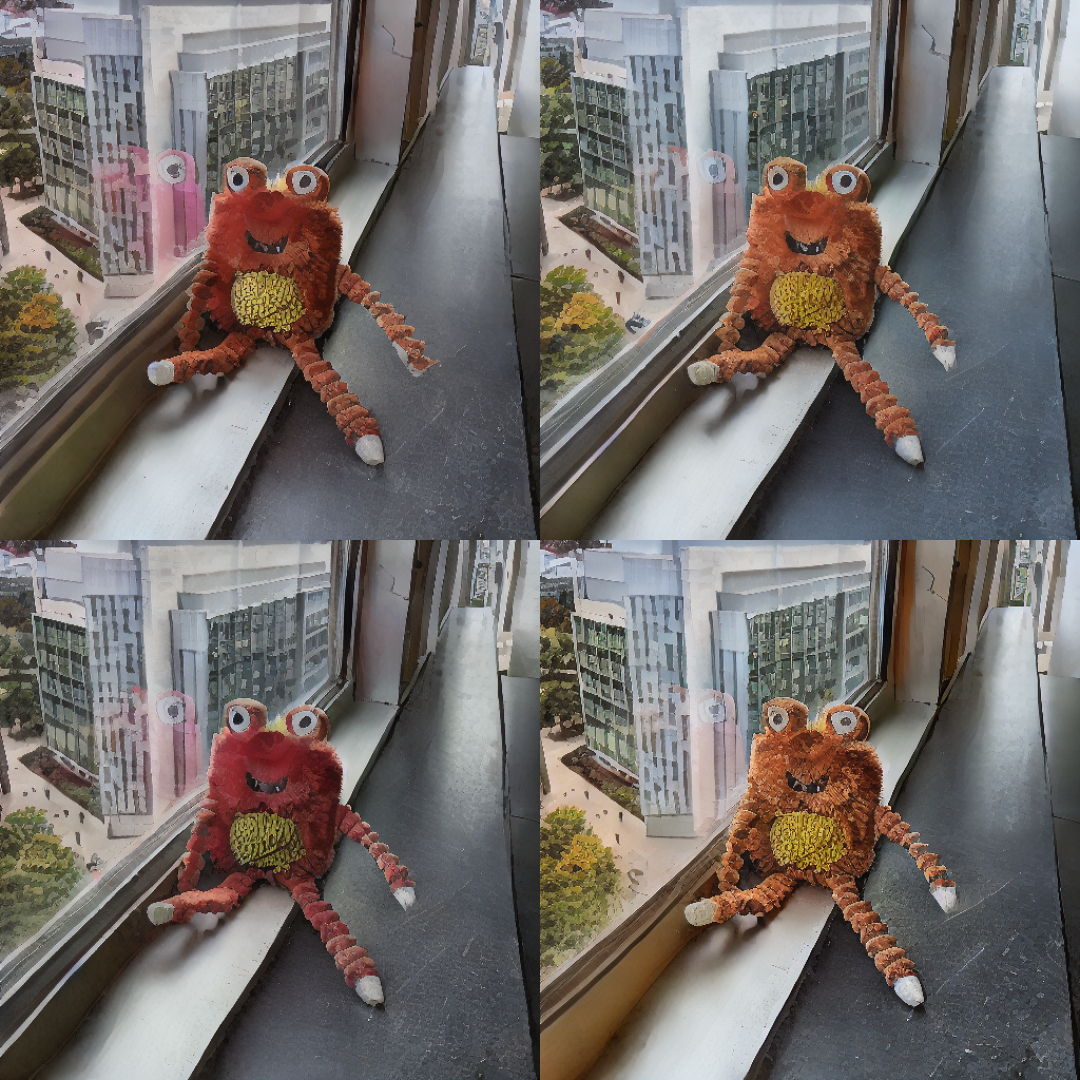} &
        \includegraphics[height=0.18\textwidth, keepaspectratio]{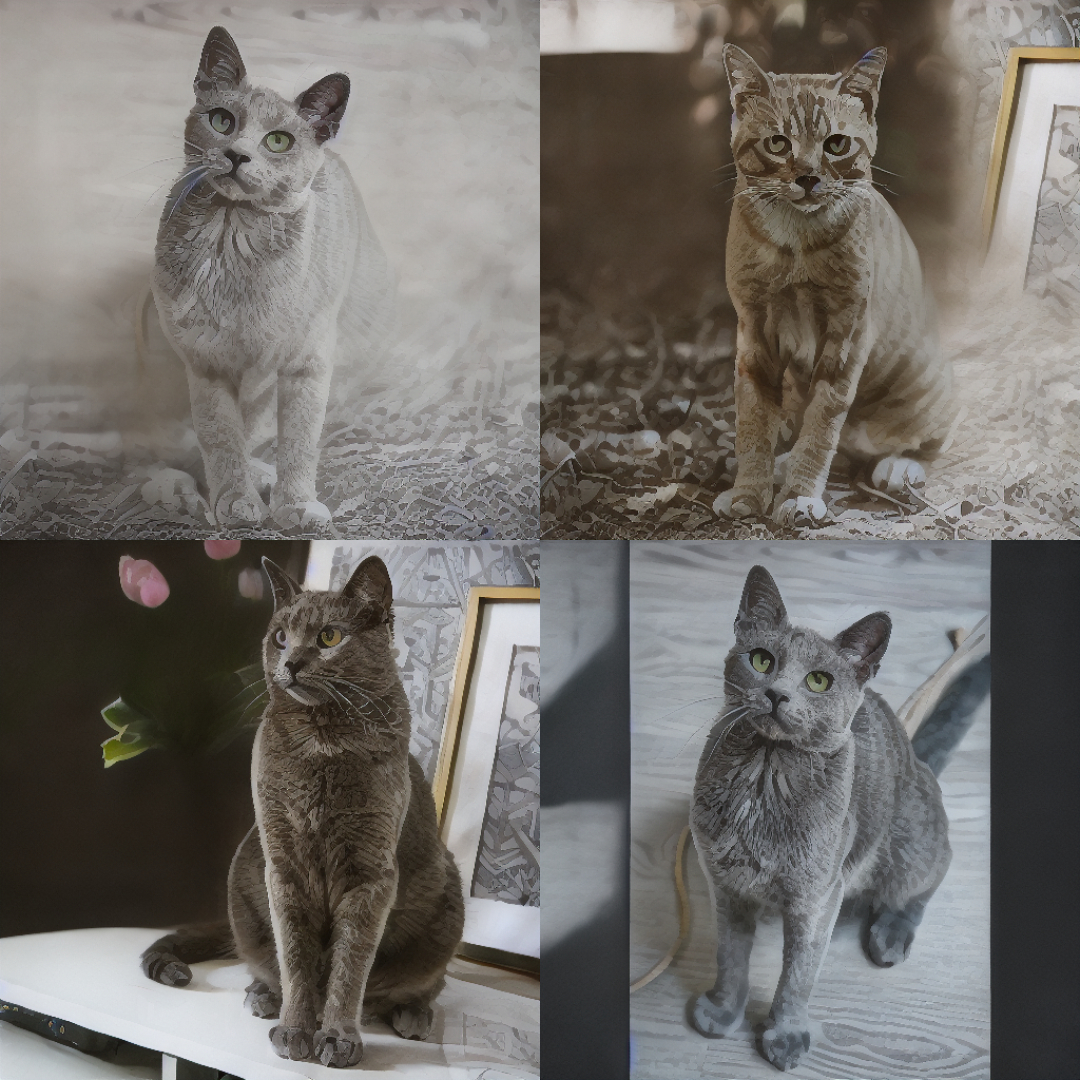} &
        \includegraphics[height=0.18\textwidth, keepaspectratio]{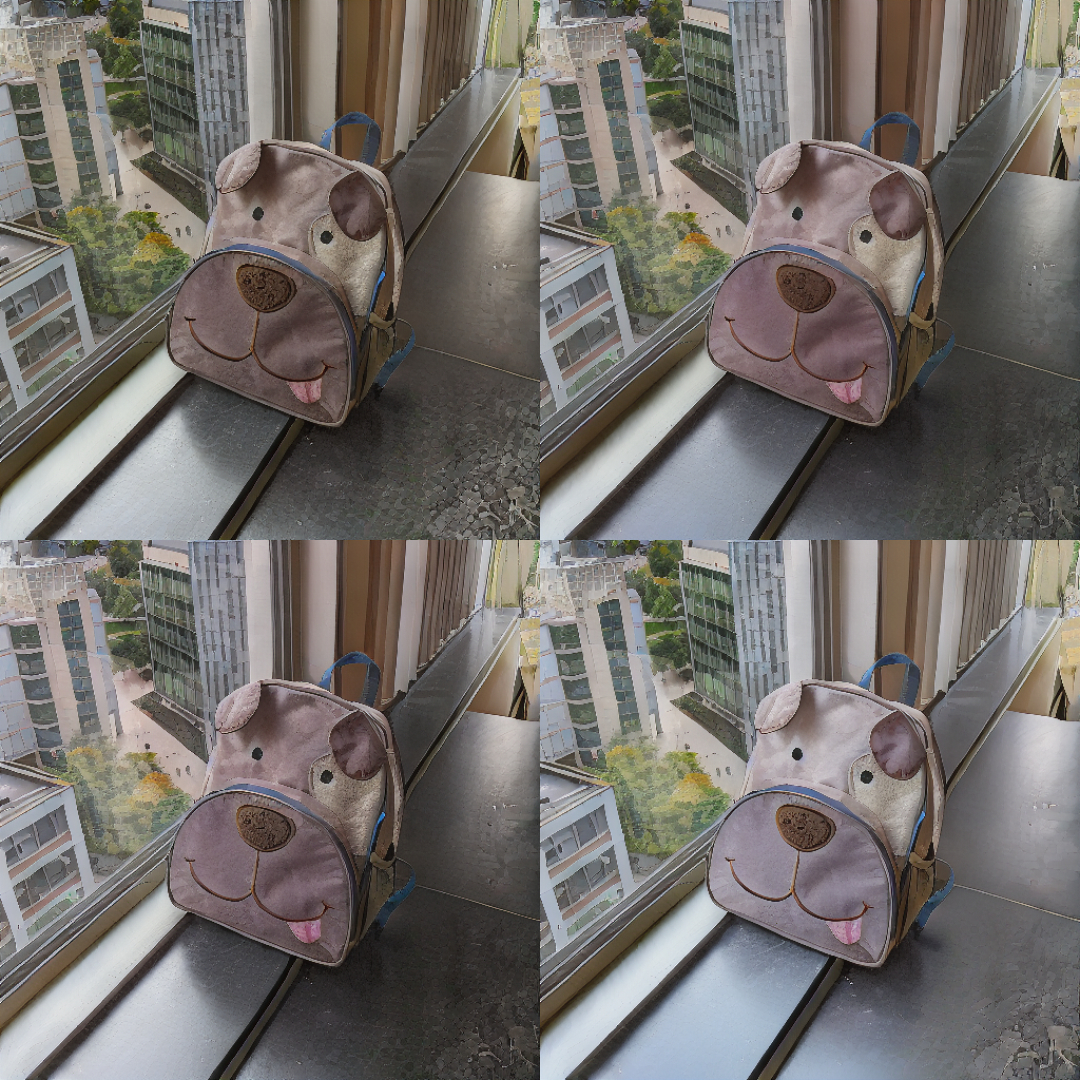} &
        \includegraphics[height=0.18\textwidth, keepaspectratio]{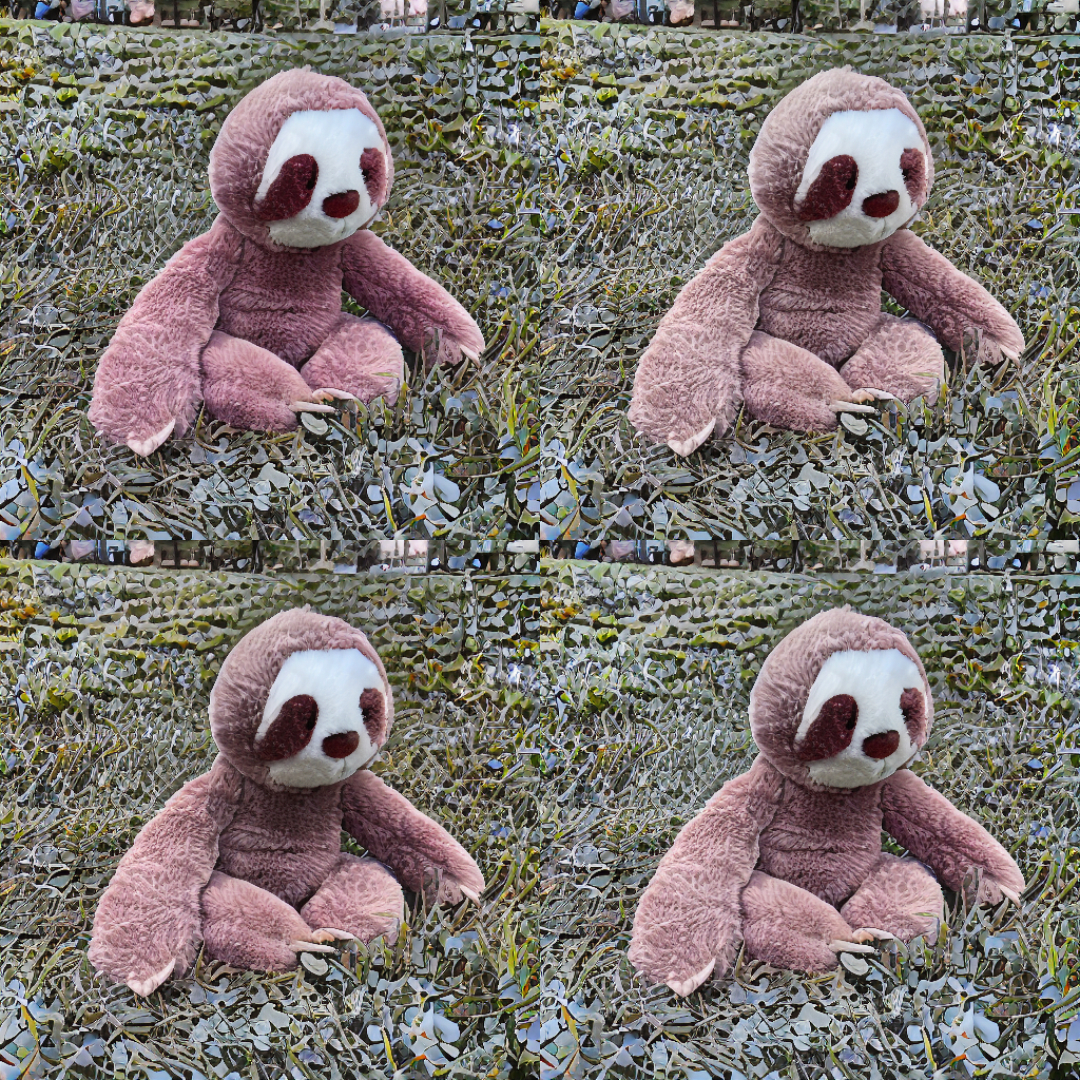} \\
        \rotatebox{90}{\parbox{3cm}{\centering\textbf{ER}}} &
        \includegraphics[height=0.18\textwidth, keepaspectratio]{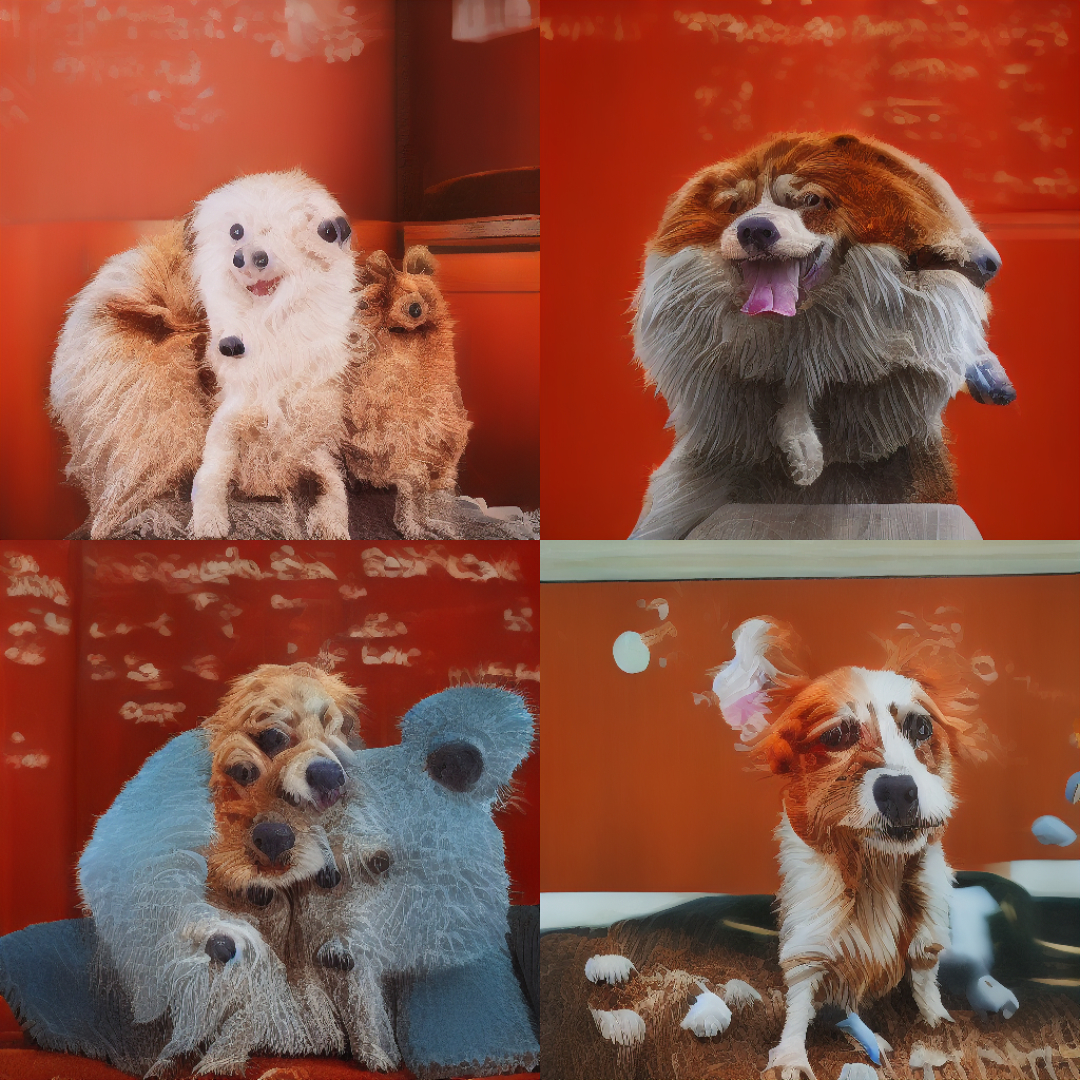} &
        \includegraphics[height=0.18\textwidth, keepaspectratio]{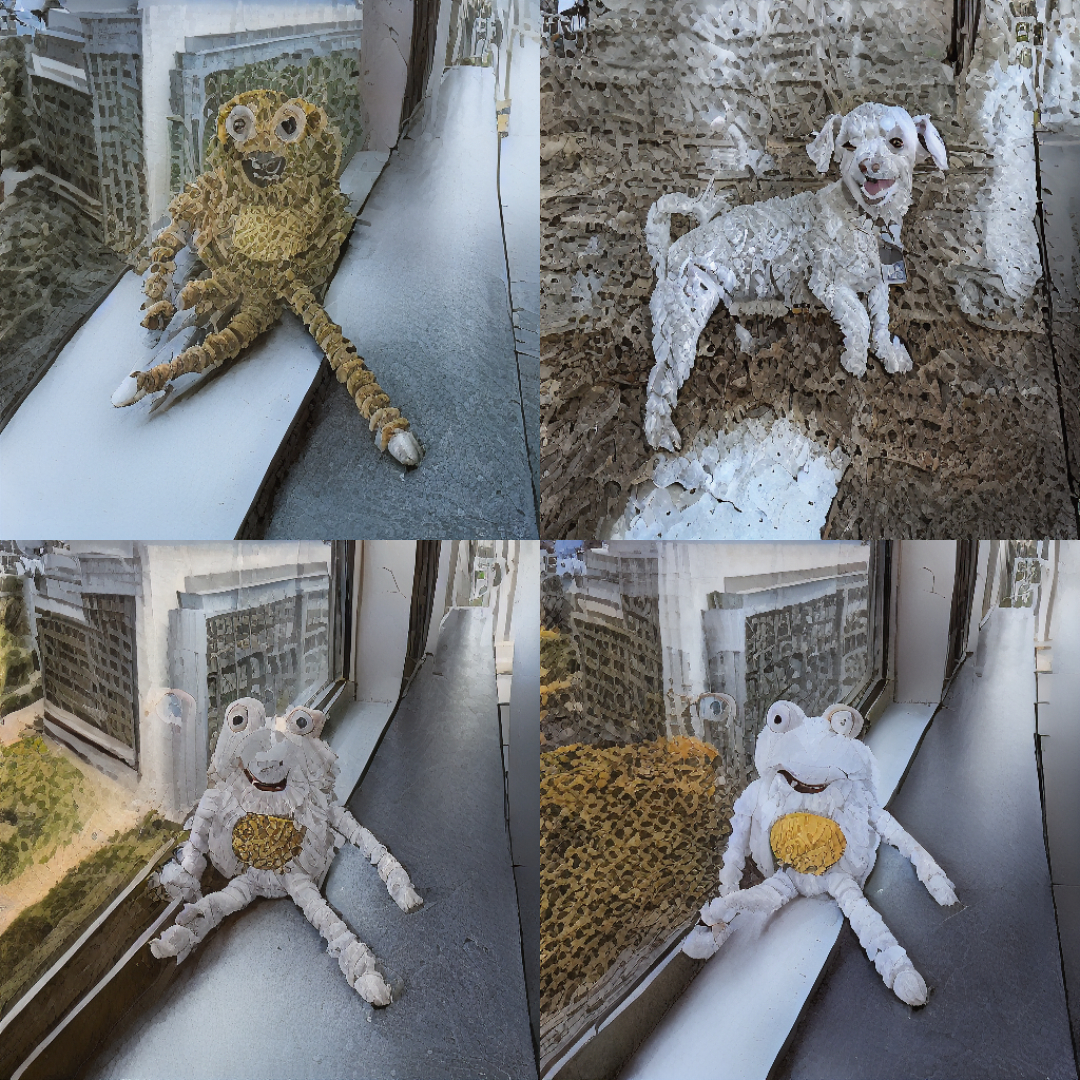} &
        \includegraphics[height=0.18\textwidth, keepaspectratio]{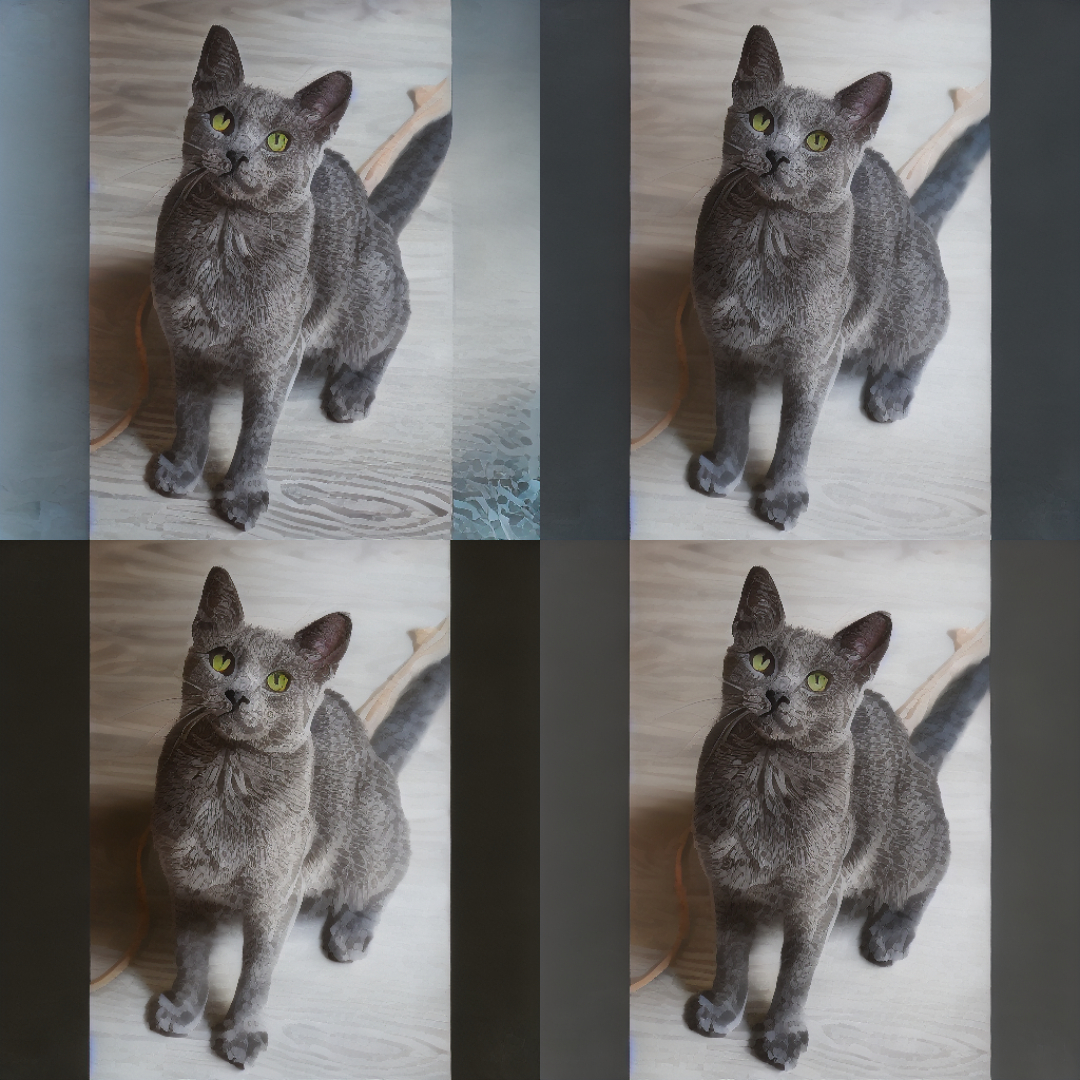} &
        \includegraphics[height=0.18\textwidth, keepaspectratio]{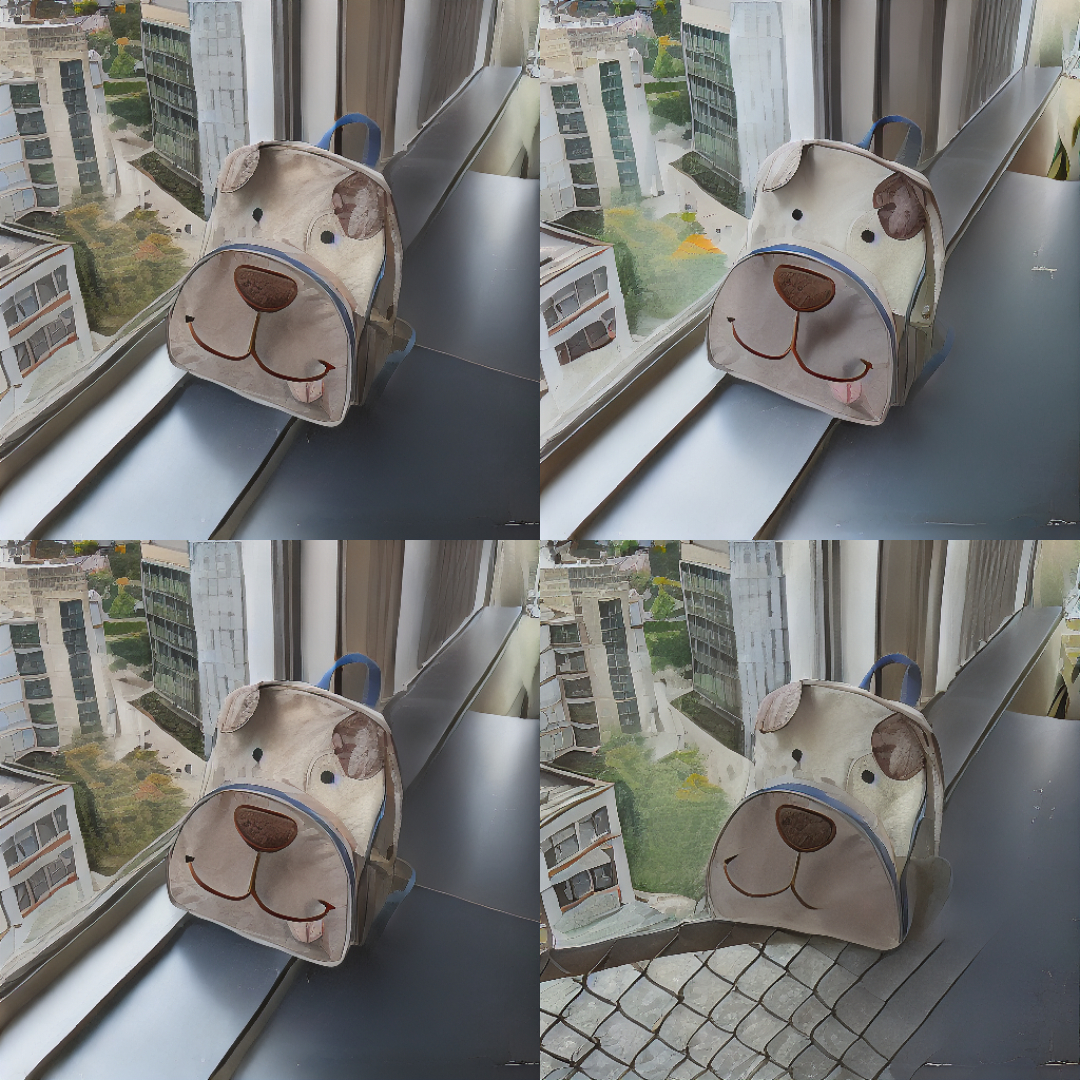} &
        \includegraphics[height=0.18\textwidth, keepaspectratio]{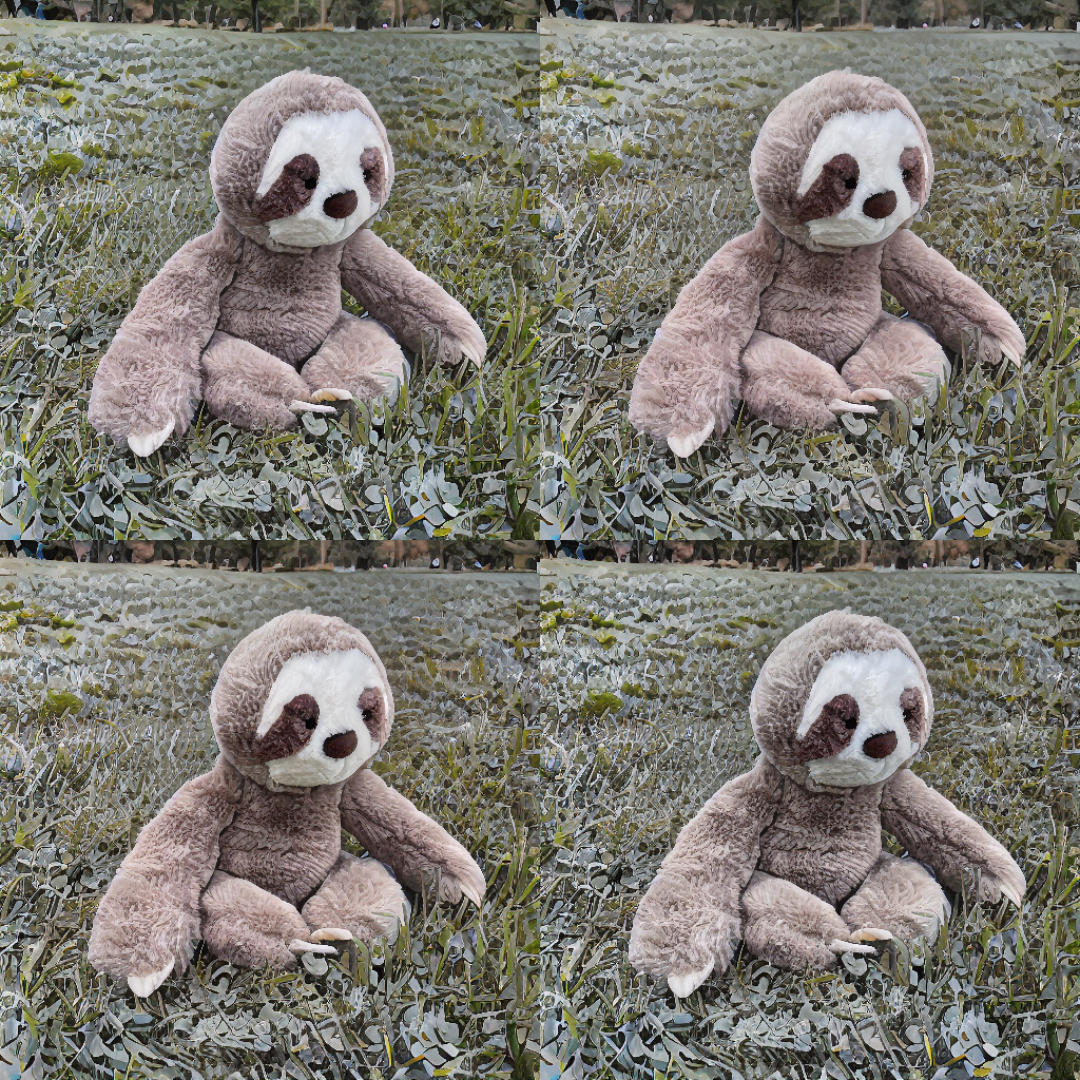} \\
        \rotatebox{90}{\parbox{3cm}{\centering\textbf{LR}}} &
        \includegraphics[height=0.18\textwidth, keepaspectratio]{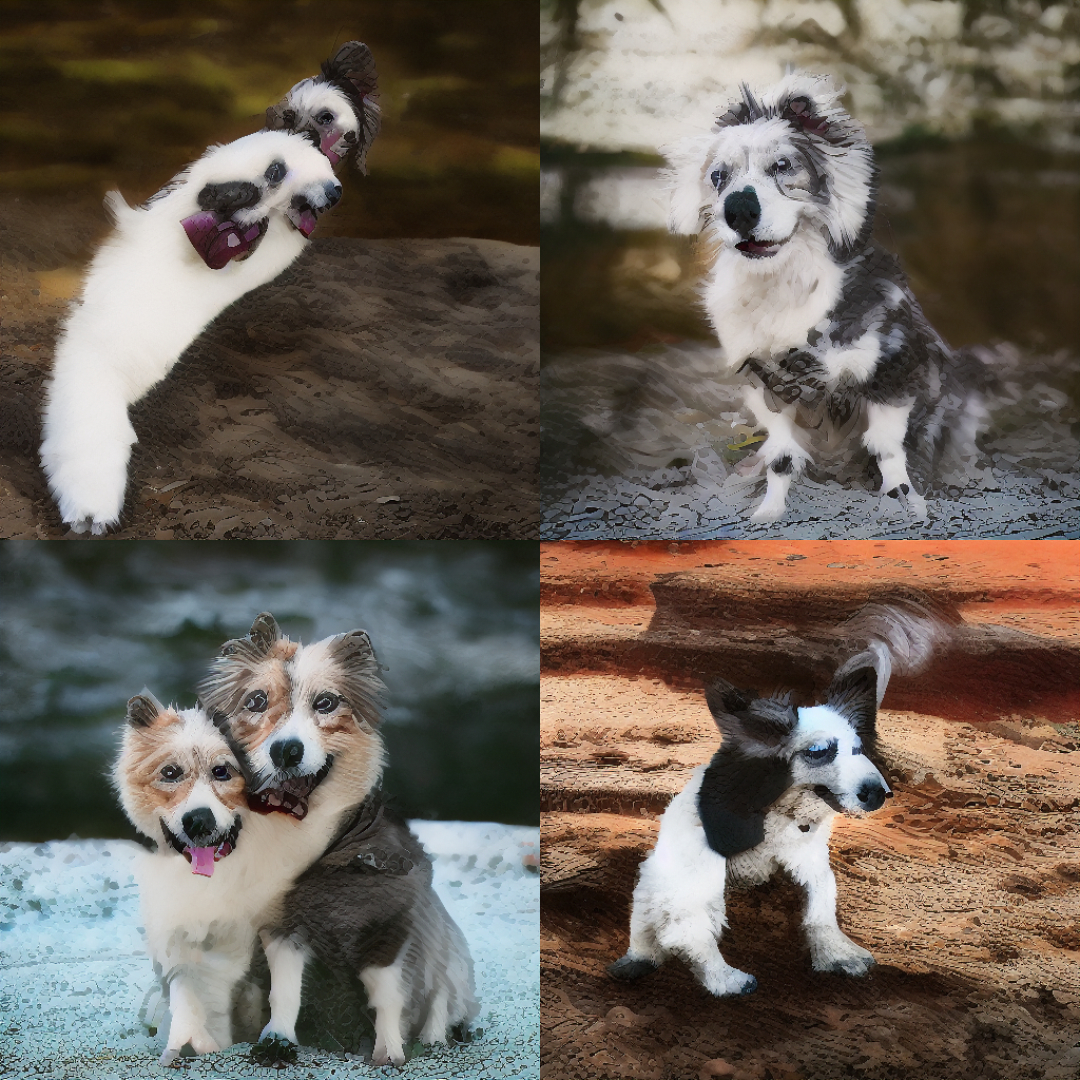} &
        \includegraphics[height=0.18\textwidth, keepaspectratio]{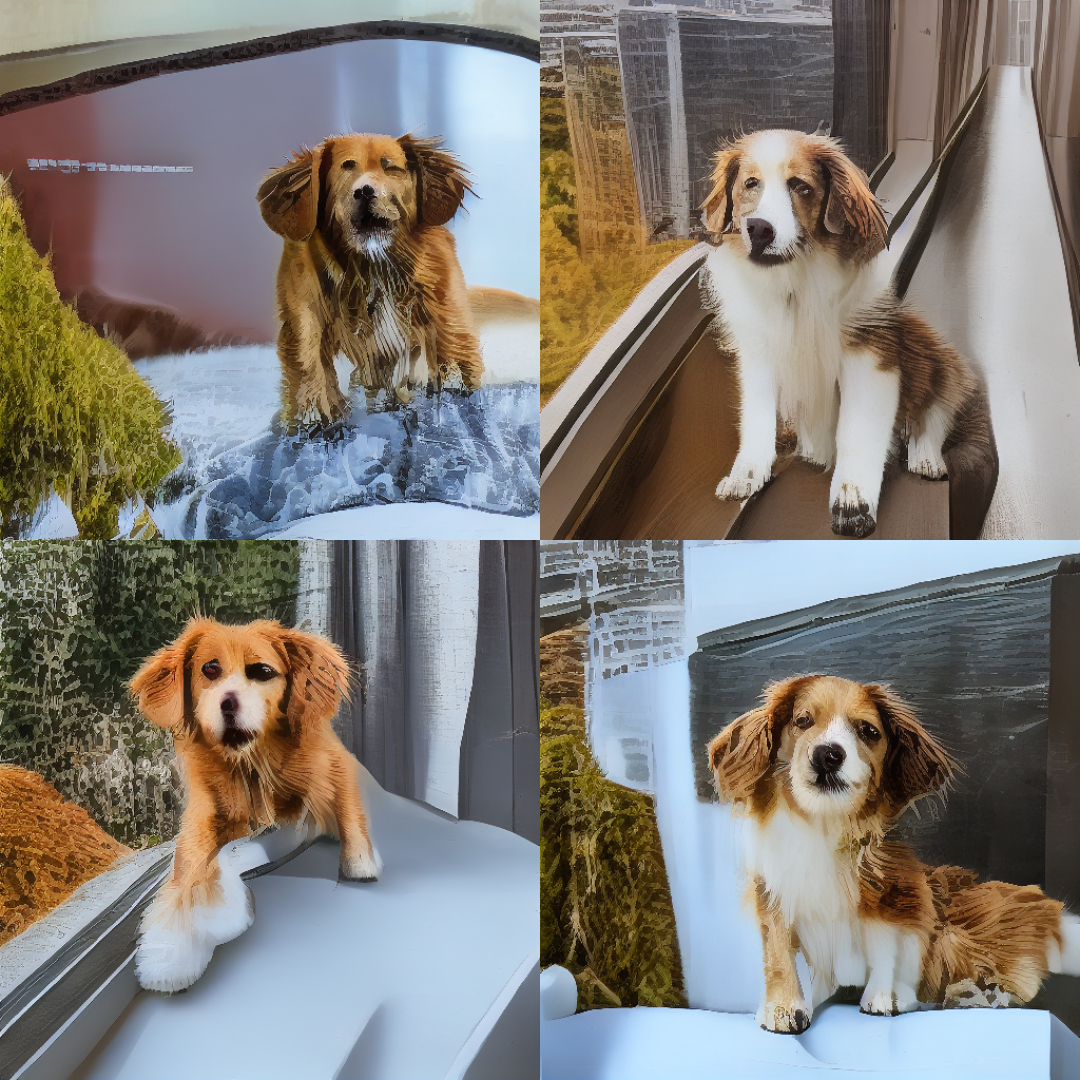} &
        \includegraphics[height=0.18\textwidth, keepaspectratio]{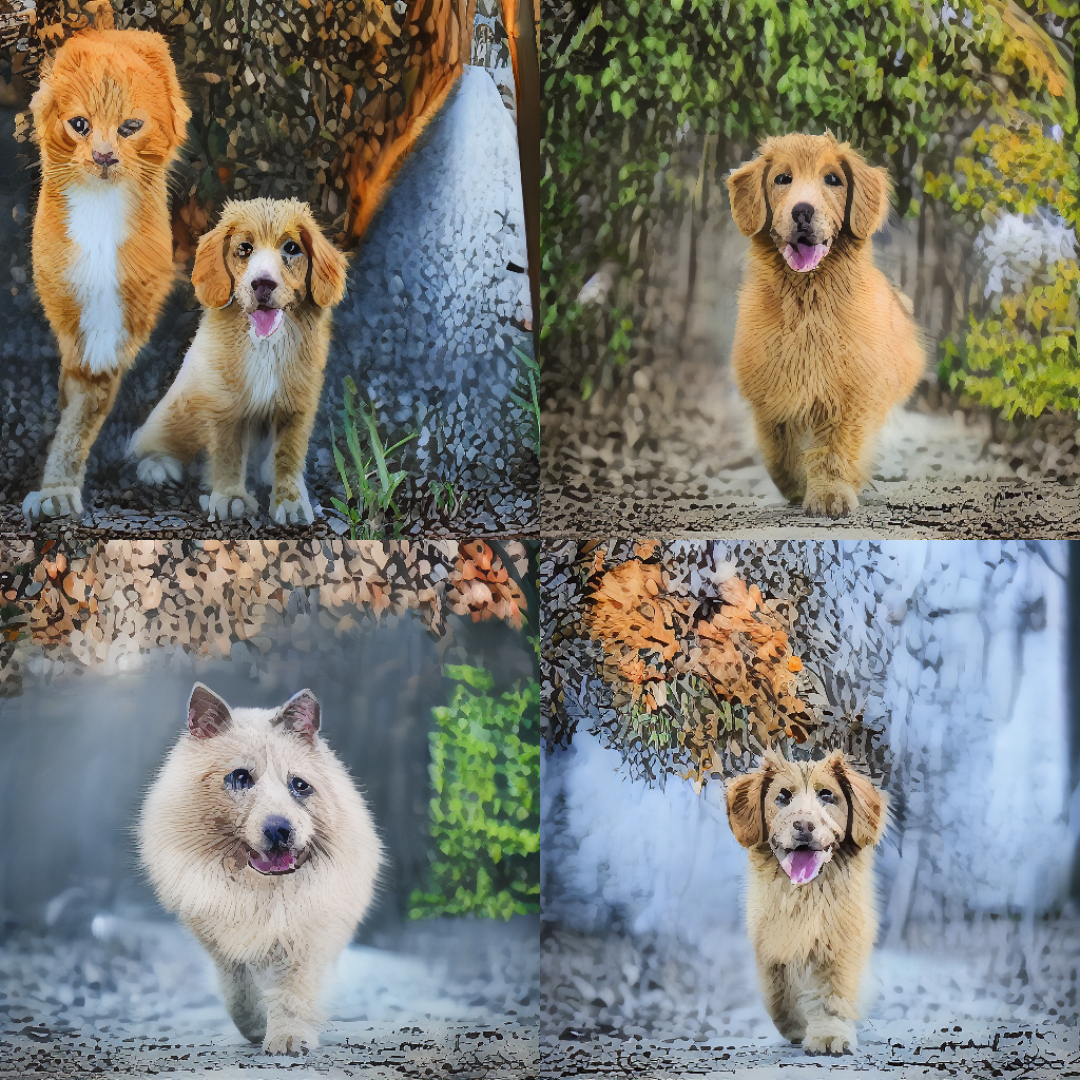} &
        \includegraphics[height=0.18\textwidth, keepaspectratio]{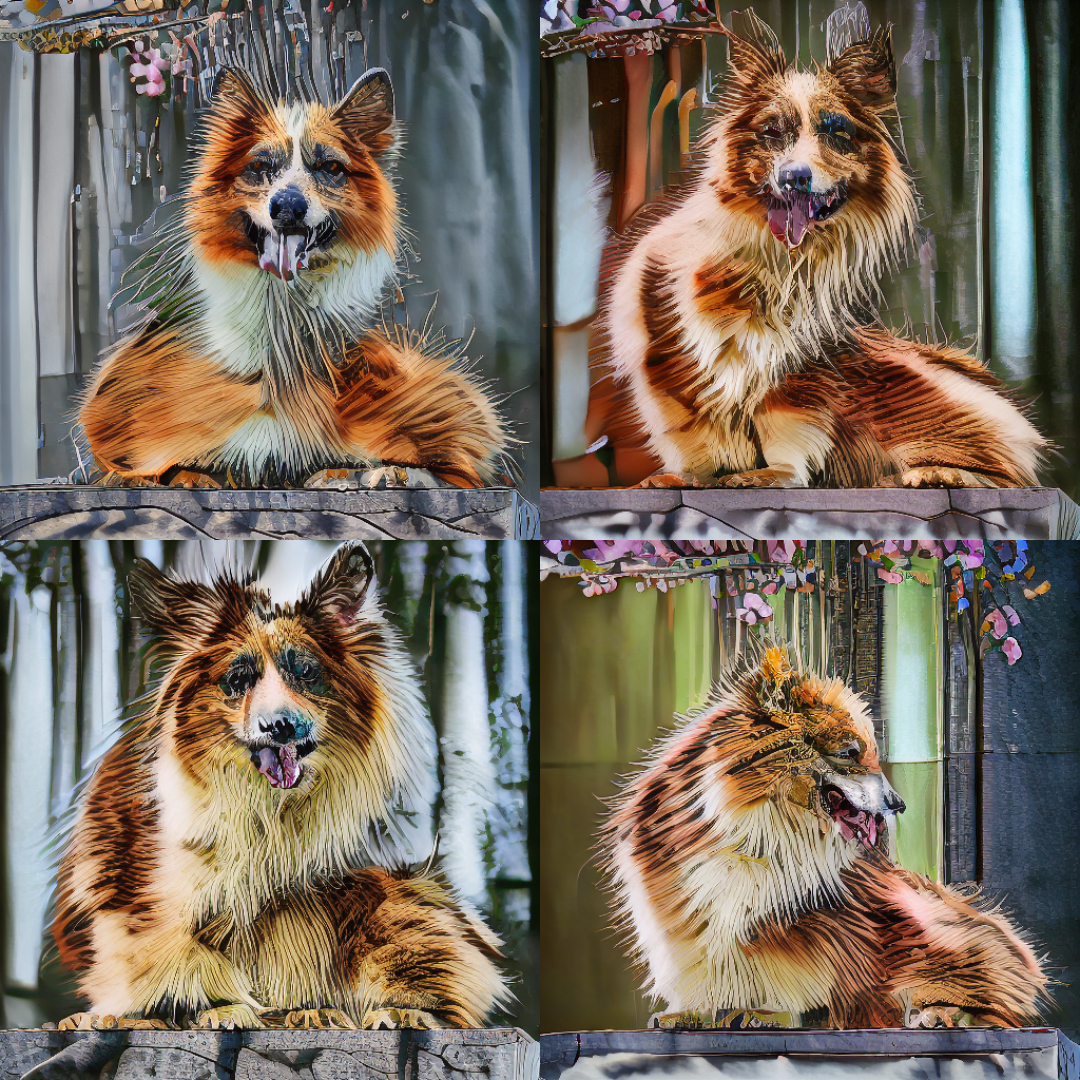} &
        \includegraphics[height=0.18\textwidth, keepaspectratio]{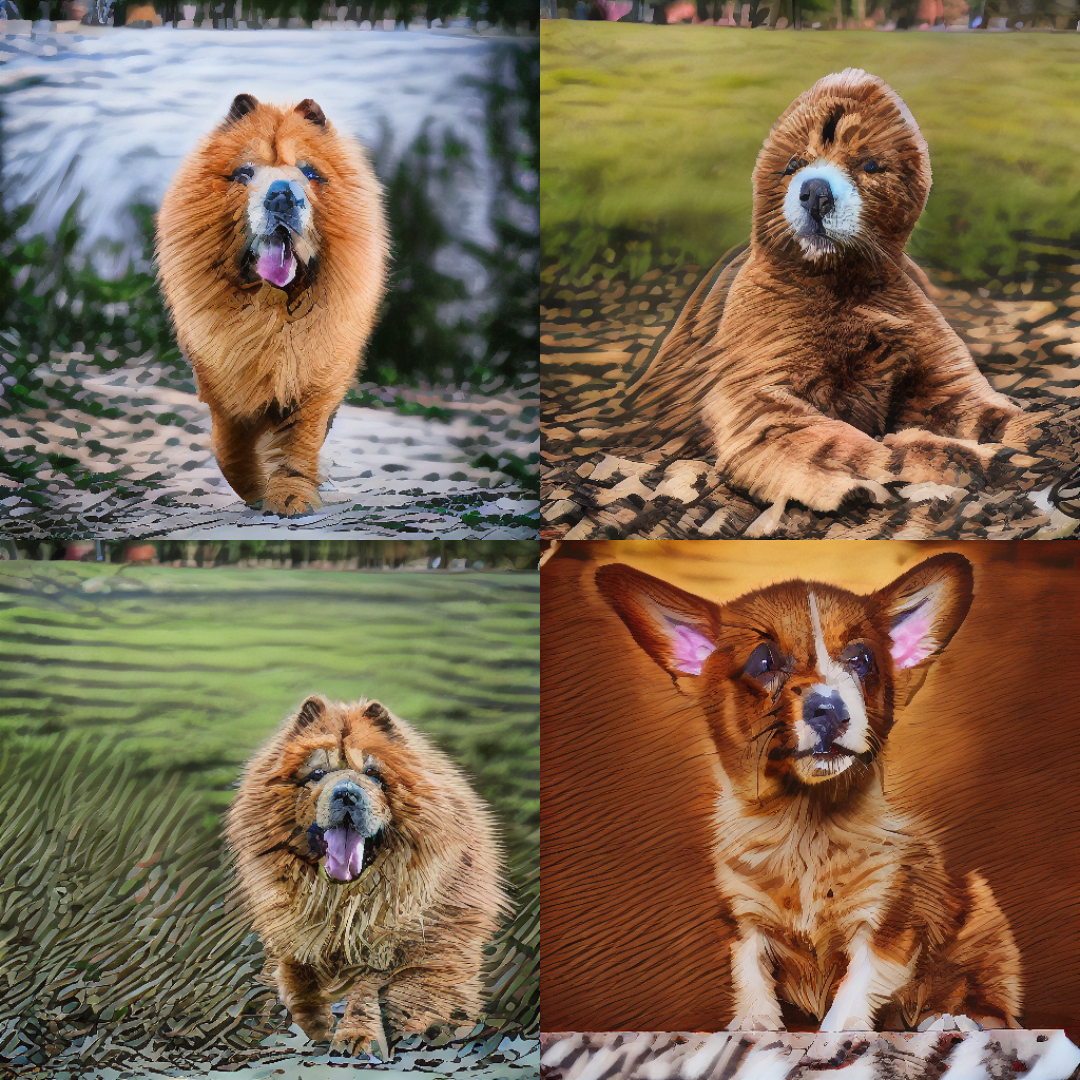} \\
        \rotatebox{90}{\parbox{3cm}{\centering\textbf{SLR}}} &
        \includegraphics[height=0.18\textwidth, keepaspectratio]{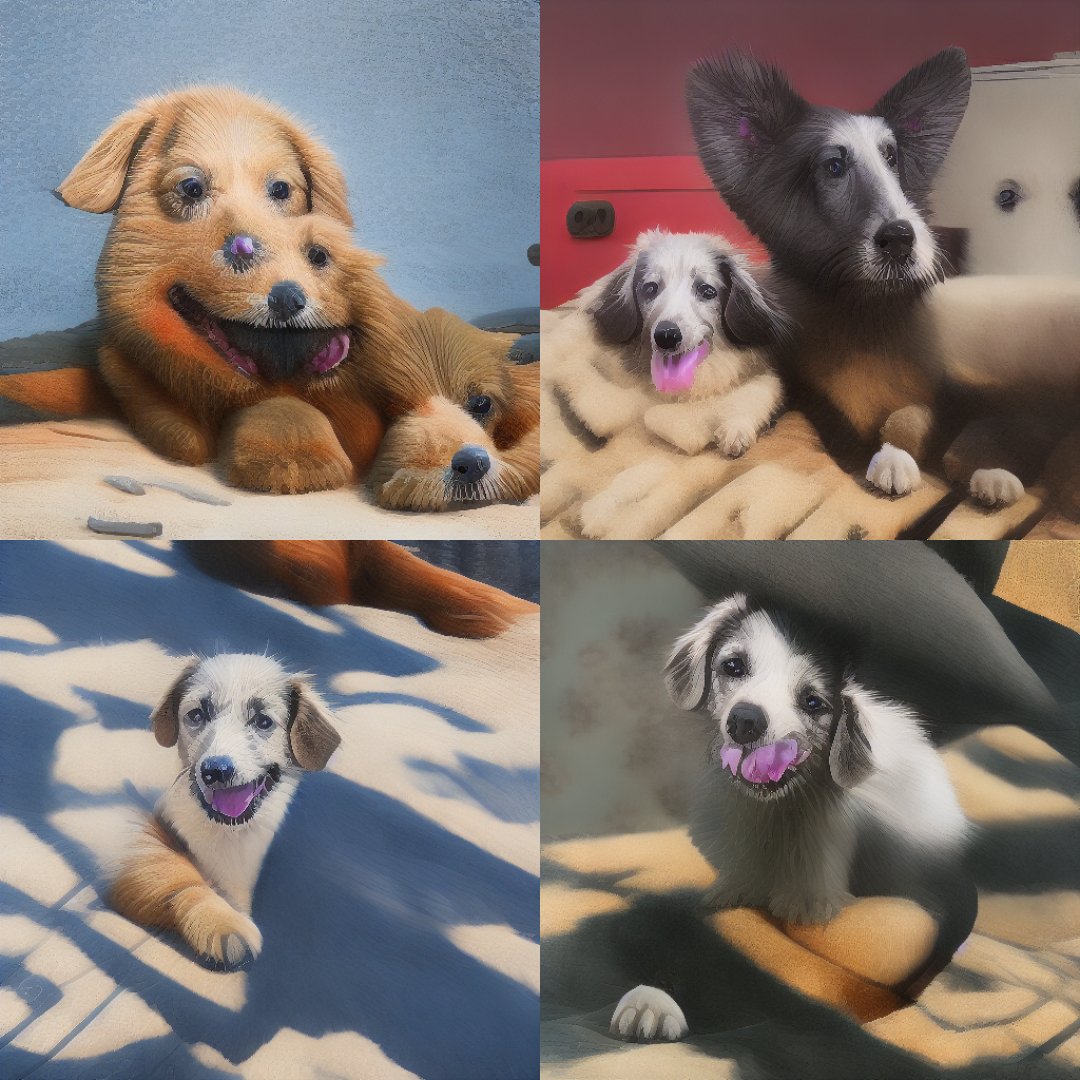} &
        \includegraphics[height=0.18\textwidth, keepaspectratio]{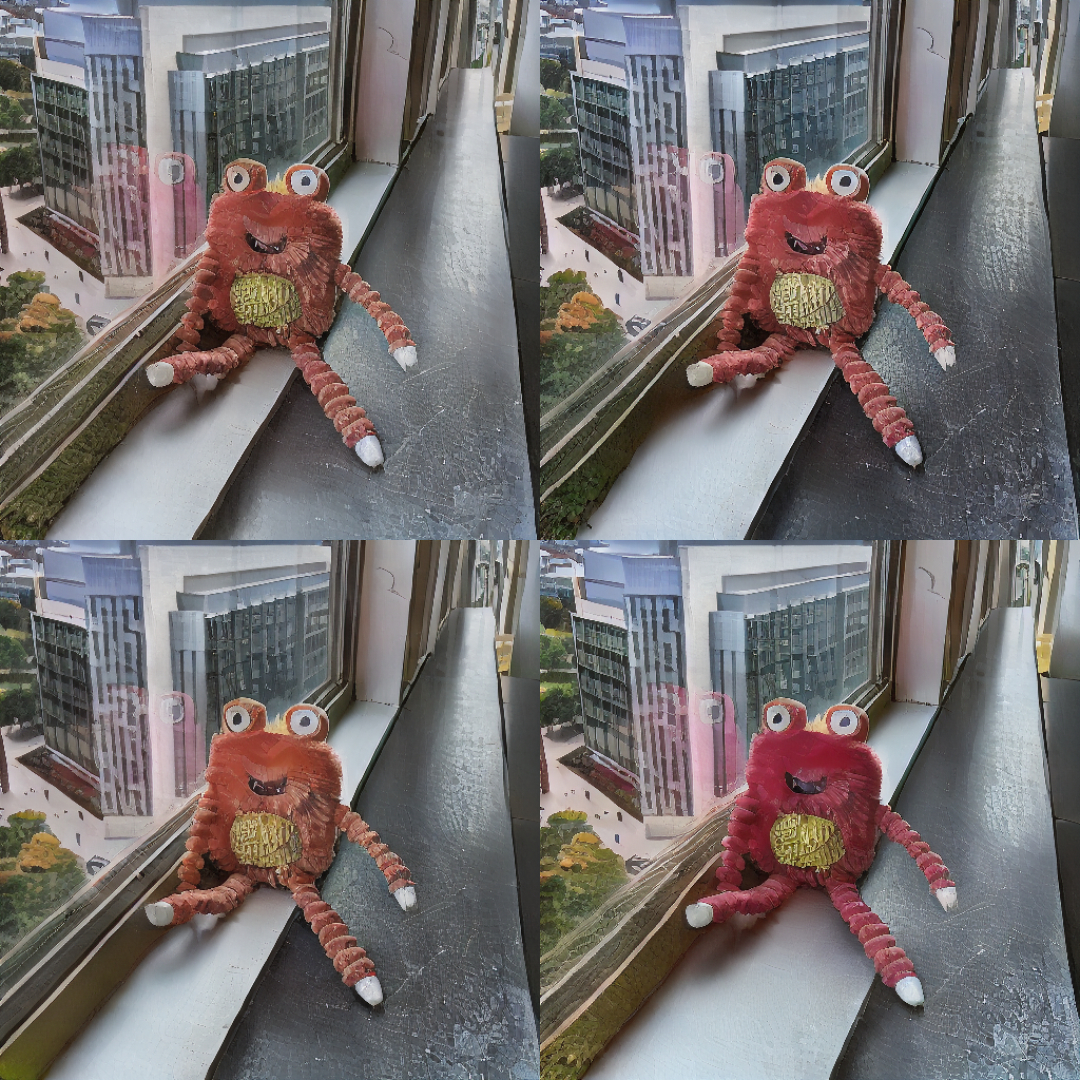} &
        \includegraphics[height=0.18\textwidth, keepaspectratio]{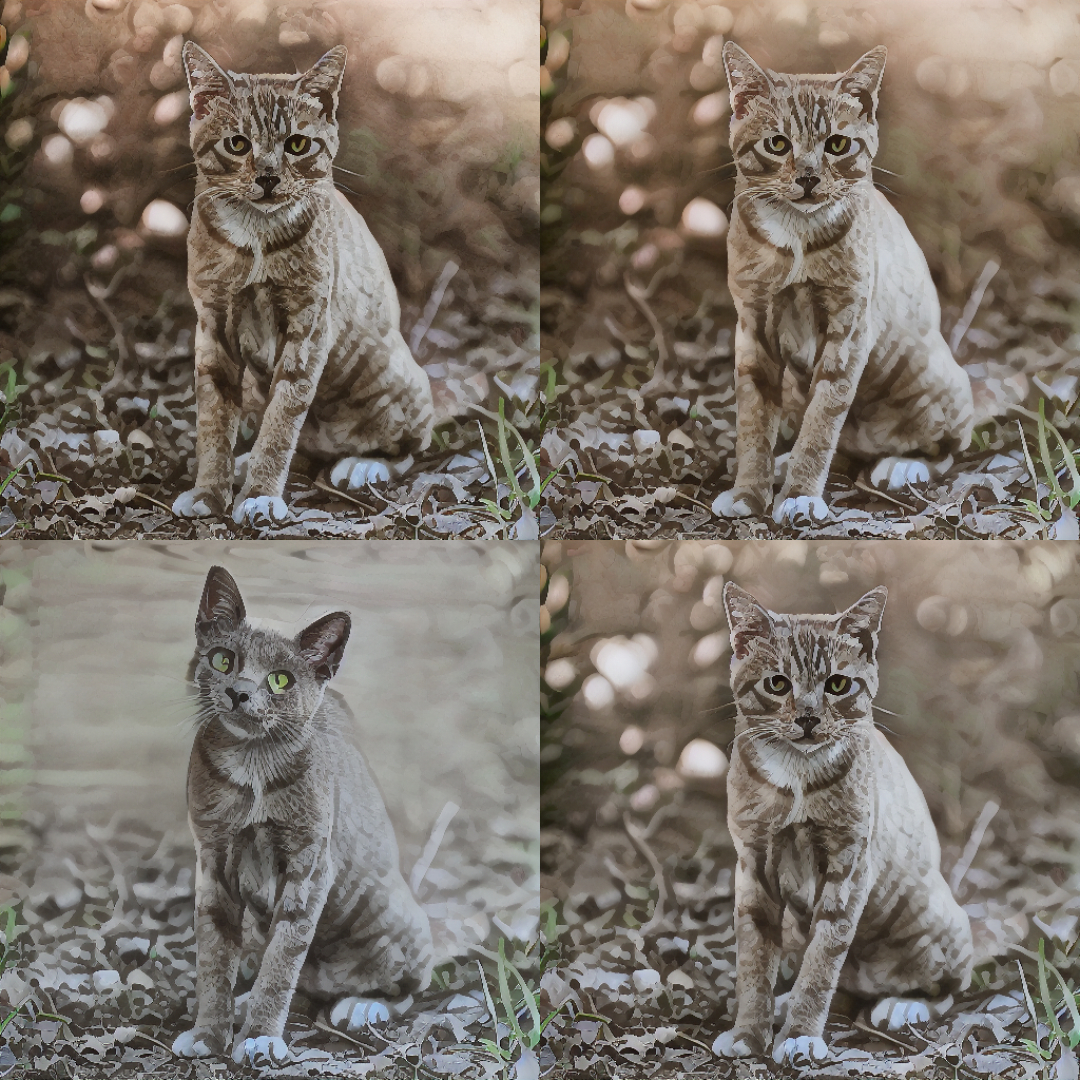} &
        \includegraphics[height=0.18\textwidth, keepaspectratio]{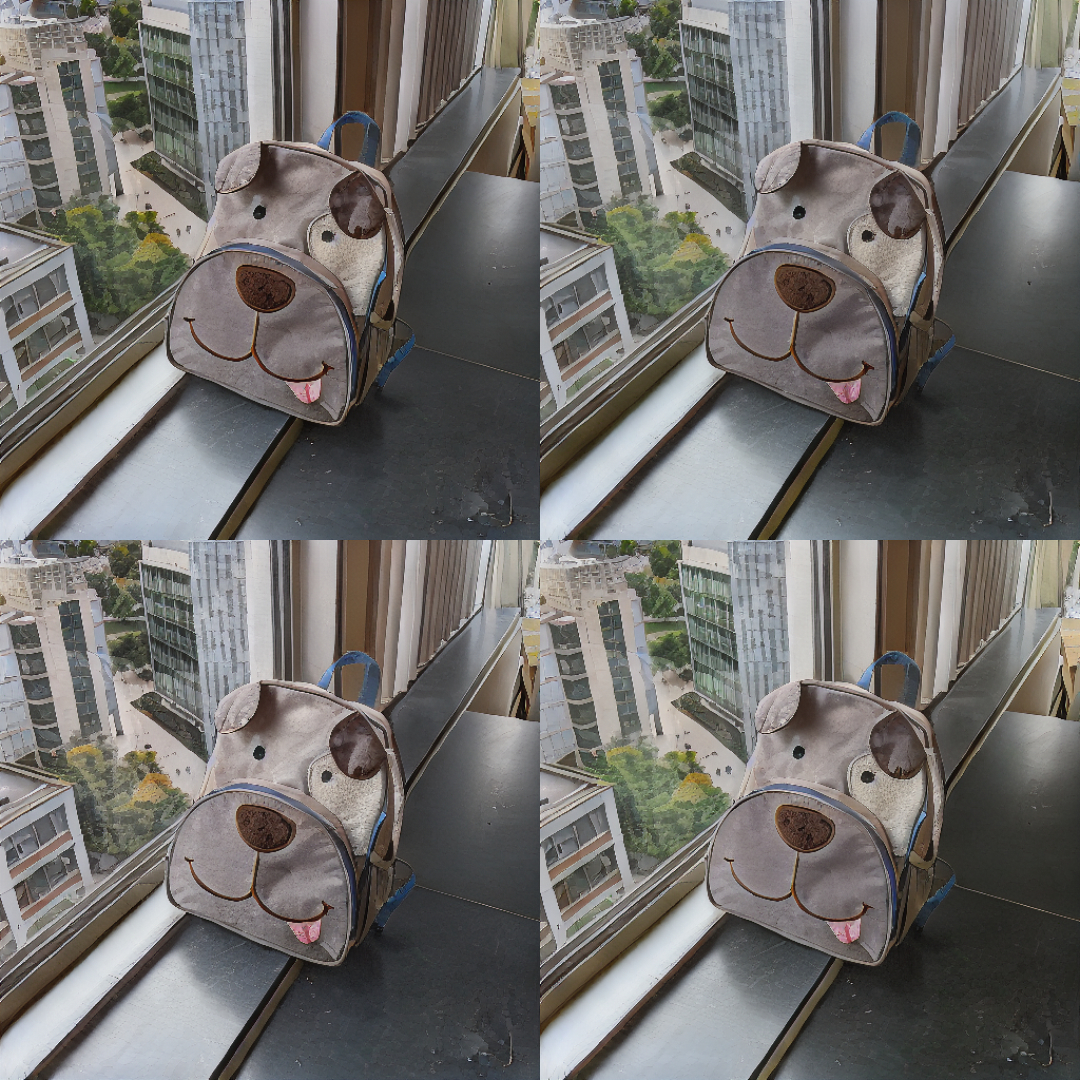} &
        \includegraphics[height=0.18\textwidth, keepaspectratio]{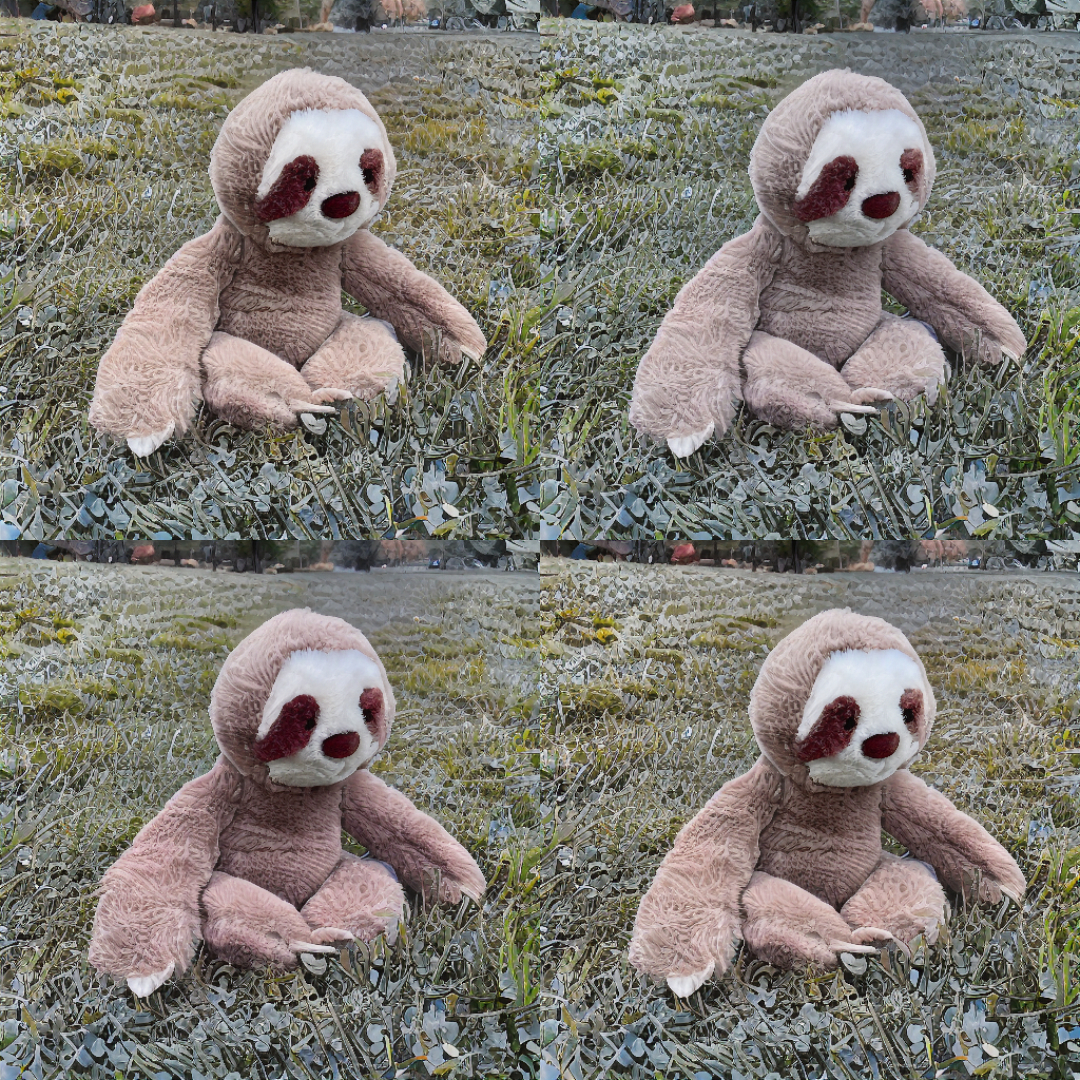} \\
        \rotatebox{90}{\parbox{3cm}{\centering\textbf{Offline}}} &
        \includegraphics[height=0.18\textwidth, keepaspectratio]{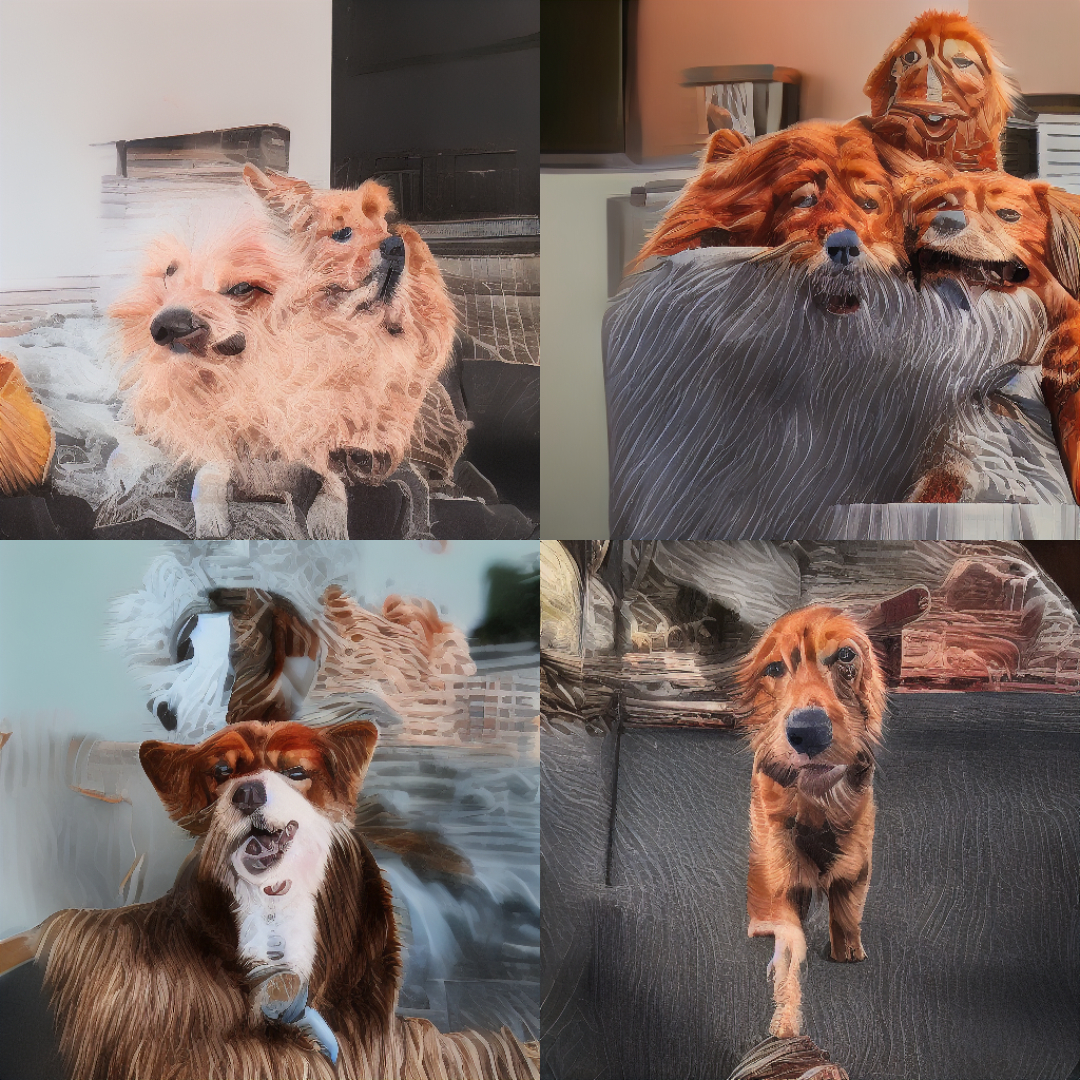} &
        \includegraphics[height=0.18\textwidth, keepaspectratio]{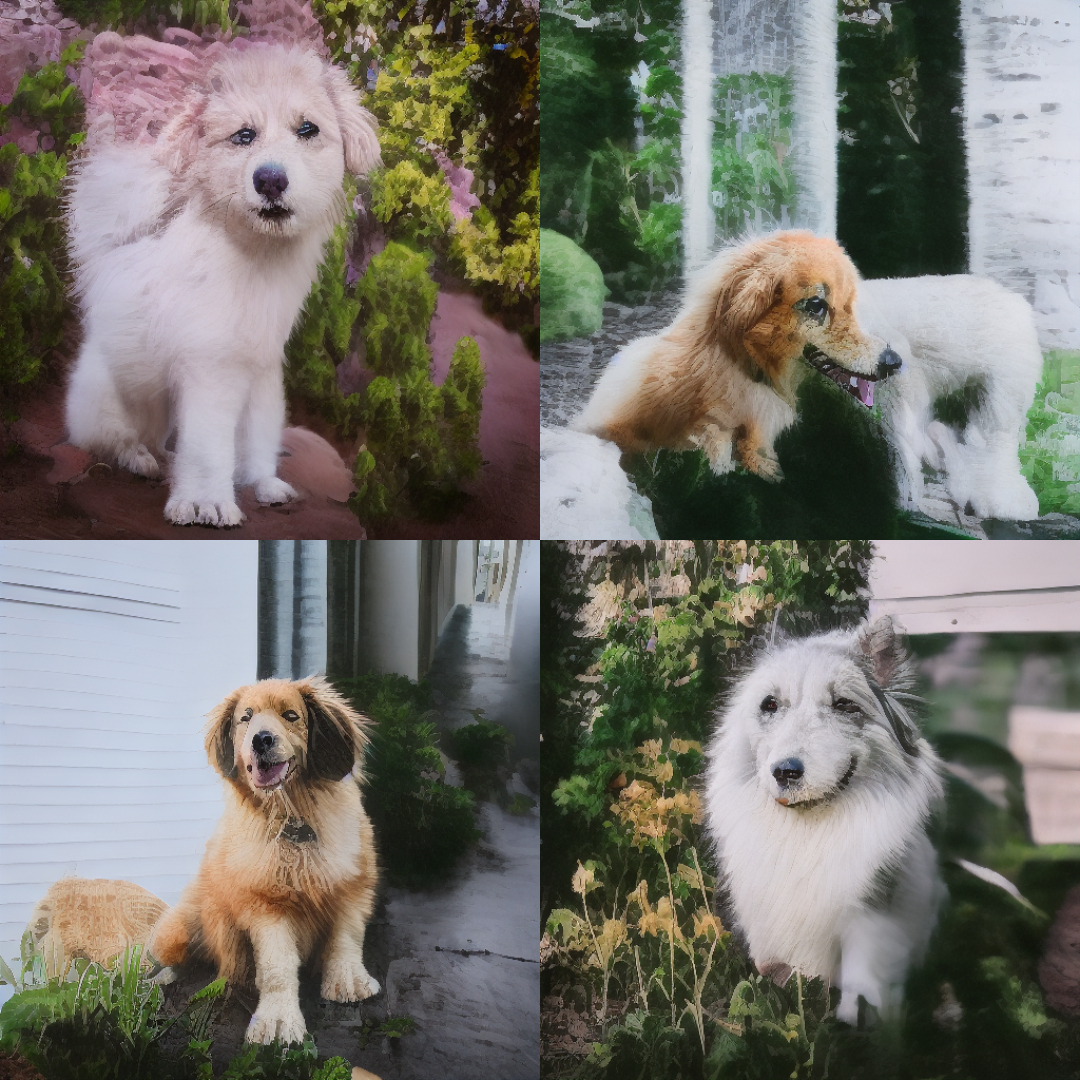} &
        \includegraphics[height=0.18\textwidth, keepaspectratio]{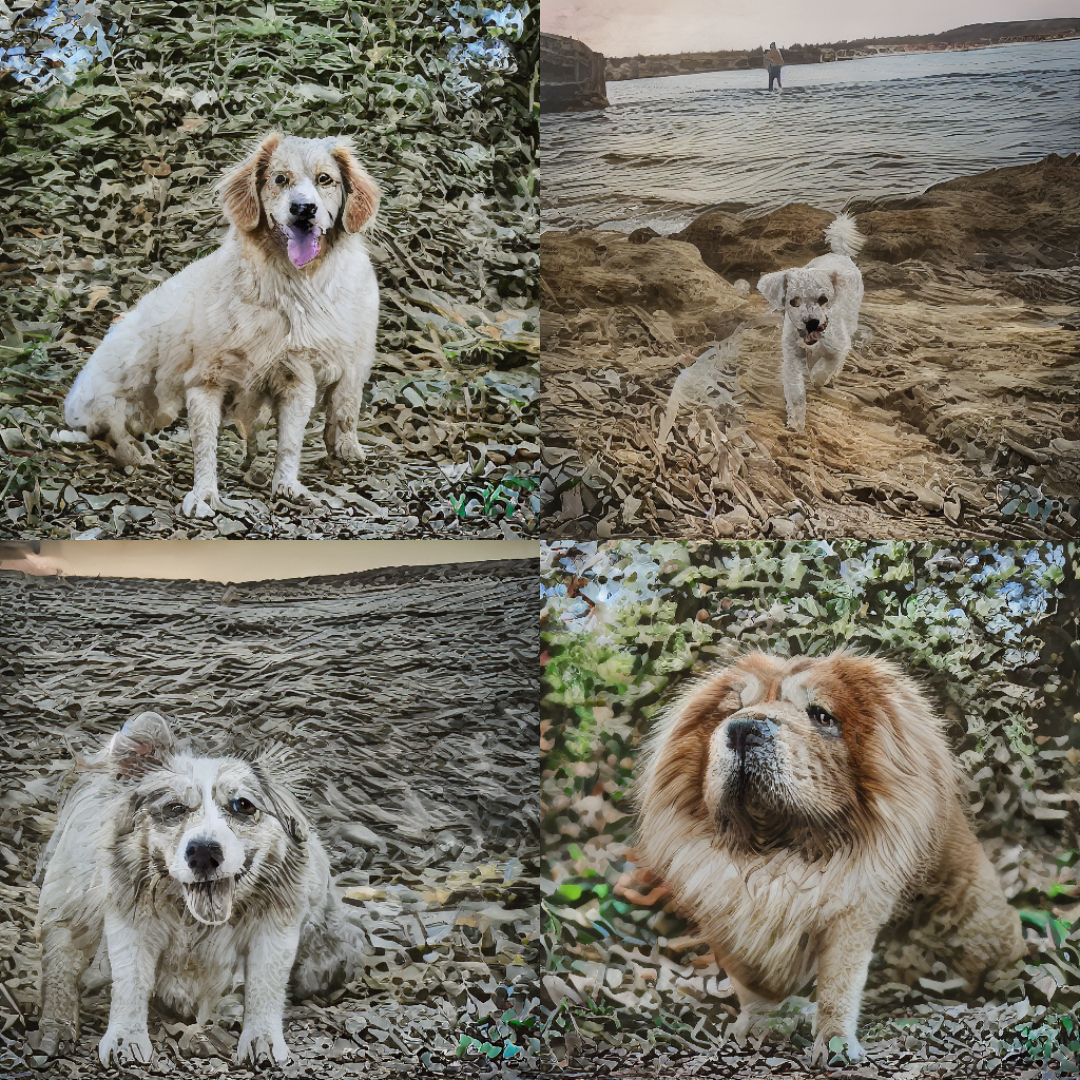} &
        \includegraphics[height=0.18\textwidth, keepaspectratio]{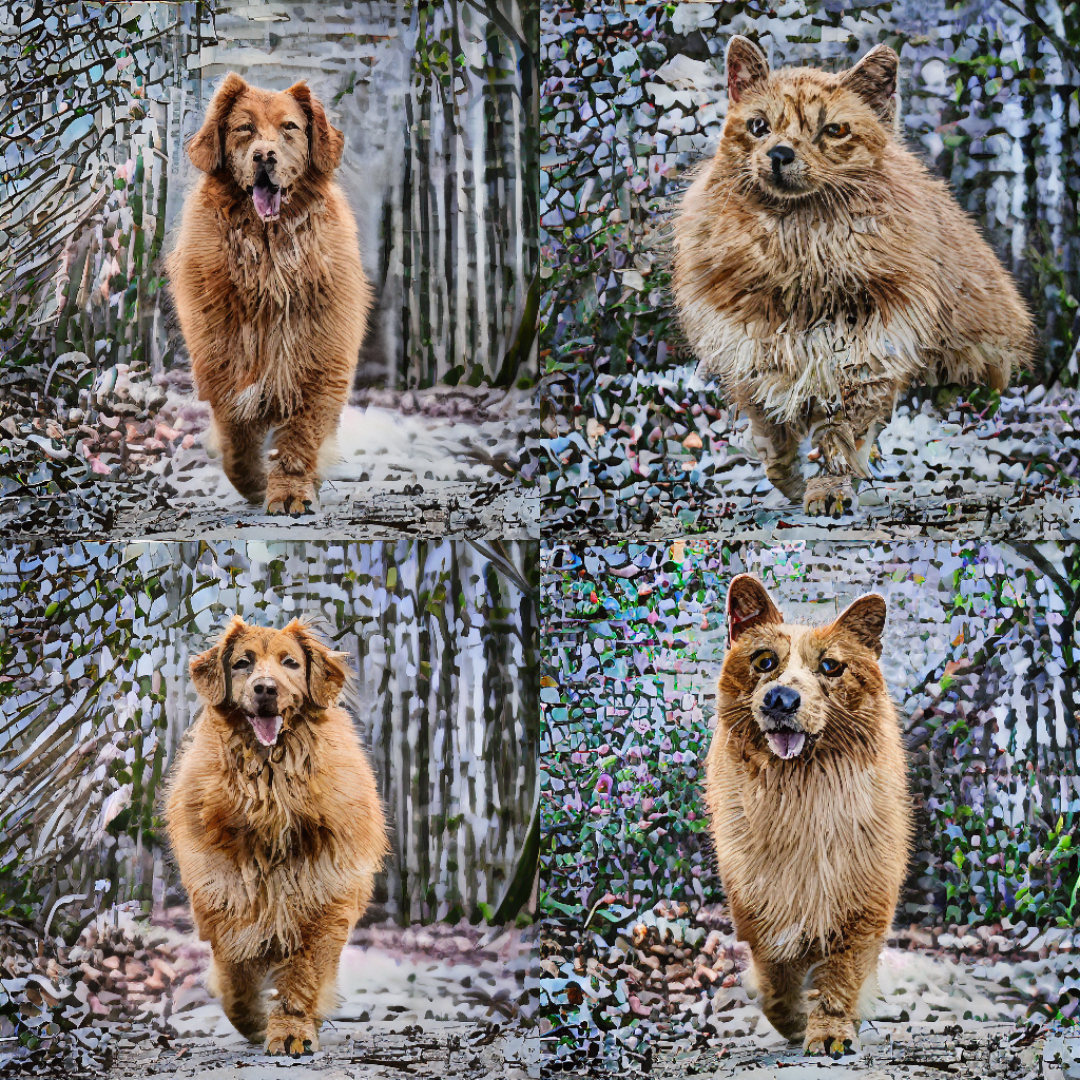} &
        \includegraphics[height=0.18\textwidth, keepaspectratio]{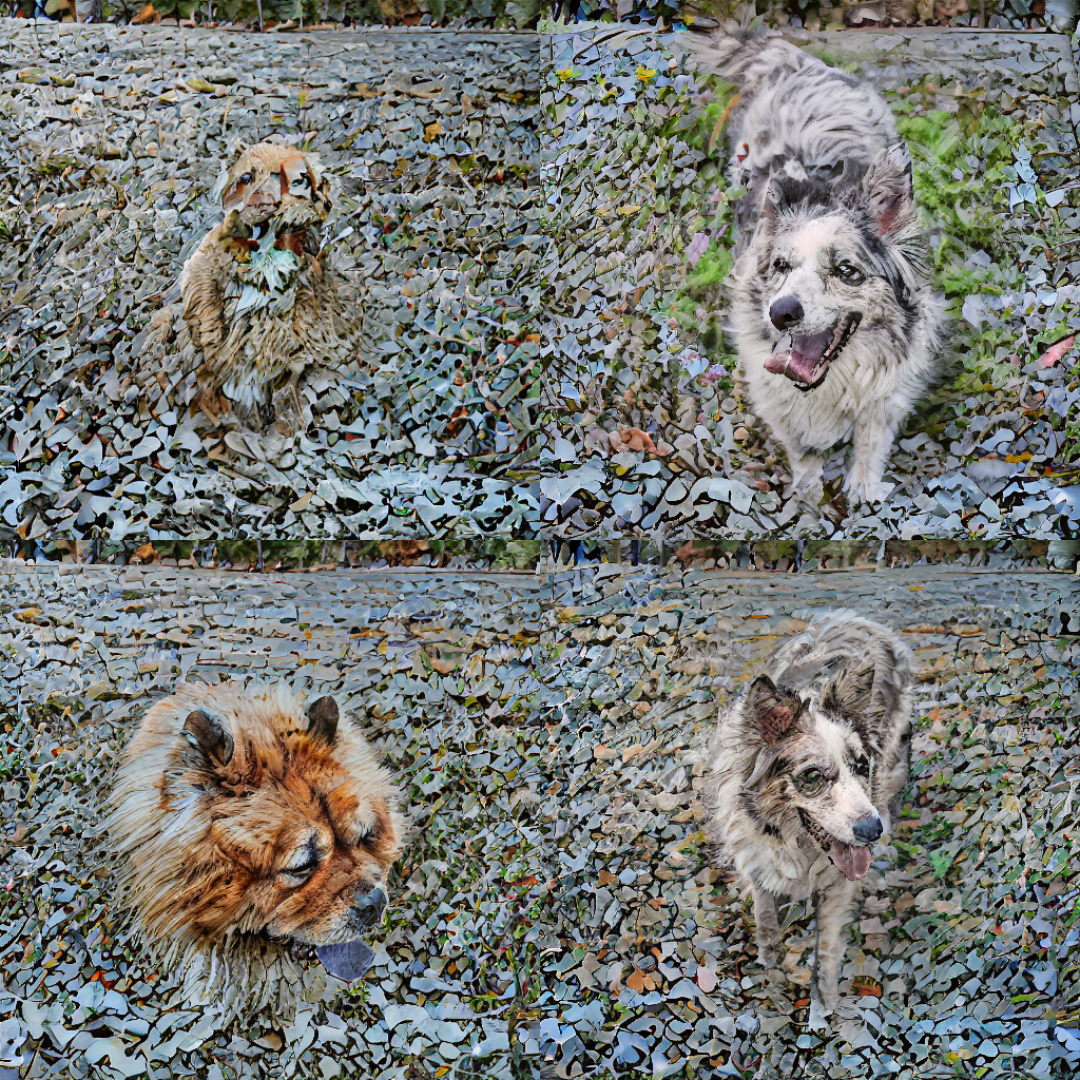} 
    \end{tabular}%
    }
    \caption{Comparison of four outputs per method for each concept immediately after learning that respective concept. LR achieves both concept preservation and high diversity (varied breeds, poses, backgrounds), while Naive, SLR, and ER generate incorrect concepts with poor variation. Only LR approaches Offline's performance while using a fraction of the memory.}
    \label{fig:multi_qualitative}
\end{figure*}

In addition to these quantitative assessments, we provide a qualitative evaluation of how well each method retains earlier concepts under a small memory setting (10 images for ER, 480 latents for LR). Specifically, we compare generation results for the prompt "a photo of V* concept" after each stage of training (\emph{dog}, \emph{toy}, \emph{cat}, \emph{backpack}, and \emph{plushie}). Our qualitative analysis consists of two complementary visualizations: a single-image comparison (Figure~\ref{fig:single_qualitative}), which shows one representative output per method at each stage of sequential learning to track how well the original concept is preserved, and a four-image comparison (Figure~\ref{fig:multi_qualitative}), which displays multiple outputs per method to demonstrate within-concept diversity at each training stage. 

\begin{figure*}[tp]
    \centering
    \resizebox{\textwidth}{!}{%
    \begin{tabular}{c@{\hspace{0.5em}}c@{\hspace{0.5em}}c@{\hspace{0.5em}}c@{\hspace{0.5em}}c@{\hspace{0.5em}}c}
        & \textbf{Dog} & \textbf{Toy} & \textbf{Cat} & \textbf{Backpack} & \textbf{Plushie} \\
        \rotatebox{90}{\parbox{3cm}{\centering\textbf{Naive}}} &
        \includegraphics[height=0.18\textwidth, keepaspectratio]{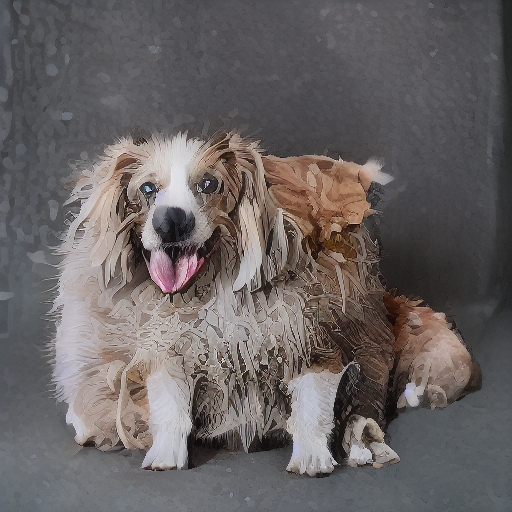} &
        \includegraphics[height=0.18\textwidth, keepaspectratio]{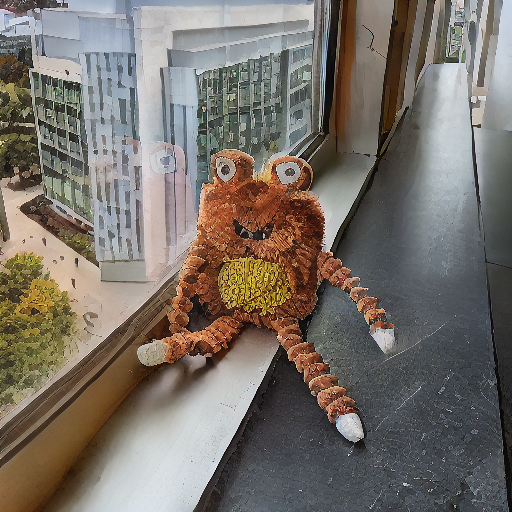} &
        \includegraphics[height=0.18\textwidth, keepaspectratio]{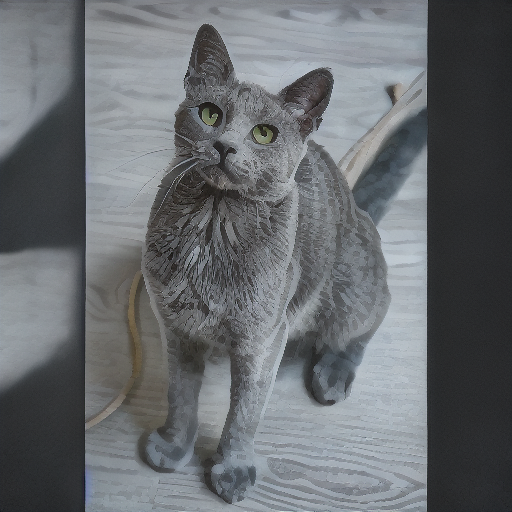} &
        \includegraphics[height=0.18\textwidth, keepaspectratio]{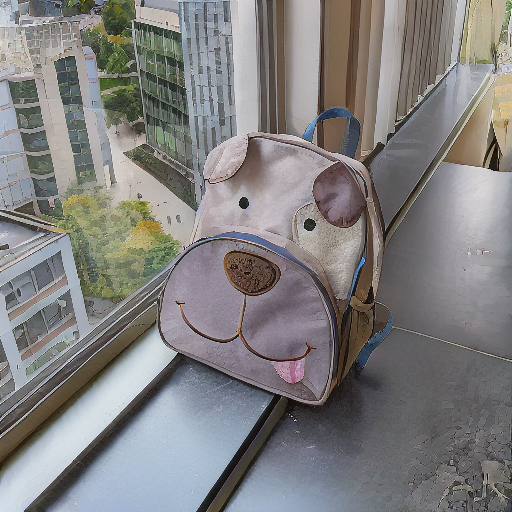} &
        \includegraphics[height=0.18\textwidth, keepaspectratio]{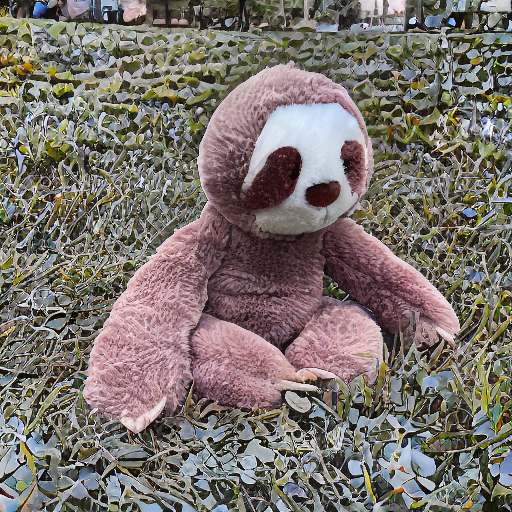} \\
        \rotatebox{90}{\parbox{3cm}{\centering\textbf{ER}}} &
        \includegraphics[height=0.18\textwidth, keepaspectratio]{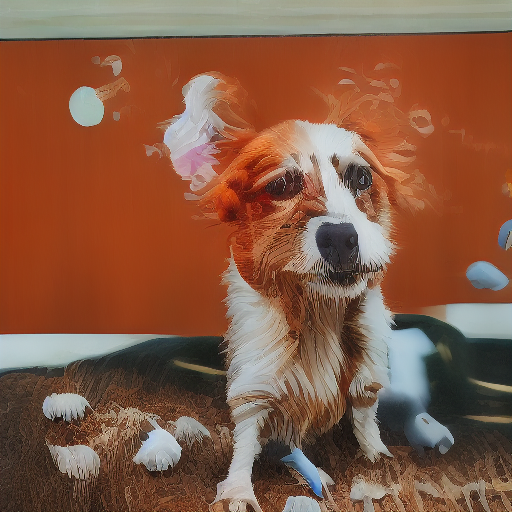} &
        \includegraphics[height=0.18\textwidth, keepaspectratio]{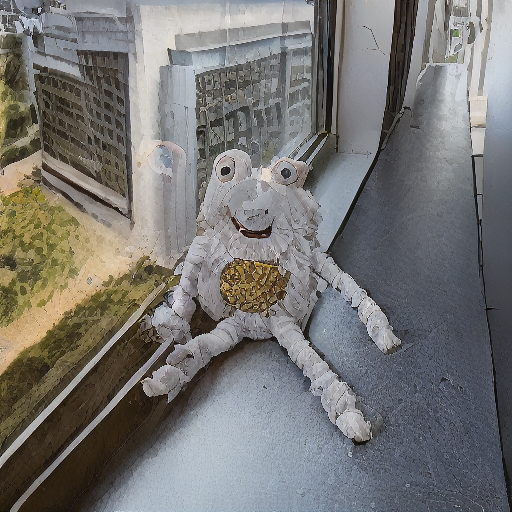} &
        \includegraphics[height=0.18\textwidth, keepaspectratio]{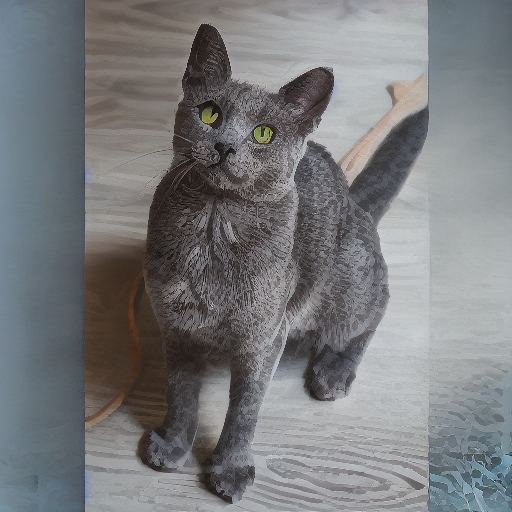} &
        \includegraphics[height=0.18\textwidth, keepaspectratio]{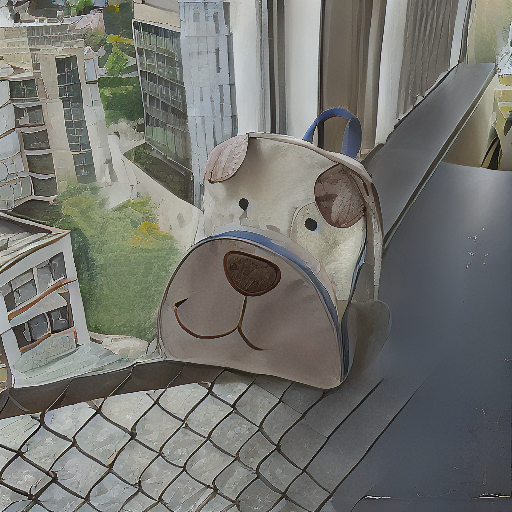} &
        \includegraphics[height=0.18\textwidth, keepaspectratio]{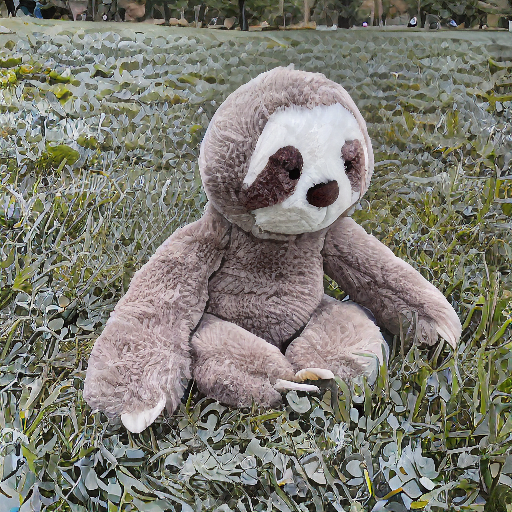} \\
        \rotatebox{90}{\parbox{3cm}{\centering\textbf{LR}}} &
        \includegraphics[height=0.18\textwidth, keepaspectratio]{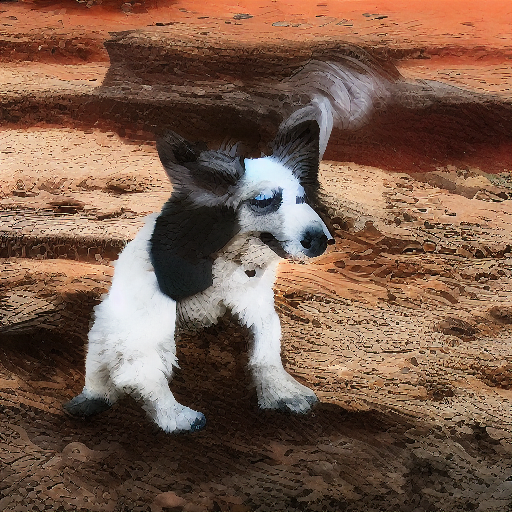} &
        \includegraphics[height=0.18\textwidth, keepaspectratio]{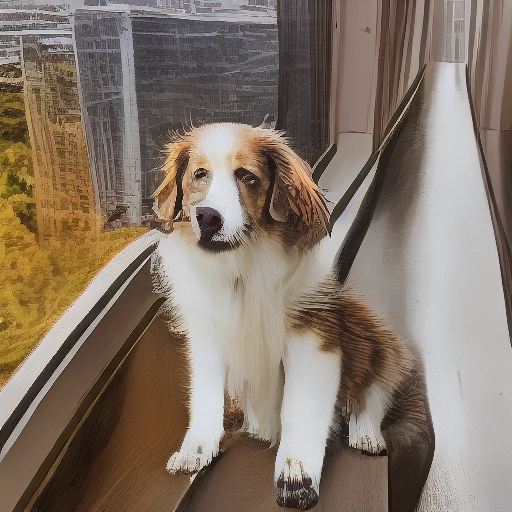} &
        \includegraphics[height=0.18\textwidth, keepaspectratio]{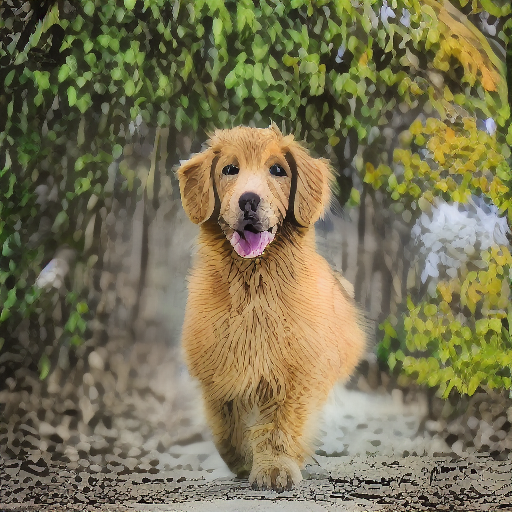} &
        \includegraphics[height=0.18\textwidth, keepaspectratio]{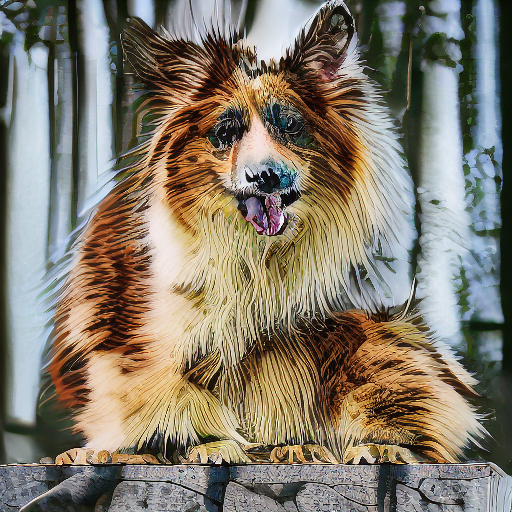} &
        \includegraphics[height=0.18\textwidth, keepaspectratio]{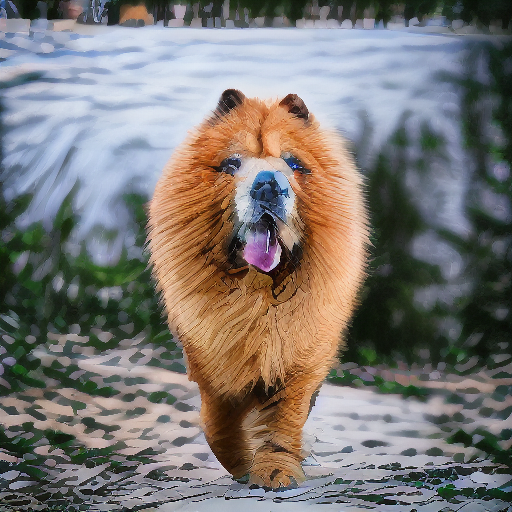} \\
        \rotatebox{90}{\parbox{3cm}{\centering\textbf{SLR}}} &
        \includegraphics[height=0.18\textwidth, keepaspectratio]{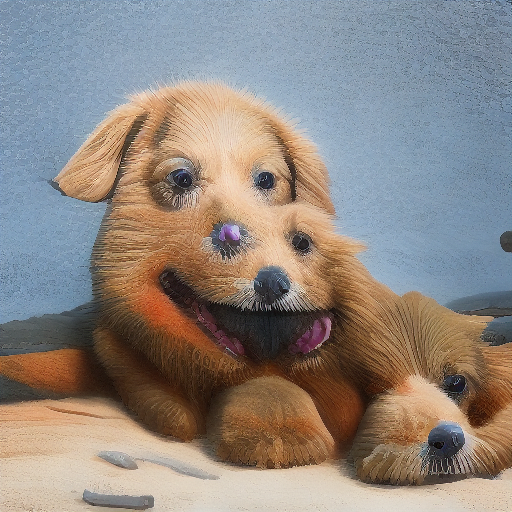} &
        \includegraphics[height=0.18\textwidth, keepaspectratio]{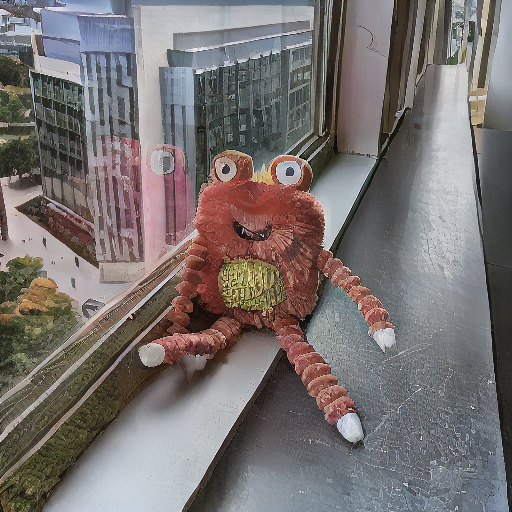} &
        \includegraphics[height=0.18\textwidth, keepaspectratio]{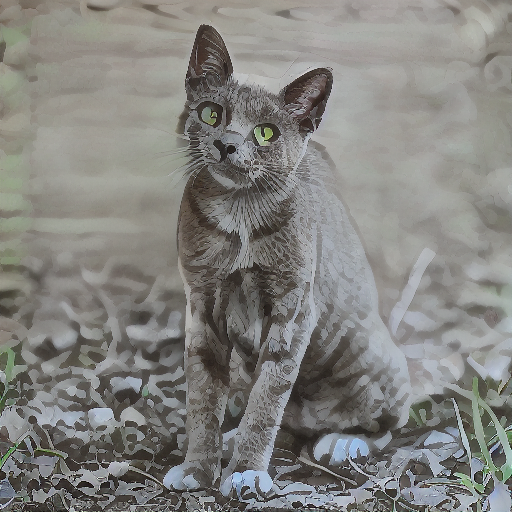} &
        \includegraphics[height=0.18\textwidth, keepaspectratio]{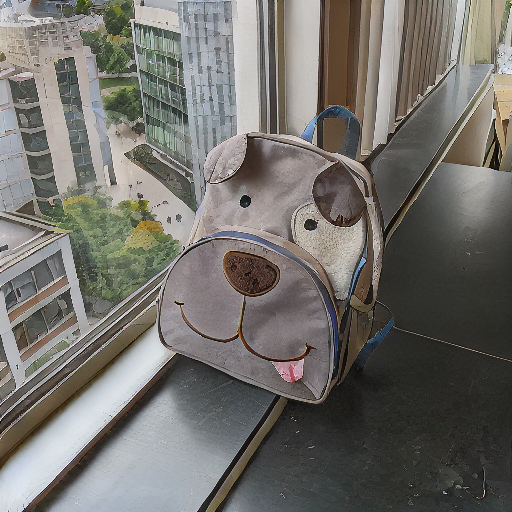} &
        \includegraphics[height=0.18\textwidth, keepaspectratio]{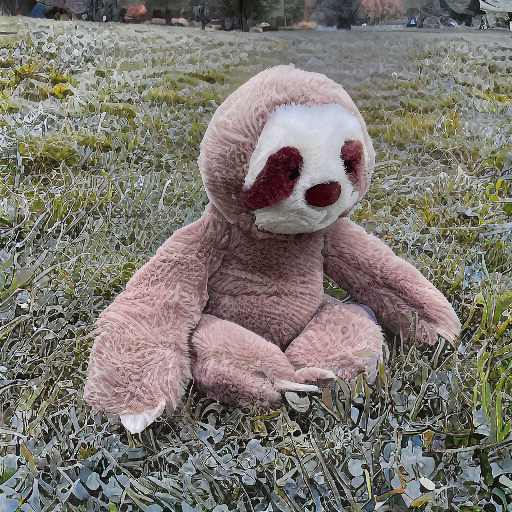} \\
        \rotatebox{90}{\parbox{3cm}{\centering\textbf{Offline}}} &
        \includegraphics[height=0.18\textwidth, keepaspectratio]{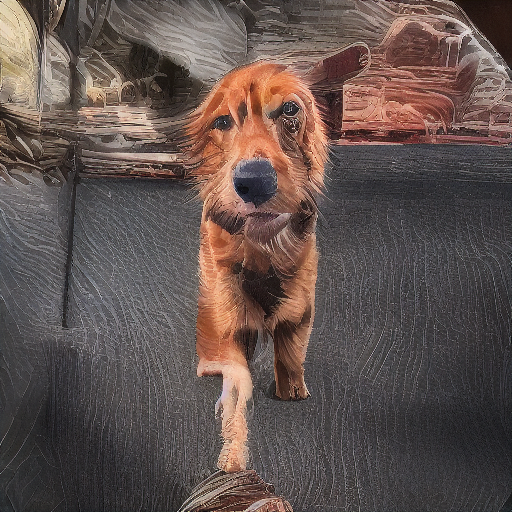} &
        \includegraphics[height=0.18\textwidth, keepaspectratio]{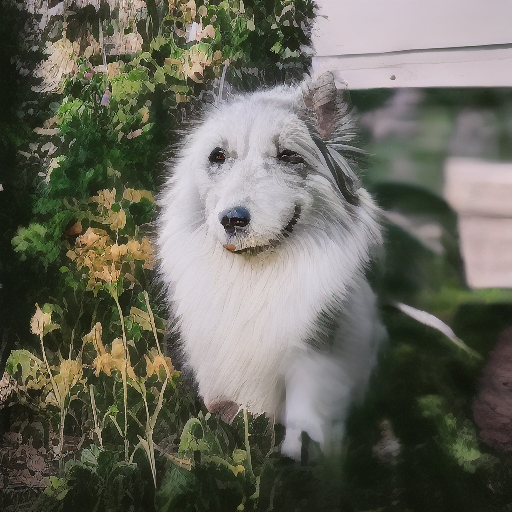} &
        \includegraphics[height=0.18\textwidth, keepaspectratio]{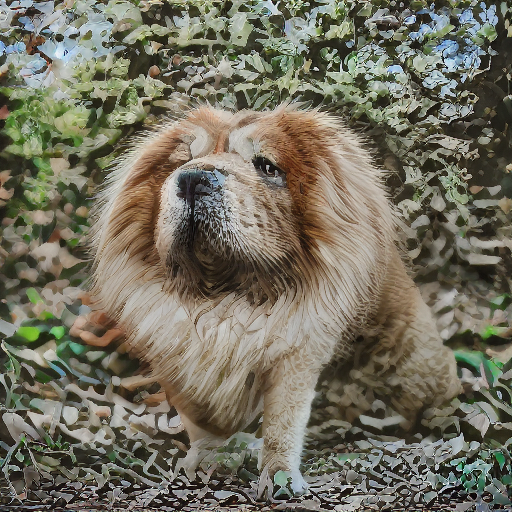} &
        \includegraphics[height=0.18\textwidth, keepaspectratio]{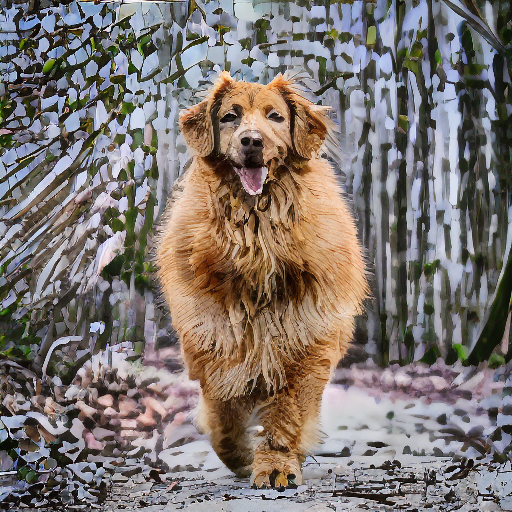} &
        \includegraphics[height=0.18\textwidth, keepaspectratio]{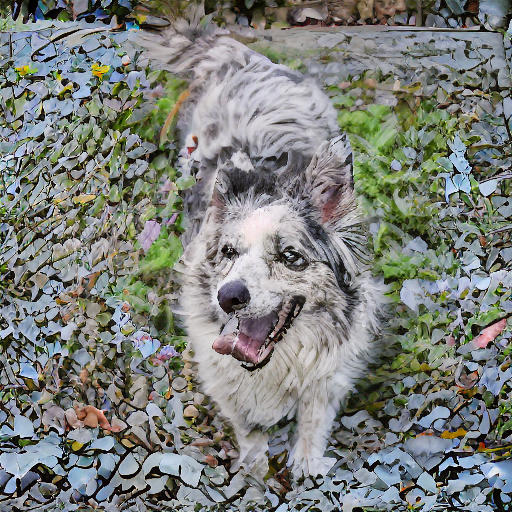}
    \end{tabular}%
    }
    \caption{Progression of dog concept retention across sequential learning. Each row represents a method (Naive, ER, LR, SLR, Offline); each column shows the result of prompting "a photo of V* dog" immediately after learning the task indicated at the top (Dog, Toy, Cat, etc.). LR maintains dog concept fidelity throughout all sequential tasks, while Naive, ER, and SLR increasingly generate the most recently learned concept instead of dogs. Offline demonstrates perfect preservation but requires all task data simultaneously.}
    \label{fig:single_qualitative}
\end{figure*}

\subsection{Single-Image Comparison (Figure~\ref{fig:single_qualitative})}

Figure~\ref{fig:single_qualitative} presents one representative output per method at each stage of sequential learning. The purpose of this comparison is to assess how well each method retains the original concept (\emph{dog}) as new tasks are introduced, highlighting differences in catastrophic forgetting across methods. 

\textbf{Naive} exhibits catastrophic forgetting almost immediately after learning \emph{toy,} with outputs drifting toward newly learned objects instead of dogs. For example, after learning the \emph{cat} task, the model generates a high-quality cat image instead of a dog, indicating that the model has completely forgotten the original concept but maintains overall generation quality.

\textbf{ER} behaves similarly to Naive despite occasionally generating partial dog shapes or elements resembling dog appearance. For example, the output for the \emph{toy} task shows a toy that is white, potentially resembling one of the replayed images of a white dog. The \emph{backpack} output also shows a background that seems to incorporate elements from a previous task. This suggests that while ER attempts to maintain previous concepts, its 10-image buffer is too small to effectively retain the older concept. However, as shown in our ablation studies (Section \ref{sec:ablation_detailed}), we find that even larger ER buffers improve but don't fully resolve this limitation.

\textbf{LR} retains a coherent dog shape and texture throughout, showing minimal forgetting. This aligns with its stronger IA/TA scores for earlier tasks. However, the generated outputs show some distortion. Interestingly, elements of successive tasks appear mixed with the original concept. For instance, after training on the \emph{toy} task, the generated image appears to show a dog but with a background (by the window in a building) that is seen in one of the \emph{toy} task training images.

\textbf{SLR} generally forgets the dog concept, similar to Naive or ER. The selective similarity approach may discard key latent examples of \emph{dog} by prioritizing latents that are similar to the current task, inadvertently accelerating concept drift.

\textbf{Offline} replays all prior data, showing perfect or near-perfect dog fidelity while also accommodating new tasks. This approach serves as the upper bound for performance.

Thus, while ER and SLR produce visually coherent images, they often reflect newly learned tasks rather than the original dog. LR manages to preserve the \emph{dog} concept, whereas Offline stands as the ideal with no forgetting and minimal distortion.

\subsection{Four-Image Comparison for Diversity (Figure~\ref{fig:multi_qualitative})}

Figure~\ref{fig:multi_qualitative} presents four representative outputs per method after each stage of sequential learning to evaluate within-task diversity. This comparison highlights how well each method maintains variation in generated images for the same concept over time, assessing whether mode collapse occurs as new tasks are introduced.

For the \textbf{Naive} approach, we observe that despite generating the wrong concept (as already established in the single-image analysis), there is generally poor diversity, as the objects and backgrounds are nearly identical for most tasks, except for the \emph{cat} task which shows different positions, styles, and types of cats. This suggests that while the issue is primarily concept substitution, mode collapse is occurring severely.

\textbf{ER}'s outputs display similar diversity characteristics to Naive, with slightly more variation in the \emph{toy} task. One of the toy outputs actually displayed a dog, and other images also show influence from a replayed dog image that is white. Other tasks show mode collapse, especially after the \emph{cat} task where the four images are nearly identical. The performance is similar to Naive in that the main problem is catastrophic forgetting and producing incorrect concepts, but there is also minimal diversity within tasks.

\textbf{LR}'s outputs stand out not only for maintaining the correct concept (dogs) but also for exhibiting remarkable diversity within that concept. The dog images show substantial variations in breed (from small to large dogs), pose (sitting, standing, running), background scenes, and coloration patterns. This visual diversity directly corresponds to LR's superior quantitative diversity score of 1.66 versus 1.17 for ER and 1.13 for Naive. LR's memory buffer effectively captures the distribution of the concept rather than just preserving a few exemplars.

\textbf{SLR} shows generated images that appear more homogeneous across samples, suggesting that similarity-based selection leads to a narrowing of the captured concept distribution. The diversity seems to be less than Naive, and not comparable to standard LR.

\textbf{Offline} training demonstrates the ideal case: high-fidelity dog images with substantial diversity across samples. It is actually similar to LR in performance. The backgrounds vary from natural outdoor scenes to indoor settings, the dogs appear in different poses and perspectives, and various breeds and colors are represented. This breadth of diversity establishes a ceiling for what's achievable when all task data is available simultaneously.

These observations highlight an important nuance that catastrophic forgetting and mode collapse are distinct but related challenges. Methods can suffer from either or both issues, and addressing one doesn't necessarily resolve the other. LR's success in maintaining both concept fidelity and diversity visually as good as Offline demonstrates why it represents a substantial improvement over traditional approaches like ER, especially in memory-constrained settings.

\subsection{Further Analysis and Suggestions}

While these results support the quantitative findings in Section~\ref{sec:quant_results}, further analyses could give deeper insights.

\begin{itemize}
    % \item \textbf{Replay Memory Evolution:} Inspect which dog examples are retained or replaced over time in the replay buffer, to see if the buffer drifts toward recent tasks at the expense of early concepts.
    \item \textbf{Hybrid Approaches:} Combining a small set of raw images with a larger latent buffer may preserve high‐fidelity exemplars alongside broader coverage.  
    \item \textbf{Adaptive Buffer Sizes:} Dynamically adjusting the buffer based on measured forgetting or domain shift could more efficiently balance memory usage and retention.
\end{itemize}

Overall, LR best balances memory footprint and forgetting mitigation in the small memory setting, whereas Offline remains an ideal but resource‐intensive solution. Naive, ER, and SLR struggle to preserve older tasks or maintain diversity when memory is severely constrained.

\section{Detailed Ablation Study Results}
\label{sec:ablation_detailed}

We conducted several ablation studies to better understand the factors affecting catastrophic forgetting and mode collapse in continual text-to-image diffusion. This section presents detailed results from these studies.

\subsection{Memory Size Variation}
\label{subsec:memory_size_ablation}

To investigate how memory capacity affects performance, we tested three different buffer sizes for both ER and LR, maintaining comparable memory footprints between methods:

\begin{itemize}
    \item \textbf{Small}: 10 images (~31MB) vs. 480 latents (~31MB)
    \item \textbf{Medium}: 20 images (~61MB) vs. 960 latents (~63MB)
    \item \textbf{Large}: 100 images (~307MB) vs. 4800 latents (~313MB)
\end{itemize}

Table~\ref{tab:memory_size_ablation} summarizes the TFR-IA (Task Forgetting Rate for Image Alignment) and final diversity scores for the Dog task (the earliest task, most prone to forgetting) across these configurations.

\begin{table}[h!]
\centering
\caption{Impact of memory size on TFR-IA and Dog task diversity. Values represent Mean $\pm$ standard deviation across ten runs. Increasing buffer size improves retention (lower TFR-IA) and diversity, but LR consistently outperforms ER at comparable memory budgets. Even with a small buffer, LR achieves lower forgetting and higher diversity than ER with a much larger buffer, demonstrating the efficiency of latent-based memory storage.}
\label{tab:memory_size_ablation}
\begin{tabular}{lcccc}
\hline
\textbf{Method} & \textbf{Memory Size} & \textbf{TFR-IA} & \textbf{Dog Diversity} & \textbf{Overhead} \\
\hline
Naive & -- & 16.16 $\pm$ 2.71 & 1.13 $\pm$ 0.08 & Lowest \\
\hline
ER & Small (10 img) & 13.59 $\pm$ 2.28 & 1.17 $\pm$ 0.12 & Low \\
ER & Medium (20 img) & 13.51 $\pm$ 2.14 & 1.12 $\pm$ 0.04 & Medium \\
ER & Large (100 img) & 7.26 $\pm$ 1.48 & 1.34 $\pm$ 0.16 & High \\
\hline
LR & Small (480 lat) & 4.84 $\pm$ 1.72 & 1.66 $\pm$ 0.21 & Low \\
LR & Medium (960 lat) & 3.97 $\pm$ 1.35 & 1.65 $\pm$ 0.22 & Medium \\
LR & Large (4800 lat) & 2.16 $\pm$ 0.95 & 1.80 $\pm$ 0.12 & High \\
\hline
SLR & Small (480 lat) & 18.48 $\pm$ 1.56 & 1.10 $\pm$ 0.03 & Low \\
SLR & Medium (960 lat) & 14.36 $\pm$ 1.48 & 1.10 $\pm$ 0.05 & Medium \\
SLR & Large (4800 lat) & 10.02 $\pm$ 1.25 & 1.11 $\pm$ 0.04 & High \\
\hline
Offline & -- & 0.32 $\pm$ 0.85 & 1.72 $\pm$ 0.18 & Highest \\
\hline
\end{tabular}
\end{table}

Both \textbf{ER} and \textbf{LR} show improvements with larger buffers, but with diminishing returns. For \textbf{LR}, increasing from medium to large (960 to 4800 latents) yields a 1.81\% improvement in TFR-IA (from 3.97\% to 2.16\%), while the improvement from small to medium (480 to 960 latents) is 0.87\% (from 4.84\% to 3.97\%). This suggests that moderate-sized buffers may offer the best trade-off between performance and resource utilization. Even the small LR configuration (480 latents) outperforms the large ER configuration (100 images) on both TFR-IA (4.84\% vs. 7.26\%) and diversity (1.66 vs. 1.34), despite using only approximately 10\% of the memory footprint (~31MB vs. ~307MB). This reinforces the central hypothesis that latent representations offer a significantly more memory-efficient way to preserve knowledge of previous tasks.

\textbf{ER} shows a substantial improvement only at the large buffer size, with TFR-IA dropping from 13.51\% (medium) to 7.26\% (large). This suggests that raw image replay requires a critical mass of examples to effectively preserve task knowledge, while LR provides strong performance even with modest buffer sizes. \textbf{LR} maintains consistently high diversity scores across all buffer sizes (1.66--1.80), approaching the diversity of the Offline upper bound (1.72). In contrast, \textbf{ER} achieves moderate diversity improvement at the large buffer size (1.34), still well below LR's small buffer performance (1.66). This indicates that latent representations more effectively capture the full distribution of a concept than raw images at equivalent memory budgets. By requiring less storage per example, latents allow a greater number of past instances to be retained in the same memory footprint, leading to a more comprehensive representation of task variations. 

As buffer sizes increase, \textbf{LR}'s performance (2.16\% TFR-IA) begins to approach that of the Offline upper bound (0.32\% TFR-IA), suggesting that with sufficient latent storage, sequential learning can achieve results competitive with simultaneous training on all tasks. This is particularly notable given that the Offline approach requires access to all data at all times, while LR maintains only a fixed-size buffer of compressed representations.

\textbf{SLR} unexpectedly performs worse than standard LR across all memory sizes, with the small buffer SLR showing TFR-IA values (18.48\%) even higher than Naive fine-tuning (16.16\%). While SLR's performance improves with larger buffers (14.36\% for medium, 10.02\% for large), it consistently underperforms compared to standard LR at equivalent memory sizes. Our analysis suggests that similarity-based selection creates a problematic bias in the replay buffer. When training on a new concept (e.g., cat), SLR selects latents from previous concepts (e.g., dog) that are most similar to the current concept's examples. This preferentially replays boundary cases or examples that share features between concepts, while under-representing the distinctive characteristics of previous concepts. Another contributing factor may be that similarity-based selection tends to repeatedly choose the same few examples for replay—those that happen to share features with the current task (e.g., a few dog images that appear cat-like or toy-like). This creates a narrow, biased representation of the previous concept, causing the model to become fixated on these boundary examples while forgetting the broader distribution of the original concept. Rather than helping the model maintain separation between concepts, this approach appears to cause concept drift, where representations of earlier concepts gradually shift toward newer ones. Random sampling in standard LR, by contrast, preserves a more diverse and representative distribution of previous concept features, maintaining clearer separation between concept manifolds.

These findings strongly suggest the memory efficiency advantage of LR over traditional ER methods, while also revealing the limitations of similarity-based selection strategies in the latent space. The strong performance of LR even at small buffer sizes suggests that this approach could be particularly valuable in resource-constrained environments or applications with strict privacy requirements that limit data retention.

\subsection{Replay Weight Analysis}
\label{subsec:replay_weight_ablation}

We investigated how the weighting parameter $\lambda_{\text{memory}}$ affects the balance between retaining old knowledge and acquiring new tasks. This parameter controls the relative importance of the replay loss versus the current task loss. Table~\ref{tab:replay_weight_ablation} shows results for LR with different $\lambda_{\text{memory}}$ values, evaluated on TFR-IA and the final IA for Task 1 (Dog) and Task 5 (Plushie).

\begin{table}[h!]
\centering
\caption{Impact of replay weight $\lambda_{\text{memory}}$ on LR performance.}
\label{tab:replay_weight_ablation}
\begin{tabular}{lccc}
\hline
\textbf{$\lambda_{\text{memory}}$} & \textbf{TFR-IA} & \textbf{Dog IA (\%)} & \textbf{Plushie IA (\%)} \\
\hline
0.1 & 9.21 $\pm$ 1.85 & 68.74 $\pm$ 3.62 & 79.55 $\pm$ 3.01 \\
0.3 & 6.38 $\pm$ 1.63 & 73.48 $\pm$ 4.65 & 78.94 $\pm$ 2.86 \\
0.5 & 4.84 $\pm$ 1.72 & 77.59 $\pm$ 5.30 & 77.72 $\pm$ 3.24 \\
0.7 & 3.69 $\pm$ 1.48 & 79.28 $\pm$ 4.87 & 74.63 $\pm$ 3.68 \\
0.9 & 2.75 $\pm$ 1.32 & 81.67 $\pm$ 3.94 & 72.19 $\pm$ 4.12 \\
\hline
\end{tabular}
\end{table}

From this, we see that higher $\lambda_{\text{memory}}$ values (0.7, 0.9) significantly reduce forgetting of earlier tasks but negatively impact performance on the most recent task. A $\lambda_{\text{memory}}$ value of 0.5 provides a balanced compromise, with reasonable performance on both early and recent tasks. This analysis empirically demonstrates the fundamental stability-plasticity trade-off in continual learning: stronger emphasis on replay (stability) reduces forgetting but impedes new learning (plasticity).

These ablation studies provide insights into the factors affecting continual learning in text-to-image diffusion models. The detailed findings confirm that LR offers the best balance of efficiency and performance, particularly with appropriate hyperparameter settings like a balanced replay weight ($\lambda_{\text{memory}}=0.5$), generous loss thresholds, and diverse replay sampling.

\subsection{Training Stability Factors}
\label{subsec:training_stability}

We explored how different loss thresholds (1.0 vs. 1.5), fixed vs. random replay batches, and restart strategies affected training stability and performance. We found that broader thresholds (1.5) were necessary to accommodate the combined replay loss, particularly for methods that balance current task loss with memory loss. More stringent thresholds often triggered unnecessary restarts, disrupting the learning process.

Additionally, enforcing a minimum of 100 steps before allowing early termination proved crucial for preventing incomplete learning of each concept. When early termination was allowed without this constraint, models sometimes terminated training prematurely before adequately learning the concept, especially for more complex objects.

\subsection{Task Order Effects}
\label{subsec:task_order_effects}

\begin{figure}[ht]
    \centering
    \includegraphics[width=\linewidth]{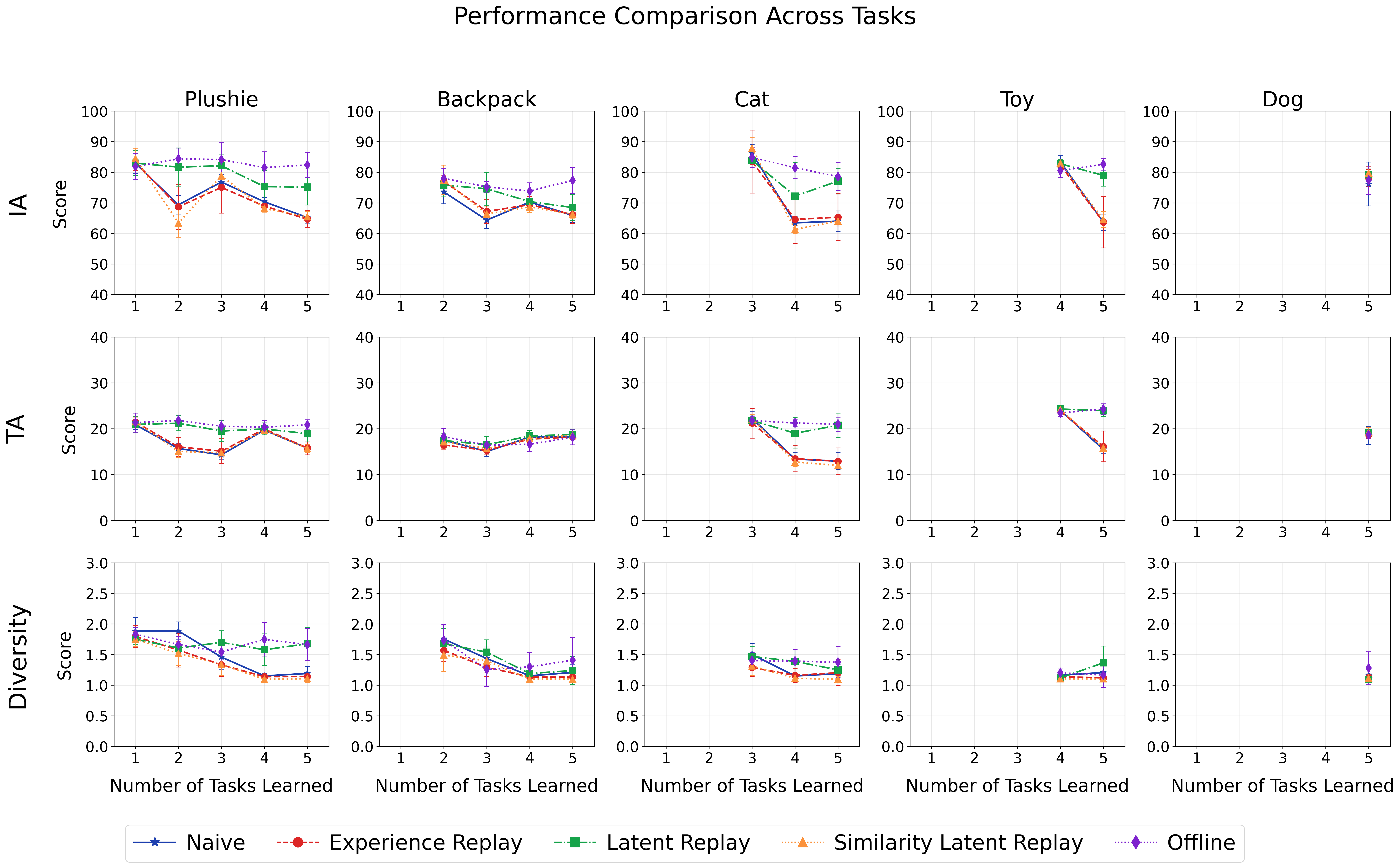}
    \caption{Performance metrics across reversed task sequence (plushie → backpack → cat → toy → dog) showing consistent patterns regardless of task order. LR maintains superior performance in Image Alignment (top), Text Alignment (middle), and Diversity (bottom) compared to Naive, ER, and SLR methods. The similar patterns between original and reversed sequences confirm that temporal position, not concept identity, primarily determines forgetting rates. LR's advantage persists regardless of sequence, approaching Offline's performance while using limited memory. Error bars represent one standard deviation across ten runs.}
    \label{fig:combinedplotsreverse}
\end{figure}

To examine how the sequence of tasks affects forgetting patterns, we evaluated alternative orderings of our five concepts. Specifically, we compared the original order (dog → toy → cat → backpack → plushie) with the reversed order (plushie → backpack → cat → toy → dog), measuring IA, TA, and Diversity metrics across all methods. Figures~\ref{fig:combinedplotsreverse} show the evolution of these metrics in the reversed task sequence.

Table~\ref{tab:task_order} compares the overall Task Forgetting Rate (TFR-IA) between the original and reversed task orders.

\begin{table}[h!]
\centering
\caption{Impact of task order on TFR-IA.}
\label{tab:task_order}
\begin{tabular}{lcc}
\hline
\textbf{Method} & \textbf{Original Order} & \textbf{Reversed Order} \\
\hline
Naive & 16.16 $\pm$ 2.71 & 15.83 $\pm$ 2.54 \\
ER & 13.59 $\pm$ 2.28 & 12.96 $\pm$ 2.42 \\
LR & 4.84 $\pm$ 1.72 & 4.61 $\pm$ 1.55 \\
SLR & 18.48 $\pm$ 1.56 & 17.95 $\pm$ 1.88 \\
Offline & 0.34 $\pm$ 0.89 & 0.41 $\pm$ 0.76 \\
\hline
\end{tabular}
\end{table}

We observe that task order had minimal impact on overall forgetting rates, with differences between original and reversed sequences not reaching statistical significance (p > 0.05 for all methods). Forgetting was predominantly determined by temporal position rather than the specific concept. In both sequences, earlier tasks showed greater forgetting regardless of which concept was taught first. In the reverse sequence, the plushie concept (now taught first) experienced similar forgetting patterns to the dog concept in the original sequence. LR consistently demonstrated superior forgetting resistance in both task orderings, maintaining a TFR-IA of approximately 4.6-4.8 compared to other methods' 13-18 range.

These findings suggest that the temporal distance between tasks is a stronger determinant of forgetting than the specific semantic relationships between concepts, though both factors play a role. Future work might systematically investigate more complex task relationships and longer sequences to better understand these dynamics.

\section{Limitations}
\label{sec:limitations}

While our study demonstrates the effectiveness of LR for continual learning in text-to-image diffusion models, several limitations should be acknowledged. Our experimental design focused on a specific set of five common objects (dog, toy, cat, backpack, plushie) which, while diverse, represents only a small subset of potential user-specific concepts in real-world personalization scenarios. The relatively small number of sequential tasks provides initial evidence for LR's efficacy but does not address long-term scalability concerns when dealing with dozens or hundreds of concepts. Furthermore, our exclusive focus on object-domain concepts leaves unexplored the potential challenges of other domains such as artistic styles, scenes, or abstract concepts, which might exhibit different forgetting dynamics, especially in images containing multiple objects with complex relationships.

Our study focuses primarily on comparing LR against ER and naive fine-tuning, but lacks direct comparison with other established continual learning techniques from the broader literature, such as regularization-based methods (e.g., EWC, SI), parameter isolation approaches, or meta-learning strategies. This limits our ability to contextualize LR's performance within the full spectrum of continual learning solutions. 

Additionally, our work assumes distinct task boundaries, which may not reflect real-world scenarios where concept boundaries are blurred or tasks arrive in a more continuous manner.

From a methodological standpoint, our architectural choice of fine-tuning only the U-Net while keeping the VAE and text encoder frozen represents just one approach among many possible configurations. Alternative fine-tuning strategies might yield different results, potentially revealing other forgetting patterns or mitigation opportunities. Our implementation relies on reservoir sampling for memory buffer updates, a straightforward but potentially suboptimal approach compared to more sophisticated memory management strategies that could prioritize examples based on forgetting risk or representational importance. While we explored several hyperparameters through ablation studies, the vast hyperparameter space remains largely unexplored, leaving open the possibility that more optimal configurations exist beyond those we tested.

Our evaluation protocol, though comprehensive in its combination of Image Alignment, Text Alignment, and Diversity metrics, may not capture all relevant aspects of generation quality. In addition, while the Vendi score provides a useful measure of output variation, it has inherent limitations in capturing perceptually meaningful diversity that aligns with human judgments. The score may overweight minor variations or undervalue semantically significant differences. Furthermore, our separate treatment of catastrophic forgetting (via IA/TA) and mode collapse (via diversity) metrics may obscure the interrelated nature of these phenomena. An integrated metric that simultaneously captures both concept fidelity and generation diversity would provide a more holistic evaluation of generative continual learning performance. 

We were also constrained by computational resources, limiting our training to 800 steps per task and 10 runs per configuration, when more extensive training or additional runs might reveal different long-term patterns or more robust statistical findings. Another limitation in our evaluation protocol is the use of 10 images per prompt rather than the 50 images per prompt used in prior works like \citet{sun2023create}. While this reduction in evaluation samples was necessary due to computational constraints, it potentially limits the statistical robustness of our diversity measurements. Nevertheless, our approach still provided sufficient evidence to distinguish between methods, with clear trends showing with this smaller sample size. Future work with greater computational resources could extend our evaluation to larger sample sizes for additional validation.

Technical limitations include our focus on Stable Diffusion v1-4 as the base model, when newer diffusion architectures might exhibit different forgetting dynamics or benefit differently from LR strategies. Our current implementation replays latents exclusively at the input level of the U-Net, whereas alternative approaches, such as replaying intermediate activations or involving different network components, remain unexplored but potentially promising directions.

These limitations suggest several directions for future research, including scaling to more diverse and numerous tasks, exploring more sophisticated memory management strategies, testing across different model architectures, and incorporating better evaluation metrics.

\section{Conclusion}
\label{sec:chapter4_conclusion}

This chapter presented a comprehensive analysis of continual learning strategies for text-to-image diffusion models, with a particular focus on the effectiveness of Latent Replay compared to traditional approaches. Our experiments across five sequential tasks provide strong evidence that Latent Replay offers significant advantages over Naive fine-tuning and Experience Replay in mitigating catastrophic forgetting and preserving generative diversity in our setup.

Our analysis reveals several important insights about the effectiveness of different forgetting mitigation strategies. Latent Replay consistently outperforms both Naive fine-tuning and Experience Replay on metrics of Image Alignment, Text Alignment, and Diversity for early tasks, with statistical analysis confirming that these differences are significant, particularly for the earliest tasks in the sequence. The memory efficiency advantage is significant, with comparable memory footprints of approximately 30MB, Latent Replay with 480 latent vectors significantly outperforms Experience Replay with just 10 images, demonstrating that compact latent representations can effectively preserve knowledge of previous concepts while minimizing storage requirements.

Beyond mere retention of concepts, Latent Replay proved especially effective at preserving generative diversity across tasks. It maintained substantially higher diversity scores for early tasks compared to other methods (with the exception of the Offline upper bound), confirming its ability to mitigate mode collapse—a critical consideration often overlooked in generative continual learning. Our ablation studies identified optimal hyperparameter settings, particularly the memory-current task weighting parameter ($\lambda_{\text{memory}}=0.5$), which effectively balances retention of previous knowledge with adaptation to new tasks, addressing the fundamental stability-plasticity dilemma inherent in continual learning systems.

Contrary to our initial hypothesis, Similarity-Based Latent Replay consistently underperformed standard Latent Replay across multiple metrics and tasks. This surprising result suggests that semantic similarity alone may not be the optimal criterion for replay sample selection in this domain, and that maintaining broader coverage of the latent space through random sampling may be more effective for preserving generative capabilities.

The visual evidence from our qualitative analysis reinforces these quantitative findings. The generated outputs from Latent Replay maintained recognizable and diverse representations of earlier concepts even after learning multiple subsequent tasks, while Naive fine-tuning and Experience Replay often generated images that reflected the most recently learned concepts regardless of the prompt. This visual confirmation underscores the practical significance of our approach for maintaining concept fidelity in sequential learning scenarios.

These findings have significant implications for real-world applications of text-to-image diffusion models. Latent Replay provides a practical pathway to sequential personalization of diffusion models without requiring substantial memory resources or access to all historical data. The demonstrated efficiency of our approach makes it particularly suitable for resource-constrained environments, such as edge devices or applications with privacy constraints that limit data storage capabilities. Our qualitative results show that Latent Replay can preserve recognizable visual concepts across sequential tasks, enabling practical use cases where models must continually learn new concepts while retaining previously learned ones.

In the next chapter, we put these findings into a cohesive framework for continual learning in generative models, discuss broader implications for the field, and outline potential directions for future research.

%% file: chapter5.tex
\section{Summary of Contributions}

Our work has made several contributions to the field of continual learning for generative models. First, we established empirical evidence for catastrophic forgetting in sequentially fine-tuned text-to-image diffusion models. Naive fine-tuning exhibited severe forgetting, with Image Alignment scores dropping by over 16 percentage points and diversity decreasing significantly for early tasks (down to final values of 1.13-1.17 compared to the Offline upper bound of 1.68-1.72). This confirms that state-of-the-art diffusion models, despite their impressive capabilities, are not inherently resistant to catastrophic forgetting—a critical limitation for real-world applications requiring lifelong learning.

Second, we developed and implemented a novel Latent Replay framework specifically tailored to the architecture of text-to-image diffusion models. Unlike previous replay methods focused primarily on classification tasks, our approach addresses the unique challenges of preserving generative diversity and concept fidelity in high-dimensional image outputs. The framework leverages the pretrained VAE encoder in Stable Diffusion to store compact latent representations, significantly reducing memory requirements while maintaining essential visual information.

Through experimentation across five sequential tasks, we demonstrated that Latent Replay significantly outperforms both Naive fine-tuning and Experience Replay in mitigating catastrophic forgetting. Statistical analyses confirmed that Latent Replay achieves significantly higher IA, TA, and diversity scores for early tasks compared to alternative methods. Notably, even with a modest buffer of 480 latents (~31MB), Latent Replay maintained 77.59\% IA for the earliest task (dog)—substantially higher than Experience Replay's 65.64\% and Naive fine-tuning's 63.56\%. Similarly, Latent Replay preserved superior Diversity scores for the dog task (1.66) compared to Experience Replay (1.17) and Naive (1.13), demonstrating its effectiveness in mitigating mode collapse alongside catastrophic forgetting.

Our ablation studies provided several key insights for optimizing Latent Replay implementation. We found that memory size exhibits diminishing returns, with improvements in TFR-IA from small to medium buffers (4.84\% to 3.97\%, a 0.87\% gain) being less substantial than from medium to large (3.97\% to 2.16\%, a 1.81\% gain). Importantly, even the small Latent Replay configuration (480 latents) significantly outperformed the large Experience Replay configuration (100 images) on both TFR-IA (4.84\% vs. 7.26\%) and diversity (1.66 vs. 1.34), despite using only approximately 10\% of the memory footprint (~31MB vs. ~307MB). This conclusively demonstrates the memory efficiency advantage of latent representations.

We also found that task ordering had minimal impact on forgetting rates, with differences between original and reversed sequences not reaching statistical significance (p > 0.05). This suggests that temporal distance between tasks is a stronger determinant of forgetting than specific semantic relationships between concepts.

A balanced replay weight ($\lambda_{\text{memory}}=0.5$) effectively addresses the stability-plasticity dilemma, providing reasonable performance on both early tasks (77.59\% dog IA) and recent tasks (77.72\% plushie IA).

Surprisingly, Similarity-Based Latent Replay consistently underperformed standard Latent Replay across memory configurations, with TFR-IA values (18.48\%, 14.36\%, 10.02\%) even worse than naive fine-tuning (16.16\%) at small buffer sizes. This counter-intuitive finding suggests that similarity-based selection creates problematic bias in the replay buffer, preferentially selecting boundary cases while under-representing the distinctive characteristics of previous concepts.

Beyond addressing catastrophic forgetting, our research demonstrated Latent Replay's ability to preserve generative diversity. By maintaining higher Vendi scores across tasks, LR effectively mitigated mode collapse, a critical concern in generative continual learning. This contribution extends beyond diffusion models, highlighting the importance of diversity-aware evaluation metrics and memory management strategies for any generative continual learning scenario.

\section{Implications and Future Directions}

\subsection{Practical Applications}

Our research enables several practical applications for text-to-image diffusion models. For sequential personalization, users can incrementally teach diffusion models new custom concepts without requiring simultaneous access to all concepts. A creative professional could personalize their model with new characters, styles, or objects over time without losing earlier customizations.

The memory efficiency of Latent Replay makes continuous learning feasible even on devices with limited storage. A mobile application could maintain a compact latent buffer rather than storing a large dataset of raw images. By storing compressed latent representations rather than raw images, Latent Replay potentially reduces privacy concerns in continual learning systems. The latent space offers a form of information reduction that may better protect sensitive content in training examples.

\subsection{Theoretical Implications}

Beyond practical applications, our work offers broader insights for continual learning theory. Our success with Latent Replay suggests that continual learning may benefit more broadly from working in compressed representation spaces rather than raw input spaces. This aligns with cognitive science theories proposing that human memory systems store compressed, semantic representations rather than detailed sensory information.

Our experiments revealed an interesting tension between preserving generation fidelity and maintaining diversity. Methods like Similarity-based Latent Replay that selected "similar" latents often produced less diverse outputs, suggesting that some level of heterogeneity in replay examples may be necessary for maintaining generative diversity. Unlike classification tasks where preserving decision boundaries is paramount, generative continual learning requires preserving entire data distributions. Our results demonstrate that effective replay strategies must explicitly consider mode coverage, not just exemplar fidelity.

Interestingly, our Latent Replay approach shares conceptual similarities with theories of human memory consolidation. Neuroscience research suggests that the hippocampus rapidly encodes detailed sensory experiences, while the neocortex gradually extracts and consolidates abstract, compressed representations. Sleep replay in the hippocampus serves to transfer and integrate these memories into neocortical structures—conceptually similar to how our Latent Replay mechanism periodically "reminds" the model of compressed representations of past experiences.

\subsection{Future Research Directions}

Our work opens several promising directions for future research. Future work could explore more sophisticated approaches to memory management, such as importance-based sampling to identify and retain the most representative or informative latent vectors, perhaps using uncertainty measures or coverage metrics. Adaptive buffer allocation could dynamically adjust the buffer size for different tasks based on their complexity or observed forgetting rates. A hierarchical memory system with different retention policies, similar to psychological models of short-term and long-term memory, could further improve performance.

Combining Latent Replay with complementary continual learning strategies could yield further improvements. Latent distillation could transfer knowledge from previous model versions into the current model, using latent representations as an intermediate representation. Generative replay with latent guidance could create synthetic examples that fill gaps in the Latent Replay buffer, ensuring broader coverage of the data manifold. Parameter regularization combined with Latent Replay could jointly constrain both model weights and ensure data diversity.

Hybrid approaches combining raw images with latent representations might offer enhanced performance. Storing a small set of high-fidelity raw images alongside a larger latent buffer could preserve critical exemplars while maintaining broader coverage. This could be particularly effective for preserving fine details in certain concepts while leveraging the memory efficiency of latent representations for coverage.

Investigating Latent Replay's effectiveness for much longer task sequences (10-100 tasks) would reveal its scalability limits and inform design modifications needed for truly lifelong learning. Studying how forgetting accumulates over many sequential tasks and whether it follows predictable patterns could inform memory management. Determining how buffer size requirements scale with the number of tasks and developing principled approaches to buffer sizing would be valuable, as would exploring optimal strategies for when and how often to replay examples from each previous task as the task sequence grows.

Extending Latent Replay to more complex scenarios could involve applying it to multi-modal tasks where the model must learn across different modalities (e.g., text-to-image, text-to-audio, text-to-3D). Investigating whether Latent Replay can help models compositionally combine previously learned concepts in novel ways could lead to more flexible generative systems. Studying how Latent Replay performs in open-world settings where the model must distinguish between known and novel concepts would be particularly relevant for real-world applications.

Drawing further inspiration from neuroscience could yield novel continual learning solutions. Neuromorphic computing architectures that more closely mimic brain structure might offer inherent resistance to catastrophic forgetting. For instance, implementing sparse, distributed representations similar to those in the neocortex could enhance stability while maintaining plasticity. Investigating mechanisms analogous to synaptic consolidation, where connections between neurons are selectively strengthened or protected based on their importance to previously learned tasks, could offer new perspectives for parameter protection strategies. Additionally, temporal separation of learning phases—analogous to different sleep stages in mammals—could potentially improve memory consolidation in diffusion models. For example, alternating between "encoding" phases (learning new concepts) and "consolidation" phases (integrating past and present knowledge) might balance stability and plasticity more effectively than simultaneous learning. These bio-inspired approaches could complement Latent Replay by addressing different aspects of the continual learning challenge, potentially leading to more robust systems capable of truly lifelong adaptation.

Developing deeper theoretical insights into why and how Latent Replay works could involve quantifying the information preserved in latent representations and how it relates to catastrophic forgetting, studying how Latent Replay affects the loss landscape and optimization dynamics during sequential learning, and investigating how the stability of latent representations over sequential tasks correlates with forgetting resistance.

\section{Conclusion}

This thesis has demonstrated that Latent Replay offers a promising approach to mitigating catastrophic forgetting in text-to-image diffusion models. By storing and replaying compact latent representations, we have shown how generative models can maintain both concept fidelity and output diversity across sequential learning tasks while minimizing memory requirements.

Perhaps most surprisingly, our work revealed that random sampling of latent vectors outperforms similarity-based selection, challenging assumptions about optimal replay strategies and highlighting the importance of maintaining diverse concept coverage. This finding, along with our other contributions, extends beyond diffusion models to inform the broader field of continual learning for generative AI.

As diffusion models continue to advance and find applications across diverse domains, the need for effective continual learning strategies will only grow more pressing. Latent Replay provides one such strategy, opening a path toward more adaptive, resource-efficient, and practically useful generative systems capable of lifelong learning - an essential capability for deploying models in dynamic, open-world environments.

%% file: thesis.bbl
\begin{thebibliography}{44}
\providecommand{\natexlab}[1]{#1}
\providecommand{\url}[1]{\texttt{#1}}
\expandafter\ifx\csname urlstyle\endcsname\relax
  \providecommand{\doi}[1]{doi: #1}\else
  \providecommand{\doi}{doi: \begingroup \urlstyle{rm}\Url}\fi

\bibitem[CompVis(2022)]{compvis_sd}
CompVis.
\newblock Stable diffusion v1.4 model card, 2022.
\newblock URL \url{https://huggingface.co/CompVis/stable-diffusion-v1-4}.

\bibitem[Cong et~al.(2020)Cong, Zhao, Li, Wang, and Carin]{cong2020ganreplay}
Yulai Cong, Miaoyun Zhao, Jiawei Li, Sofia Wang, and Lawrence Carin.
\newblock Gan memory with no forgetting.
\newblock \emph{Advances in Neural Information Processing Systems (NeurIPS)}, 33:\penalty0 16481--16494, 2020.

\bibitem[Gal et~al.(2022)Gal, Alaluf, Atzmon, Patashnik, Bermano, Chechik, and Cohen-Or]{gal2022textual}
Rinon Gal, Yuval Alaluf, Yuval Atzmon, Or~Patashnik, Amit~H. Bermano, Gal Chechik, and Daniel Cohen-Or.
\newblock An image is worth one word: Personalizing text-to-image generation using textual inversion.
\newblock \emph{arXiv preprint}, 2022.

\bibitem[Gao and Liu(2023)]{gao2023ddgr}
Ruiying Gao and Wei Liu.
\newblock Ddgr: Continual learning with deep diffusion-based generative replay.
\newblock \emph{Proceedings of the International Conference on Machine Learning (ICML)}, pages 10744--10763, 2023.

\bibitem[Goodfellow et~al.(2014)Goodfellow, Pouget-Abadie, Mirza, Xu, Warde-Farley, Ozair, Courville, and Bengio]{goodfellow2014gan}
Ian Goodfellow, Jean Pouget-Abadie, Mehdi Mirza, Bing Xu, David Warde-Farley, Sherjil Ozair, Aaron Courville, and Yoshua Bengio.
\newblock Generative adversarial nets.
\newblock \emph{Advances in Neural Information Processing Systems (NeurIPS)}, 27, 2014.

\bibitem[Goodfellow et~al.(2013)Goodfellow, Mirza, Xiao, Courville, and Bengio]{goodfellow2013empirical}
Ian~J. Goodfellow, Mehdi Mirza, Da~Xiao, Aaron Courville, and Yoshua Bengio.
\newblock An empirical investigation of catastrophic forgetting in gradient-based neural networks.
\newblock \emph{arXiv preprint}, 2013.

\bibitem[Ho and Salimans(2021)]{ho2021classifier}
Jonathan Ho and Tim Salimans.
\newblock Classifier-free diffusion guidance.
\newblock \emph{NeurIPS 2021 Workshop on Deep Generative Models and Downstream Applications}, 2021.

\bibitem[Ho et~al.(2020)Ho, Jain, and Abbeel]{ho2020denoising}
Jonathan Ho, Ajay Jain, and Pieter Abbeel.
\newblock Denoising diffusion probabilistic models.
\newblock \emph{Advances in Neural Information Processing Systems (NeurIPS)}, 33:\penalty0 6840--6851, 2020.

\bibitem[Huang et~al.(2018)Huang, Vaswani, Uszkoreit, Shazeer, Simon, Hawthorne, Dai, Hoffman, Dinculescu, and Eck]{huang2018music}
Cheng-Zhi~Anna Huang, Ashish Vaswani, Jakob Uszkoreit, Noam Shazeer, Ian Simon, Curtis Hawthorne, Andrew~M. Dai, Matthew~D. Hoffman, Monica Dinculescu, and Douglas Eck.
\newblock Music transformer.
\newblock \emph{arXiv preprint}, 2018.

\bibitem[Kingma and Welling(2013)]{kingma2013vae}
Diederik~P Kingma and Max Welling.
\newblock Auto-encoding variational bayes.
\newblock \emph{arXiv preprint}, 2013.

\bibitem[Kirkpatrick et~al.(2017)Kirkpatrick, Pascanu, Rabinowitz, Veness, Desjardins, Rusu, Milan, Quan, Ramalho, Grabska-Barwińska, et~al.]{kirkpatrick2017overcoming}
James Kirkpatrick, Razvan Pascanu, Neil Rabinowitz, Joel Veness, Guillaume Desjardins, Andrei~A. Rusu, Kieran Milan, John Quan, Tiago Ramalho, Agnieszka Grabska-Barwińska, et~al.
\newblock Overcoming catastrophic forgetting in neural networks.
\newblock \emph{Proceedings of the National Academy of Sciences}, 114\penalty0 (13):\penalty0 3521--3526, 2017.

\bibitem[Kumari et~al.(2022)Kumari, Zhang, Zhang, Shechtman, and Zhu]{kumari2022multiconcept}
Nithin Kumari, Bowen Zhang, Richard Zhang, Eli Shechtman, and Jun-Yan Zhu.
\newblock Multiconcept customization of text-to-image diffusion.
\newblock \emph{arXiv preprint}, 2022.

\bibitem[Li et~al.(2019)Li, Qi, Torr, and Lukasiewicz]{li2019controllable}
Bowen Li, Xiaojuan Qi, Philip Torr, and Thomas Lukasiewicz.
\newblock Controllable text-to-image generation.
\newblock \emph{Advances in Neural Information Processing Systems (NeurIPS)}, 32, 2019.

\bibitem[Li and Hoiem(2017)]{li2017learning}
Zhizhong Li and Derek Hoiem.
\newblock Learning without forgetting.
\newblock \emph{IEEE Transactions on Pattern Analysis and Machine Intelligence (TPAMI)}, 40\penalty0 (12):\penalty0 2935--2947, 2017.

\bibitem[Lin et~al.(2024)Lin, Wu, Zhu, Zhai, and Xu]{lin2024dynamic}
Ching-Yao Lin, Hao Wu, Feng Zhu, Yihua Zhai, and Chang Xu.
\newblock Dynamic expansion for continual learning.
\newblock \emph{arXiv preprint}, 2024.

\bibitem[Mallya and Lazebnik(2018)]{mallya2018packnet}
Arun Mallya and Svetlana Lazebnik.
\newblock Packnet: Adding multiple tasks to a single network by iterative pruning.
\newblock \emph{Proceedings of the IEEE Conference on Computer Vision and Pattern Recognition (CVPR)}, pages 7765--7773, 2018.

\bibitem[McCloskey and Cohen(1989)]{mccloskey1989catastrophic}
Michael McCloskey and Neal~J. Cohen.
\newblock Catastrophic interference in connectionist networks: The sequential learning problem.
\newblock \emph{Psychology of Learning and Motivation}, 24:\penalty0 109--165, 1989.

\bibitem[Naeem et~al.(2022)Naeem, Baek, Woo, Park, Lee, Paull, Kim, and Kim]{naeem2022vendi}
Muhammad~Farrukh Naeem, Jooyoung Baek, Sanghyuk Woo, Joonseok Park, Junho Lee, Liam Paull, Sungjoo Kim, and Gunhee Kim.
\newblock The vendi score: A diversity evaluation metric for machine learning.
\newblock \emph{arXiv preprint}, 2022.
\newblock URL \url{https://arxiv.org/abs/2210.02410}.

\bibitem[Parisi et~al.(2019)Parisi, Kemker, Part, Kanan, and Wermter]{parisi2019continual}
German~I. Parisi, Ronald Kemker, Jose~L. Part, Christopher Kanan, and Stefan Wermter.
\newblock Continual lifelong learning with neural networks: A review.
\newblock \emph{Neural Networks}, 113:\penalty0 54--71, 2019.

\bibitem[Pellegrini et~al.(2020)Pellegrini, Graffieti, Lomonaco, and Maltoni]{pellegrini2020latent}
Lorenzo Pellegrini, Gabriele Graffieti, Vincenzo Lomonaco, and Davide Maltoni.
\newblock Latent replay for real-time continual learning.
\newblock \emph{IEEE/RSJ International Conference on Intelligent Robots and Systems (IROS)}, pages 4436--4443, 2020.

\bibitem[Popescu et~al.(2009)Popescu, Balas, Perescu-Popescu, and Mastorakis]{popescu2009mlp}
Marian-Ciprian Popescu, Valentina~E. Balas, Liliana Perescu-Popescu, and Nikos Mastorakis.
\newblock Multilayer perceptron and neural networks.
\newblock \emph{WSEAS Transactions on Circuits and Systems}, 8\penalty0 (7):\penalty0 579--588, 2009.

\bibitem[Radford et~al.(2021)Radford, Kim, Hallacy, Ramesh, Goh, Agarwal, Sastry, Askell, Mishkin, Clark, Krueger, and Sutskever]{radford2021clip}
Alec Radford, Jong~Wook Kim, Chris Hallacy, Aditya Ramesh, Gabriel Goh, Sandhini Agarwal, Girish Sastry, Amanda Askell, Pamela Mishkin, Jack Clark, Gretchen Krueger, and Ilya Sutskever.
\newblock Learning transferable visual models from natural language supervision.
\newblock In Marina Meila and Tong Zhang, editors, \emph{Proceedings of the International Conference on Machine Learning (ICML)}, volume 139 of \emph{Proceedings of Machine Learning Research}, pages 8748--8763. PMLR, 2021.
\newblock URL \url{http://proceedings.mlr.press/v139/radford21a.html}.

\bibitem[Ramesh et~al.(2021)Ramesh, Pavlov, Goh, Gray, Voss, Radford, Chen, and Sutskever]{ramesh2021zero}
Aditya Ramesh, Mikhail Pavlov, Gabriel Goh, Scott Gray, Chelsea Voss, Alec Radford, Mark Chen, and Ilya Sutskever.
\newblock Zero-shot text-to-image generation.
\newblock \emph{arXiv preprint}, 2021.

\bibitem[Rebuffi et~al.(2017)Rebuffi, Kolesnikov, Sperl, and Lampert]{rebuffi2017icarl}
Sylvestre-Alvise Rebuffi, Alexander Kolesnikov, Georg Sperl, and Christoph~H. Lampert.
\newblock icarl: Incremental classifier and representation learning.
\newblock \emph{Proceedings of the IEEE Conference on Computer Vision and Pattern Recognition (CVPR)}, pages 2001--2010, 2017.

\bibitem[Riemer et~al.(2018)Riemer, Cases, Ajemian, Liu, Rish, Tu, and Tesauro]{riemer2018learning}
Matthew Riemer, Ignacio Cases, Robert Ajemian, Miao Liu, Irina Rish, Yuhai Tu, and Gerald Tesauro.
\newblock Learning to learn without forgetting by maximizing transfer and minimizing interference.
\newblock \emph{arXiv preprint}, 2018.

\bibitem[Rombach et~al.(2022)Rombach, Blattmann, Lorenz, Esser, and Ommer]{rombach2022high}
Robin Rombach, Andreas Blattmann, Dominik Lorenz, Patrick Esser, and Björn Ommer.
\newblock High-resolution image synthesis with latent diffusion models.
\newblock \emph{Proceedings of the IEEE/CVF Conference on Computer Vision and Pattern Recognition (CVPR)}, pages 10684--10695, 2022.

\bibitem[Ruiz et~al.(2023)Ruiz, Li, Jampani, Pritch, Rubinstein, and Aberman]{ruiz2022dreambooth}
Nataniel Ruiz, Yuanzhen Li, Varun Jampani, Yael Pritch, Michael Rubinstein, and Kfir Aberman.
\newblock Dreambooth: Fine tuning text-to-image diffusion models for subject-driven generation.
\newblock \emph{Proceedings of the IEEE/CVF Conference on Computer Vision and Pattern Recognition (CVPR)}, pages 22500--22510, 2023.

\bibitem[Rusu et~al.(2016)Rusu, Rabinowitz, Desjardins, Soyer, Kirkpatrick, Kavukcuoglu, Pascanu, and Hadsell]{rusu2016progressive}
Andrei~A. Rusu, Neil Rabinowitz, Guillaume Desjardins, Hubert Soyer, James Kirkpatrick, Koray Kavukcuoglu, Razvan Pascanu, and Raia Hadsell.
\newblock Progressive neural networks.
\newblock \emph{arXiv preprint}, 2016.

\bibitem[Saharia et~al.(2022)Saharia, Chan, Saxena, Li, Whang, Denton, Ghasemipour, Gontijo~Lopes, Karagol~Ayan, Salimans, Ho, Fleet, and Norouzi]{saharia2022photorealistic}
Chitwan Saharia, William Chan, Saurabh Saxena, Lala Li, Jay Whang, Emily Denton, Seyed~Kamyar Ghasemipour, Raphael Gontijo~Lopes, Belinda Karagol~Ayan, Tim Salimans, Jonathan Ho, David Fleet, and Mohammad Norouzi.
\newblock Photorealistic text-to-image diffusion models with deep language understanding.
\newblock \emph{arXiv preprint}, 2022.

\bibitem[Seo et~al.(2023)Seo, Kang, and Park]{seo2023lfsgan}
Jeon-Hyeong Seo, Junmo Kang, and Gyeongbae Park.
\newblock Lfs-gan: Lifelong few-shot image generation.
\newblock \emph{Proceedings of the IEEE/CVF International Conference on Computer Vision (ICCV)}, pages 11322--11332, 2023.

\bibitem[Shi et~al.(2023)Shi, Siddharth, Brooks, Torr, and Liao]{shi2023deep}
Yuanbo Shi, Nikhil Siddharth, Philip~A. Brooks, Philip~H.S. Torr, and Yanjie Liao.
\newblock Deep generative models on 3d representations: A survey.
\newblock \emph{arXiv preprint}, 2023.

\bibitem[Shin et~al.(2017)Shin, Lee, Kim, and Kim]{shin2017continual}
Hanul Shin, Jung~Kwon Lee, Jaehong Kim, and Jiwon Kim.
\newblock Continual learning with deep generative replay.
\newblock \emph{Advances in Neural Information Processing Systems (NeurIPS)}, 30, 2017.

\bibitem[Smith et~al.(2023)Smith, Tian, Halbe, Hsu, and Kira]{smith2023continual}
James~Seale Smith, Jianwei Tian, Suhail Halbe, Yung-Chieh Hsu, and Zsolt Kira.
\newblock Continual diffusion: Continual customization of text-to-image diffusion with c-lora.
\newblock \emph{arXiv preprint}, 2023.

\bibitem[Smith et~al.(2024)Smith, Hsu, Zhang, Hua, Kira, Shen, and Jin]{smith2024stamina}
James~Seale Smith, Yung-Chieh Hsu, Ling Zhang, Te~Hua, Zsolt Kira, Yujun Shen, and Huiwen Jin.
\newblock Continual diffusion with stamina: Stack-and-mask incremental adapters.
\newblock \emph{arXiv preprint}, 2024.

\bibitem[Song and Ermon(2020)]{song2020scorebased}
Yang Song and Stefano Ermon.
\newblock Improved techniques for training score-based generative models.
\newblock \emph{Advances in Neural Information Processing Systems (NeurIPS)}, 33:\penalty0 12438--12448, 2020.

\bibitem[Sun et~al.(2024)Sun, Liang, Dong, Li, Ding, and Cong]{sun2023create}
Gan Sun, Wenbo Liang, Jun Dong, Jun Li, Zhengming Ding, and Yulai Cong.
\newblock Create your world: Lifelong text-to-image diffusion.
\newblock \emph{IEEE Transactions on Pattern Analysis and Machine Intelligence}, 2024.

\bibitem[van~den Oord et~al.(2016)van~den Oord, Dieleman, Zen, Simonyan, Vinyals, Graves, Kalchbrenner, Senior, and Kavukcuoglu]{vandenOord2016wavenet}
Aäron van~den Oord, Sander Dieleman, Heiga Zen, Karen Simonyan, Oriol Vinyals, Alex Graves, Nal Kalchbrenner, Andrew Senior, and Koray Kavukcuoglu.
\newblock Wavenet: A generative model for raw audio.
\newblock \emph{arXiv preprint}, 2016.

\bibitem[Wortsman et~al.(2020)Wortsman, Ramanujan, Liu, Kembhavi, Rastegari, Yosinski, and Farhadi]{wortsman2020supermasks}
Mitchell Wortsman, Vivek Ramanujan, Rosanne Liu, Aniruddha Kembhavi, Mohammad Rastegari, Jason Yosinski, and Ali Farhadi.
\newblock Supermasks in superposition.
\newblock \emph{Advances in Neural Information Processing Systems (NeurIPS)}, 33:\penalty0 15173--15184, 2020.

\bibitem[Wu et~al.(2018)Wu, Herranz, Liu, Wang, van~de Weijer, and Raducanu]{wu2018memorygan}
Chenshen Wu, Luis Herranz, Xialei Liu, Yaxing Wang, Joost van~de Weijer, and Bogdan Raducanu.
\newblock Memory replay gans: Learning to generate images from new categories without forgetting.
\newblock \emph{Advances in Neural Information Processing Systems (NeurIPS)}, 31, 2018.

\bibitem[Zeng et~al.(2022)Zeng, Vahdat, Williams, Gojcic, Litany, Fidler, and Kreis]{zeng2022lion}
Xiaohui Zeng, Arash Vahdat, Francis Williams, Zan Gojcic, Or~Litany, Sanja Fidler, and Karsten Kreis.
\newblock Lion: Latent point diffusion models for 3d shape generation.
\newblock \emph{arXiv preprint}, 2022.

\bibitem[Zenke et~al.(2017)Zenke, Poole, and Ganguli]{zenke2017continual}
Friedemann Zenke, Ben Poole, and Surya Ganguli.
\newblock Continual learning through synaptic intelligence.
\newblock \emph{Proceedings of the International Conference on Machine Learning (ICML)}, pages 3987--3995, 2017.

\bibitem[Zhai et~al.(2021)Zhai, Chen, and Mori]{zhai2021lit}
Miaoyun Zhai, Liang Chen, and Greg Mori.
\newblock Hyper-lifelonggan: Scalable lifelong learning for image conditioned generation.
\newblock \emph{Proceedings of the IEEE/CVF Conference on Computer Vision and Pattern Recognition (CVPR)}, pages 2246--2255, 2021.

\bibitem[Zhang et~al.(2024)Zhang, Zhou, Lin, Ye, Zhu, Wang, Gao, Wang, and Liang]{zhang2024clog}
Haotian Zhang, Junting Zhou, Haowei Lin, Hang Ye, Jianhua Zhu, Zihao Wang, Liangcai Gao, Yizhou Wang, and Yitao Liang.
\newblock Clog: Benchmarking continual learning of image generation models.
\newblock \emph{arXiv preprint arXiv:2406.04584}, 2024.

\bibitem[Zhang et~al.(2023)Zhang, Rao, and Agrawala]{zhang2023controlnet}
Lvmin Zhang, Anyi Rao, and Maneesh Agrawala.
\newblock Adding conditional control to text-to-image diffusion models.
\newblock \emph{Proceedings of the IEEE/CVF International Conference on Computer Vision (ICCV)}, pages 3836--3847, 2023.

\end{thebibliography}
